\documentclass[10pt,twocolumn,letterpaper]{article}

\usepackage[pagenumbers]{cvpr} 

\definecolor{cvprblue}{rgb}{0.21,0.49,0.74}
\usepackage[pagebackref,breaklinks,colorlinks,allcolors=cvprblue]{hyperref}

\usepackage{amsmath}
\usepackage{adjustbox}
\usepackage{graphicx}

\def\paperID{26} 
\def\confName{CVPR}
\def\confYear{2025}

\title{DeclutterNeRF: Generative-Free 3D Scene Recovery for Occlusion Removal}

\author{
  Wanzhou Liu$^{1}$ \quad 
  Zhexiao Xiong$^{1}$ \quad 
  Xinyu Li$^{2}$ \quad 
  Nathan Jacobs$^{1}$ \\
  $^{1}$Washington University in St. Louis \quad
  $^{2}$Georgia Institute of Technology \\
  \small \texttt{\{l.wanzhou, x.zhexiao, jacobsn\}@wustl.edu, xli3212@gatech.edu}
}

\begin{document}

\twocolumn[{%
\renewcommand\twocolumn[1][]{#1}%
\maketitle
\begin{center}
    \centering
    \includegraphics[trim={0cm 1.6cm 0cm 1.6cm},width=\linewidth]{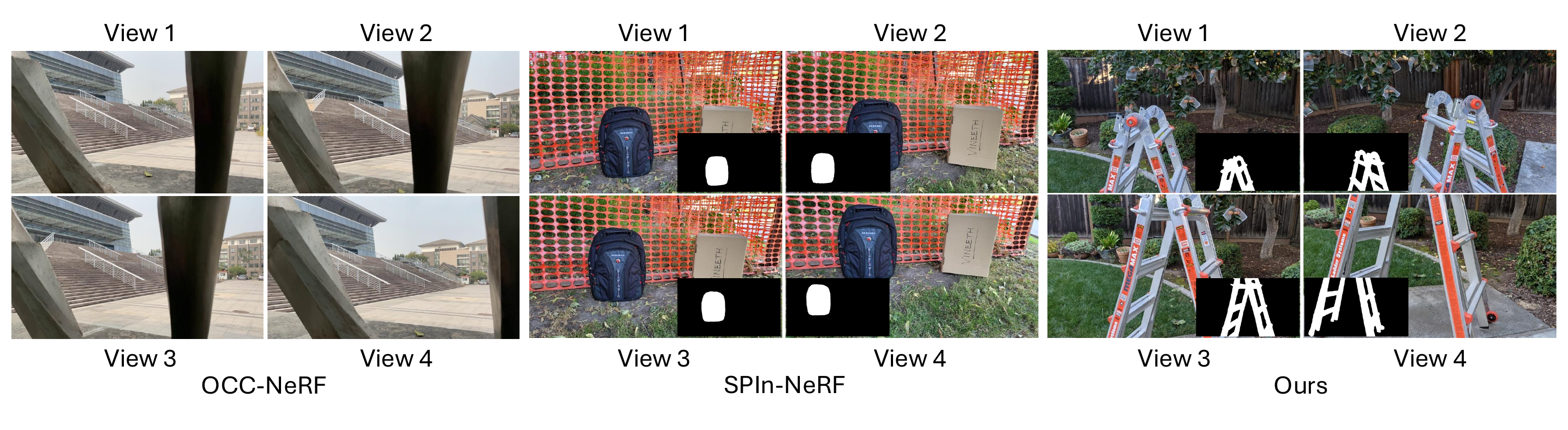}
    \captionof{figure}{\textbf{Comparison of Mainstream Occlusion Removal Datasets.} DeclutterSet is a new dataset reflecting real-world challenges and complexity in occlusion removal. For each dataset, we show four evenly spaced views per scene. As seen in both the RGB images and masks, DeclutterSet exhibits: (i) wider distance distribution, (ii) larger occluded regions, (iii) greater relative motion between viewpoints and occluders, and (iv) more uncertain occluder shapes and mask layouts. In contrast, the OCC-NeRF dataset~\cite{zhu2023occlusion} does not employ masks during selection, limiting it to foreground occlusions and requiring a strict separation between foreground and background, reducing its suitability for complex scenarios. SPIn-NeRF~\cite{spinnerf} provides limited challenge for cross-view consistency, as it is constrained to small viewpoint variations, keeping occluders and background nearly static across rendered views. A detailed analysis is provided in Sec.~\ref{sec:experimental_setup}.
    }\label{fig:teaser}
\end{center}%
\vspace{6pt}
}]

\maketitle

\begin{abstract}

Recent novel view synthesis (NVS) techniques, including Neural Radiance Fields (NeRF) and 3D Gaussian Splatting (3DGS) have greatly advanced 3D scene reconstruction with high-quality rendering and realistic detail recovery.
Effectively removing occlusions while preserving scene details can further enhance the robustness and applicability of these techniques. 
However, existing approaches for object and occlusion removal predominantly rely on generative priors, which, despite filling the resulting holes, introduce new artifacts and blurriness. 
Moreover, existing benchmark datasets for evaluating occlusion removal methods lack realistic complexity and viewpoint variations. 
To address these issues, we introduce \textbf{DeclutterSet}, a novel dataset featuring diverse scenes with pronounced occlusions distributed across foreground, midground, and background, exhibiting substantial relative motion across viewpoints. 
We further introduce \textbf{DeclutterNeRF}, an occlusion removal method free from generative priors. 
DeclutterNeRF introduces joint multi-view optimization of learnable camera parameters, occlusion annealing regularization, and employs an explainable stochastic structural similarity loss, ensuring high-quality, artifact-free reconstructions from incomplete images. 
Experiments demonstrate that DeclutterNeRF significantly outperforms state-of-the-art methods on our proposed DeclutterSet, establishing a strong baseline for future research. 
The code and data are available at \href{https://github.com/wanzhouliu/declutter-nerf}{DeclutterNeRF}.

\end{abstract}    
\section{Introduction}
\label{sec:intro}

Recent novel view synthesis (NVS) techniques including Neural Radiance Fields (NeRF)~\cite{mildenhall2020nerf} and 3D Gaussian Splatting (3DGS)~\cite{kerbl20233d} have advanced realistic and efficient 3D scene reconstruction.
Removing unwanted objects from rendered scenes would further enhance the flexibility and applicability of these methods for applications in AR, VR, robotics, and autonomous driving~\cite{zhao2024hybridocc, chubin2023occnerf, pan2024renderocc, tonderski2024neurad}.
Notably, these real-world scenarios often involve far more complex scene settings than current mainstream occlusion and object removal benchmarks and demand reliable rendering results.
This remains a major challenge in 3D reconstruction and calls for a rethinking of existing approaches.

Traditional methods rely on stereo geometry for occlusion handling~\cite{zitnick2007stereo, hirschmuller2007stereo, wang2008modeling, 10.1145/383259.383296, cheng2010repfinder}. With the advent of neural view synthesis, filtering-based and optimization-driven techniques have emerged for occlusion selection and removal~\cite{zhu2023occlusion, Sabour2023RobustNeRFID, Sabour2024SpotlessSplatsID}, but their effectiveness remains limited by overly simplified scene assumptions. 
Recent NeRF and 3DGS approaches have embraced generative models~\cite{suvorov2022resolution, kirillov2023segment, ravi2024sam2, liu2022nerf, Weder2023Removing, spinnerf, yin2023ornerf, wang2024innerf360, weber2023nerfiller, MVIPNeRF, Chen2024SingleMaskIF, Lin2024TamingLD, Cao2024MVInpainterLM, Ni2025EfficientI3, Salimi2025GeometryAwareDM, Huang20253DGI, Shi2025IMFine3I, Wu2025AuraFusion360AU, Wang2024GScreamL3}, which can marginally improve reconstruction quality but often introduce significant computational overhead, limiting their practicality. 
Importantly, most existing methods are developed on OCC-NeRF~\cite{zhu2023occlusion} and SPIn-NeRF~\cite{spinnerf} datasets, both of which introduce limiting assumptions. As shown in Fig.~\ref{fig:teaser}, OCC-NeRF considers only foreground occlusions, while SPIn-NeRF assumes all objects lie on a background plane with minimal relative motion across viewpoints. When these assumptions are violated, \ie, when objects are at different distances or exhibit large motion relative to viewpoints, both generative and non-generative methods struggle, leading to severe artifacts, inconsistent geometry, and unrealistic texture.

To address these limitations, we introduce DeclutterSet, a novel dataset designed to reflect real-world occlusion complexities. Unlike the settings in the previous datasets, DeclutterSet carefully considers the spatial distribution of objects at varying distances, ensuring that occlusions exhibit substantial motion relative to obvious viewpoint changes. 
By incorporating diverse scenarios where foreground, midground, and background objects shift across views, DeclutterSet provides a more realistic benchmark for evaluating occlusion removal methods.

Building on our DeclutterSet benchmark, we propose DeclutterNeRF, a straightforward optimization-driven approach that leverages NeRF's inherent cross-view consistency to tackle recovery after occlusion removal. Rather than relying on generative models, we demonstrate that targeted improvements to the classic NeRF framework can achieve superior results for this task. Using SAM~\cite{kirillov2023segment} for initial occlusion segmentation, our approach focuses on optimizing reconstruction from visible regions with minimal computational overhead. We first observe that occlusion presence alters camera parameter estimation, leading to suboptimal pose reconstruction. Inspired by camera posture estimation methods in 3D reconstruction~\cite{wang2021nerf, Levy2023MELONNW, Fu2023COLMAPFree3G}, we incorporate camera parameter optimization as a learnable component, allowing multi-view joint optimization to correct pose shifts and mitigate local minima issues. To ensure stable learning after occlusion removal, where only limited pixels are available for rendering, we propose Occlusion Annealing Regularization (OAR), which reduces the impact of occluded regions, improving training stability and preventing overfitting. Finally, we employ Stochastic Structural Similarity Loss (S3IM)~\cite{xie2023s3im} to address the long-tail distribution of background pixels caused by non-fixed occlusion regions, which leads to imbalanced ray sampling. Our experiments demonstrate that these targeted optimizations enable DeclutterNeRF to significantly outperform both previous optimization-based and generative methods in occlusion removal and recovery tasks, while maintaining computational efficiency. We summarize our contributions as follows:

\begin{itemize}
    \item We introduce DeclutterSet, a novel occlusion removal dataset with diverse real-world occlusion scenarios, capturing multi-position spatial distributions and viewpoint-dependent changes.
    \item We propose DeclutterNeRF, a generative-free occlusion removal framework that reconstructs 3D scenes using NeRF's implicit multi-view consistency, ensuring reliable and high-quality results without additional training costs.
    \item We highlight the impact of occlusion removal on camera pose estimation, incorporate multi-view joint learnable camera parameter optimization, and propose Occlusion Annealing Regularization (OAR) to improve robust rendering progress and stabilize training after occlusion removal, mitigating local minima and overfitting issues.
    \item We theoretically and experimentally validate the ``Unreasonable Effectiveness'' of random structural similarity~\cite{xie2023s3im}, showing its broader applicability in our task.
\end{itemize}

\begin{figure*}[t]
    \centering
    \includegraphics[width=\linewidth]{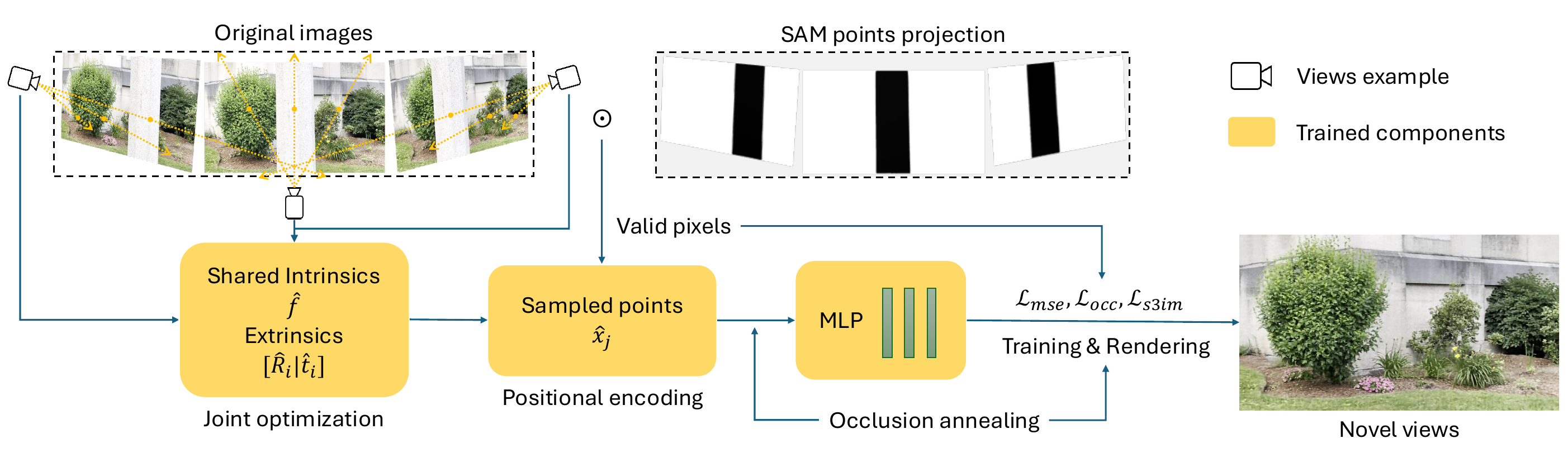}
    \caption{\textbf{Overview of Our Optimization Framework.} Our method builds on the NeRF architecture to recover occluded scenes without generative priors. Starting with a single-view SAM segmentation method~\cite{yin2023ornerf}, we propagate occluder masks across views via stereo matching. Camera parameters are jointly optimized with masked photometric supervision to correct occlusion-induced pose errors (Sec.~\ref{methodcam}). To stabilize training and mitigate overfitting to visible regions, we propose Occlusion Annealing Regularization (Sec.~\ref{methodann}). The Stochastic Structural Similarity loss (Sec.~\ref{methodssim}) enforces global consideration across views and improves reconstruction under long-tail visibility.
    }
    \label{fig:nerf_overview}
    \vspace{-4mm}
\end{figure*}

\section{Related Work}
\label{sec:related_work}

\textbf{Occlusion and Object removal.} Traditional approaches to occlusion removal and object deletion often rely on stereo geometry and multi-view consistency cues, such as disparity maps~\cite{jonna2017stereo}, dense flow fields~\cite{xue2015computational}, and synthetic apertures~\cite{vaish2006reconstructing}. Later, deep learning techniques leveraging temporal information~\cite{du2018accurate} and optical flow~\cite{liu2020learning} emerged. These methods often have limitations due to restricted camera movement and poor generalization to novel viewpoints.

Recent progress in novel view synthesis is driven by NeRF and 3DGS~\cite{mildenhall2020nerf, kerbl20233d}, which offer high-fidelity reconstruction via ray tracing and real-time performance via point-based rendering. Both have been extended to generative-free and optimization-based occlusion removal~\cite{zhu2023occlusion, Sabour2023RobustNeRFID, Sabour2024SpotlessSplatsID} with simplified assumptions. Concretely, OCC-NeRF~\cite{zhu2023occlusion} removes close-range occluders based on bidirectional depth inconsistency, but assumes all occlusions lie in the foreground, resulting in missing foreground details indiscriminately. RobustNeRF~\cite{Sabour2023RobustNeRFID} and SpotlessSplats~\cite{Sabour2024SpotlessSplatsID} handle transient occlusions by removing outliers that appear sporadically across views, but are not designed for persistent or structured obstacles. In contrast, our approach flexibly removes occlusions across diverse object categories, distributions and varying depth ranges.

\noindent\textbf{2D \& 3D Inpainting.}  
Early image inpainting techniques restore missing regions via local texture synthesis or structural propagation, using exemplar-based~\cite{ballester2001filling} or PDE-driven~\cite{10.1145/344779.344972} approaches. With deep learning, 2D inpainting evolved to adopt RGB priors (\eg, LaMa~\cite{suvorov2022resolution}) and explicit depth cues~\cite{liu2022nerf, Weder2023Removing, spinnerf, yin2023ornerf, wang2024innerf360}, achieving recovery reconstruction in single and multi-view settings. 

Beyond the 2D inpainting methods, diffusion-based techniques have been integrated into 3D reconstruction process with NeRF and 3DGS~\cite{MVIPNeRF, weber2023nerfiller, Chen2024SingleMaskIF, Lin2024TamingLD, Salimi2025GeometryAwareDM, Liu2024InFusionI3, Huang20253DGI, Wang2024GScreamL3, Shi2025IMFine3I, Wu2025AuraFusion360AU}. MVIP-NeRF~\cite{MVIPNeRF} leverages diffusion and cross-view distillation to hallucinate missing content, but at the cost of high memory and training time. GScream~\cite{Wang2024GScreamL3} incorporates depth supervision, which makes it sensitive to the quality of depth estimation. While these generative approaches aim to improve visual fidelity, they are also prone to introducing artifacts, suffer from geometric inconsistencies, and incur significant computational overhead, which limits their scalability in real-world applications.

\noindent\textbf{3D Reconstruction from Limited Pixels.} 
Traditional approaches for 3D reconstruction under incomplete observations rely on stereo correspondence~\cite{zitnick2007stereo, hirschmuller2007stereo}, image-based priors~\cite{wang2008modeling, cheng2010repfinder}, or local texture synthesis~\cite{10.1145/383259.383296}. Segment-based stereo matching improves robustness at object boundaries, while image quilting demonstrates the feasibility of patch-based texture propagation. Despite their contributions, these methods typically involve handcrafted priors and computationally intensive optimization, limiting efficiency and scalability in complex scenes.

The availability of powerful segmentation tools such as SAM~\cite{kirillov2023segment} and SAM2~\cite{ravi2024sam2} has popularized object-level masking in novel view synthesis~\cite{yin2023ornerf, Ni2025EfficientI3, Huang20253DGI, Wang2024GScreamL3}. This trend amplifies the need for 3D reconstruction from incomplete images, especially when large scene portions are masked out. In this work, we focus on recovering occluded geometry directly from the visible regions, without relying on synthetic content. By leveraging NeRF’s cross-view consistency and introducing optimization methods to occlusion scenarios, our method ensures structurally coherent and robust reconstruction under different occlusion scenarios.

\begin{figure}[t]
    \centering
    \begin{minipage}{0.49\columnwidth}
        \centering
        \includegraphics[width=\linewidth]{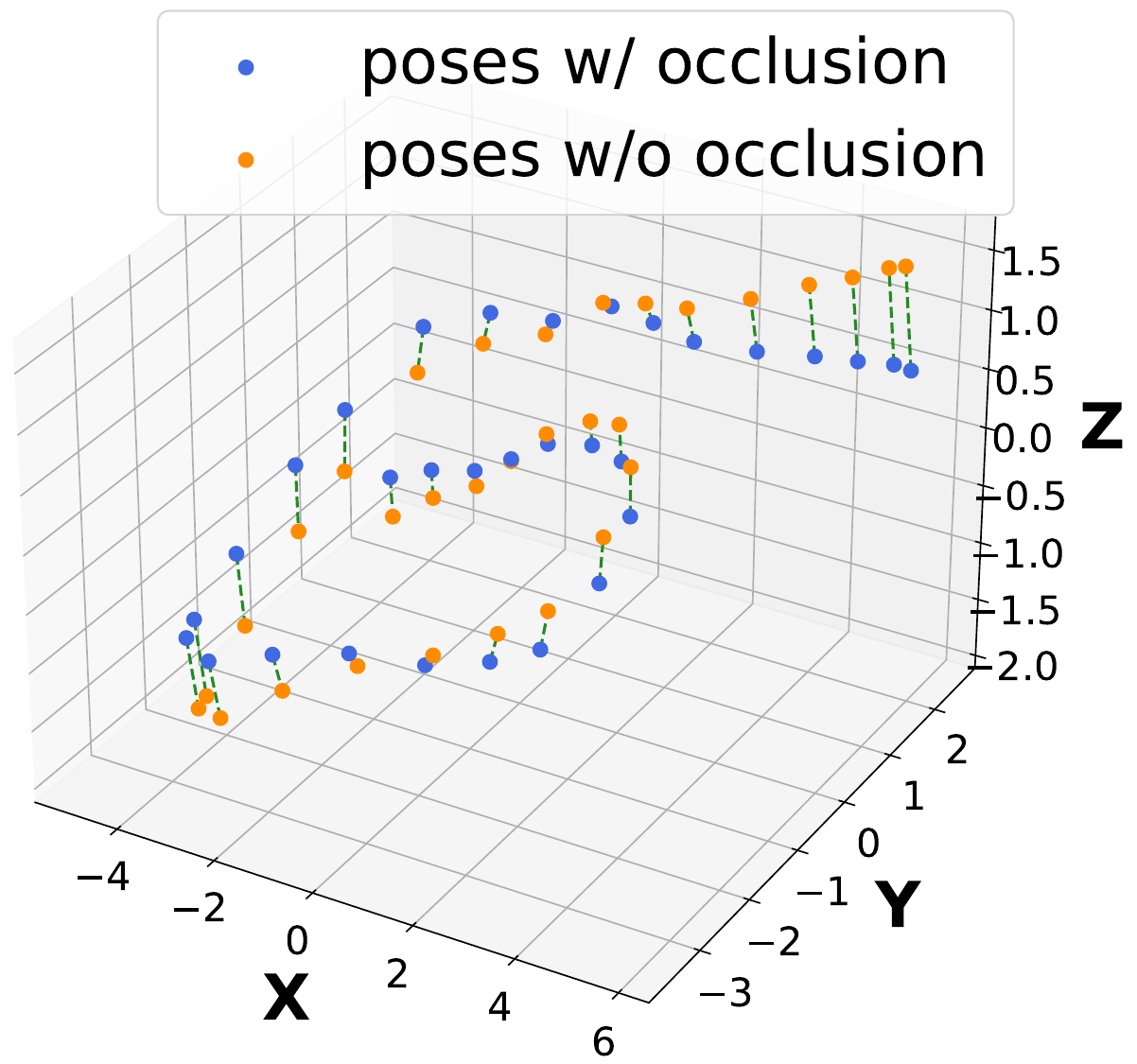}
    \end{minipage}
    \hfill
    \begin{minipage}{0.49\columnwidth}
        \centering
        \includegraphics[width=\linewidth]{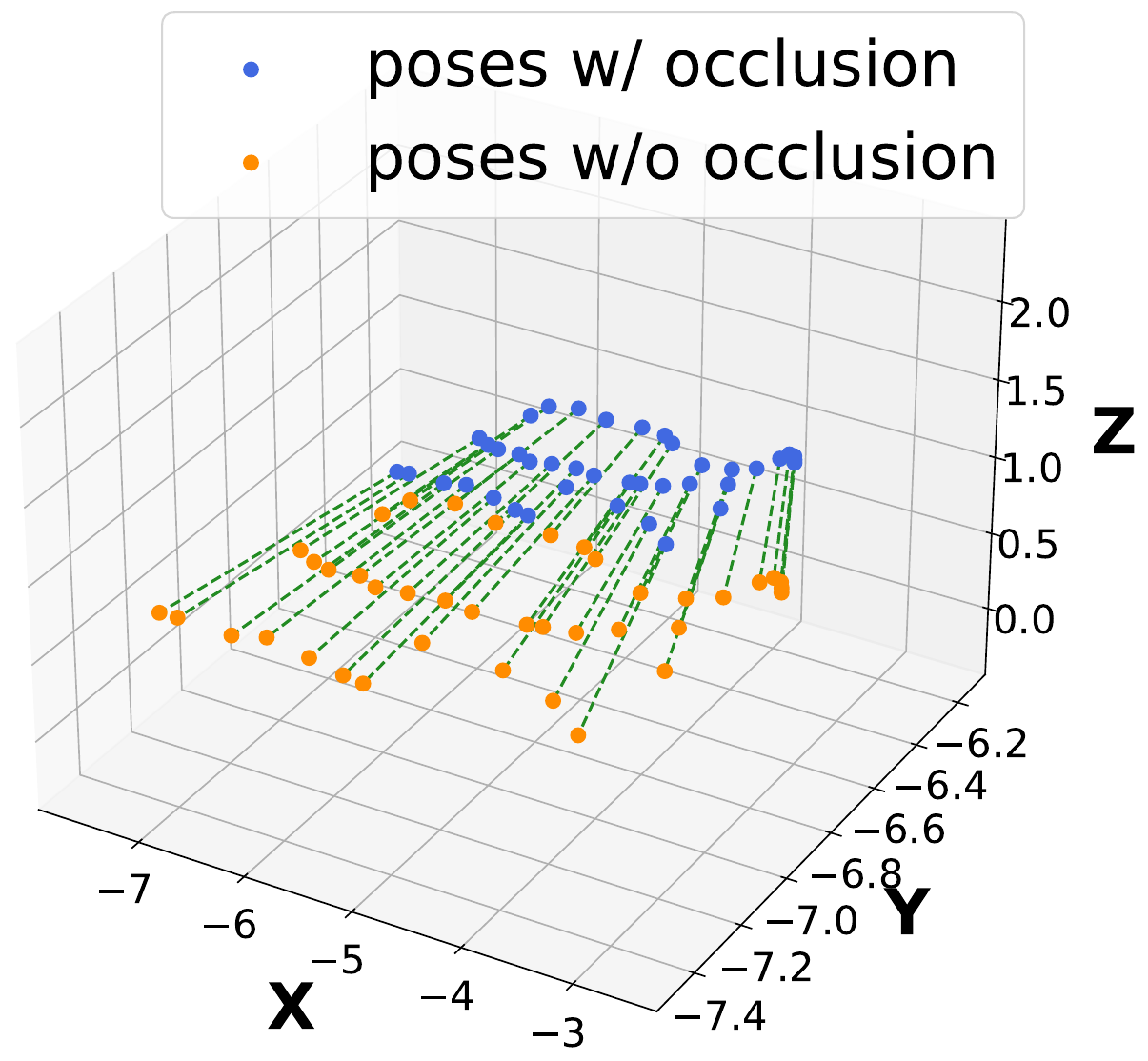}
    \end{minipage}
    \caption{\textbf{Visualization of the Impact of Obstacles on Pose Estimation.} Structure-from-motion methods, including the widely used COLMAP~\cite{schoenberger2016sfm, schoenberger2016mvs} and the recently proposed GLOMAP~\cite{pan2024glomap}, struggle to maintain stable camera pose estimation after occlusion is removed. This is illustrated in the Ladder scene (left) and the Lamp Post scene (right). Green dashed lines connect corresponding samples before and after occlusion removal, highlighting positional shifts. Axes are rotated for clearer visualization.}
    \label{fig_pose}
    \vspace{-4mm}
\end{figure}

\section{Method}
\label{sec:method}

\subsection{Preliminaries}
\label{sec:preliminaries}

\textbf{Neural Radiance Fields.}  NeRF~\cite{mildenhall2020nerf} is an approach to view synthesis, encoding scenes as implicit continuous volumetric functions. 
Let $\mathbf{x} = (x, y, z)$ denote a 3D point in space and $\mathbf{d} = (\theta, \phi)$ represent a viewing direction. 
The core of NeRF is a multi-layer perceptron (MLP) $F_\Theta: (\mathbf{x}, \mathbf{d}) \rightarrow (\mathbf{c}, \sigma)$, where $\Theta$ are the parameters of MLP. It maps a 3D location and viewing direction to a color $\mathbf{c} = (r, g, b)$ and volume density $\sigma$. The camera poses $\mathbf{x}$ are primarily derived from the pose estimation tool COLMAP~\cite{schoenberger2016sfm, schoenberger2016mvs}.

\noindent\textbf{Positional Encoding in NeRF.} 
Directly optimizing over raw inputs $(\mathbf{x}, \mathbf{d})$ makes it difficult for NeRF to capture high-frequency details.
To mitigate this, the mapping $F_\Theta$ is decomposed as $F'_\Theta \circ \lambda$, where $\lambda$ encodes inputs into a higher-dimensional space $\mathbb{R}^{2L}$.
The positional encoding $\lambda(\cdot)$ is defined as:
\begin{equation}
\begin{split}
\lambda(p) = \big( &\sin(2^0 \pi p), \cos(2^0 \pi p), \dots, \\
&\sin(2^{L-1} \pi p) \cos(2^{L-1} \pi p) \big)
\end{split}
\end{equation}
where $L$ is a hyperparameter that controls the highest encoded frequency. The encoding is applied to each component of the 3D position vector $\mathbf{x}$ (normalized to $[-1, 1]$) and the viewing direction vector $\mathbf{d}$ (unit vector in $[-1, 1]$). In most cases, it has $L = 10$ for $\lambda(\mathbf{x})$ and $L = 4$ for $\lambda(\mathbf{d})$.

\noindent\textbf{Stochastic Structural Similarity (S3IM).} S3IM\cite{xie2023s3im} is a patch-based, stochastic variant of SSIM~\cite{wang2004image}, designed to introduce global structural supervision into NeRF training. Instead of point-wise loss MSE or local supervise SSIM, S3IM computes SSIM between randomly sampled image patches, capturing non-local and cross-view structural consistency.
Given rendered radiance fields $\hat{R}$ and ground-truth images $R$, the loss is computed by sampling $M$ patch pairs ${P^{(m)}(\hat{C}), P^{(m)}(C)}_{m=1}^M$ from both rendered images $\hat{C}$ and ground-truth $C$. Each patch is of size $K \times K$, sampled with stride $s = K$. The final S3IM loss is defined as:
\begin{equation}
\text{S3IM}(\hat{\mathcal{R}}, \mathcal{R}) = \frac{1}{M} \sum_{m=1}^M \text{SSIM}(\mathcal{P}^{(m)}(\hat{C}), \mathcal{P}^{(m)}(C))
\end{equation}
where $\text{SSIM}(\cdot, \cdot)$ denotes the structural similarity between corresponding patches.

\subsection{Joint Optimization for Camera Parameters}
\label{methodcam}

Our concern about the impact of occlusions on camera parameter estimation originates from classical insights in computer vision and graphics~\cite{wang1994representing, szeliski1998geometrically}. As demonstrated in Fig.~\ref{fig_pose}, occluders can disturb pose estimation, leading to reconstruction degradation. A straightforward solution would be to recalibrate camera parameters after occlusion removal using the cleaner, occlusion-free 2D observations, which typically enhances reconstruction quality. However, for fair comparison and to test robustness, we retain the original camera parameters estimated under occlusions. Our goal is to leverage the occlusion-free setting as a means to further refine these parameters. To this end, we incorporate the camera parameters into our joint optimization framework. With photometric loss as the major supervision, our framework progressively corrects camera poses, resulting in improved reconstruction performance.

Similar work was proposed by~\cite{wang2021nerf, Levy2023MELONNW}, which aimed to completely resolve NeRF's dependence on camera parameters. However, these approaches introduced the problem of easily falling into local minima during training. OCC-NeRF~\cite{zhu2023occlusion}, which also employs this method, often produces poor reconstruction quality and can only handle small camera position movements due to this issue. 
Based on our analysis, they sample only one single image per training iteration, although OCC-NeRF utilized a pretrained ResNet to extract features for the warped feature map, such high-level feature extraction and projection transformation cannot meet NeRF's requirements for fine-grained geometric details, making it difficult to effectively handle subtle differences in camera poses. 

The improvement in our method is intuitive and easy to understand. 
As illustrated in the initial sampling process in Fig.~\ref{fig:nerf_overview}, instead of the traditional approach of sampling from a single view, we jointly optimize across all views by uniformly sampling valid pixels from the entire image set, enabling simultaneous refinement of camera parameters for all viewpoints.
This approach is well-founded. 
Firstly, previous results~\cite{wang2021nerf} show that when NeRF parameters are trapped in local minima, focal length parameters often deviate significantly from calibrated values.
Since focal length is shared across all input views, distributing focal length sampling across all views contributes to its stable optimization. Second, as illustrated in the sampling process in Fig.~\ref{fig:nerf_overview}, where multiple intersecting rays are intentionally drawn as a demonstrative example, the volumetric rendering process handles multiple intersecting rays during multi-view optimization. This leverages the advantage of stereoscopic input, where intersecting rays jointly optimize shared parameters, enhancing stability. This is also consistent with the recent prevailing trends in NeRF training methods. Finally, by adjusting the learning rates and implementing a delayed camera optimization strategy, we avoid potential local minima issues during training.

Let $\Theta$ denote the parameters of our MLP, $\phi$ represent the camera parameters, $\phi_t$ represent the current camera parameters at step $t$, $T$ be the total number of the iteration and $I$ be the set of input images. We formulate our joint optimization objective as:
\vspace{-11pt}

\begin{equation}
\resizebox{\linewidth}{!}{$
\begin{aligned}
\arg\min_{\Theta, \{\phi_t\}_{t=t_c}^T} & \sum_{t=1}^T \mathbb{E}_{B \sim \mathcal{U}(I, b)} \left[ \frac{1}{|B|} \sum_{(i,j) \in B} \mathcal{L}_{\text{photo}}(R_{\Theta, \phi_t}(r_{i,j}), I_{i,j}) \right] \\
& \text{where } \phi_t = \begin{cases}
    \phi_0, & \text{if } t < t_c \\
    \text{optimized}, & \text{if } t \geq t_c
\end{cases}
\end{aligned}
$}
\end{equation}
where $t_c$ is our delayed camera parameters optimization start step, $B \sim \mathcal{U}(I, b)$ indicates a batch of $B$ rays are used in total, and has uniform $b$ samples from each image. $\mathcal{L}_{\text{photo}}$ is the photometric loss between the rendered images and limited visible ground truth pixels. In our joint optimization framework, it is primarily supervised by $\mathcal{L}_{\text{MSE}}$. The specific weighting of this and other losses is detailed in Sec.~\ref{sec:experimental_setup}.

\subsection{Occlusion Annealing Regularization (OAR)}
\label{methodann}

In reconstruction after occlusion removal, the most significant issue arises from the variability of the visible region due to the non-fixed distribution of the occlusion. This results in two effects: 1) some areas being underfitted because they are not adequately visible across multiple views,  and 2) other non-occluded regions may be overfitted due to being fully visible in every input image. 
This imbalance leads to underfitting in regions that are sparsely visible and overfitting in consistently visible ones, often causing rendering artifacts within the same view.
Our objective is to reduce the impact of overfitting on the final rendering effect while ensuring sufficient generalization across views. Given that neural networks tend to learn low-frequency features~\cite{rahaman2019spectral, mildenhall2020nerf}, and inspired by regularization work in neural rendering across different frequencies~\cite{yang2023freenerf, lin2021barf, park2021nerfies}, we first use lower-dimensional pose embeddings to learn consistent low-frequency scene features from various perspectives, then gradually increase the dimension to standard frequency encoding. This simple approach leads to blurred boundaries where occlusions exist. Additionally, it delays the training convergence in areas with higher visibility frequencies, contributing to the generation of consistent rendering effects.

Due to limited multi-view supervision, artifacts in occluded regions frequently manifest as floaters near the camera, where rendering is more sensitive to density misestimation.
While prior work~\cite{yang2023freenerf} penalizes near-camera rays to suppress floaters, we find that early-stage low-frequency training can cause unstable feature clustering in these regions, making global penalties detrimental and prone to collapse.
To mitigate this, we propose Occlusion Annealing Regularization (OAR), which gradually introduces occlusion loss during frequency ramp-up, stabilizing training under view redundancy.

The position and direction encodings at iteration $t$ are represented as:
\begin{equation}
\begin{aligned}
\mathbf{e}_{\text{pos}}(t) &= \mathbf{x} \odot \mathbf{m}_{\text{pos}}(t), \\
\mathbf{e}_{\text{dir}}(t) &= \mathbf{d} \odot \mathbf{m}_{\text{dir}}(t),
\end{aligned}
\end{equation}
where $\mathbf{x}$ and $\mathbf{d}$ are the original position and direction encodings, and $\mathbf{m}_{\text{pos}}(t)$ and $\mathbf{m}_{\text{dir}}(t)$ are frequency masks that depend on the current iteration $t$. 

The masks are defined as:
\begin{equation}
\mathbf{m}_{\text{pos,dir}}(t) = 
\begin{cases}
1, & \text{if } f \leq f_{\text{max}}(t), \\
0, & \text{otherwise},
\end{cases}
\end{equation}
where $f$ is the frequency of each encoding dimension, and $f_{\text{max}}(t)$ is the maximum allowed frequency at iteration $t$, which increases linearly from 0 to the maximum frequency over the course of training. Through the masks, low-frequency and high-frequency information is progressively exposed to the network.

This progressive exposure is synchronized with an annealed occlusion loss weight, defined via a cosine schedule between iterations $t_{\text{start}}$ and $t_{\text{end}}$, ensuring a smooth transition toward full supervision as frequency increases:
\begin{equation}
w_{\text{occ}}(t) = 
\begin{cases}
0, & \text{if } t < t_{\text{start}}, \\
\frac{w_{\text{full}}}{2} \left(1 + \cos\left(\pi \frac{t_{\text{end}} - t}{t_{\text{end}} - t_{\text{start}}}\right)\right), & \text{if }  t_{\text{start}} < t < t_{\text{end}}, \\
w_{\text{full}}, & \text{if } t \geq t_{\text{end}}.
\end{cases}
\end{equation}
Here, $w_{\text{full}}$ is the full occlusion loss weight, $t_{\text{start}}$ is the iteration to start introducing the occlusion loss, and corresponding $t_{\text{end}}$ is the iteration when the full weight is reached.

The occlusion loss is then calculated as:
\begin{equation}
    \mathcal{L}_{\text{occ}}(t) = w_{\text{occ}}(t) \cdot \mathcal{L}_{\text{occ\_base}}
\end{equation}
\begin{equation}
\mathcal{L}_{occ\_base} = \frac{\boldsymbol{\sigma}_K^{\mathbf{T}} \cdot \mathbf{m}_K}{K} = \frac{1}{K} \sum_K \sigma_k \cdot m_k,
\end{equation}
where $\mathbf{m}_k$ is a binary mask vector and $\boldsymbol{\sigma}_K$ denotes the density values of the $K$ sampled points along the ray.
The frequency regularization end ($t_{\text{freq\_end}}$) and occlusion annealing are connected through $\lambda$:
\begin{equation}
    t_{\text{end}} = \frac{t_{\text{freq\_end}}}{\lambda}
\end{equation}

This coordination between frequency and occlusion schedules promotes stable learning early on while effectively penalizing artifacts later in training.

\begin{figure}[t]
    \centering
    \includegraphics[width=0.9\linewidth]{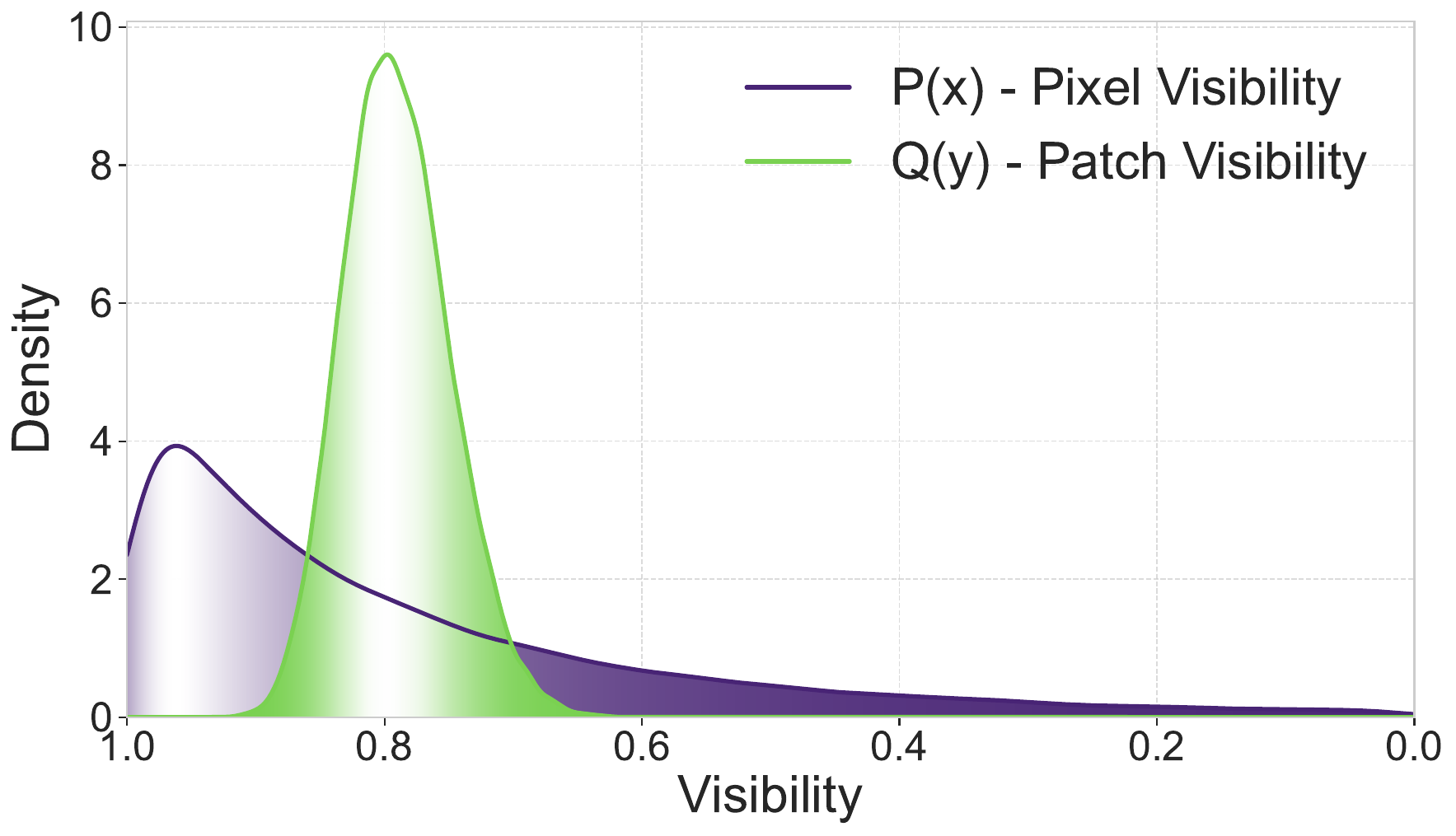}
    \caption{\textbf{Visualization of Sampling Distribution.} For a demonstration of the principle of global patched S3IM, the distribution of pixels exhibits a marked imbalance. This issue can be addressed through our patch reorganization. The distribution of each patch becomes more concentrated and uniform, eliminating the regional long-tail distribution of pixels and promoting stable model iteration. Darker regions indicate more extreme long-tailed visibility, which require targeted optimization.}
    \label{fig:vis_dis}
    \vspace{-4mm}
\end{figure}

\subsection{S3IM in Occluded Long-Tailed Visibility}
\label{methodssim}

Occluded scenes often exhibit a long-tailed distribution of pixel visibility, \ie, most pixels appear frequently across views, while a minority are rarely observed. This imbalance hampers stable training and leads to biased reconstructions, as frequently visible pixels dominate the optimization signal.
We employ a patch-based global stochastic structural similarity method~\cite{xie2023s3im} to address this issue and validate its effectiveness in our task. As described in Sec.~\ref{sec:preliminaries} with all the notations, S3IM lies within $[-1, 1]$ and is positively correlated with image quality, so its loss definition is:
\vspace{-3mm}

\begin{equation}
\begin{aligned}
\mathcal{L}_{S3IM}(\Theta, \mathcal{R}) &= 1 - S3IM(\hat{\mathcal{R}}, \mathcal{R}) \\
&= 1 - \frac{1}{M} \sum_{m=1}^M SSIM(\mathcal{P}^{(m)}(\hat{C}), \mathcal{P}^{(m)}(C)).
\end{aligned}
\end{equation}

This involves a patch-based stochastic structural similarity SSIM~\cite{wang2004image} but in a global range. To better understand how this loss formulation mitigates long-tailed visibility issues, we begin by characterizing the underlying pixel visibility distribution. In the illustration for occluded scenes, pixel visibility follows a long-tailed pattern:

\begin{equation}
    P(x) \approx \frac{1}{(x_{max} - x + 1)^\alpha}, \quad \alpha > 1
\end{equation}
where $x$ represents pixel visibility, the number of views in which a pixel is visible, and $x_{max}$ corresponds to the maximum visibility. S3IM randomly samples rays across all views and group them into $K \times K$ patches. Each patch’s visibility is defined as:
\vspace{-2mm}

\begin{equation}
    \text{vis}(p) = \frac{1}{K^2} \sum_{i=1}^{K^2} \text{vis}(pixel_i)
\end{equation}
which converts the per-pixel visibility distribution $P(x)$ into a patch-level distribution $Q(y)$, where
\vspace{-3mm}

\begin{equation}
    Q(y) = P(y = \frac{1}{K^2} \sum_{i=1}^{K^2} X_i), \quad X_i \sim P(x)
\end{equation}

\noindent Compared with $x$ and $x_{max}$, $y$ is the average visibility of a patch. By aggregating over both high and low visibility pixels, each patch visibility becomes naturally moderated:
\vspace{-3mm}

\begin{equation}
    \min_{i \in p} \text{vis}(pixel_i) \leq \text{vis}(p) \leq \max_{i \in p} \text{vis}(pixel_i)
\end{equation}

This mixing effect shortens the tail of the visibility distribution (as visualized in Fig.~\ref{fig:vis_dis}), resulting in more centralized gradients during optimization. By balancing supervision across visibility levels, it enhances stability and improves reconstruction in sparsely observed regions.

\begin{figure}[t]
    \centering
    \includegraphics[trim={0.5cm 0.5cm 0.5cm 0.5cm},width=\linewidth]{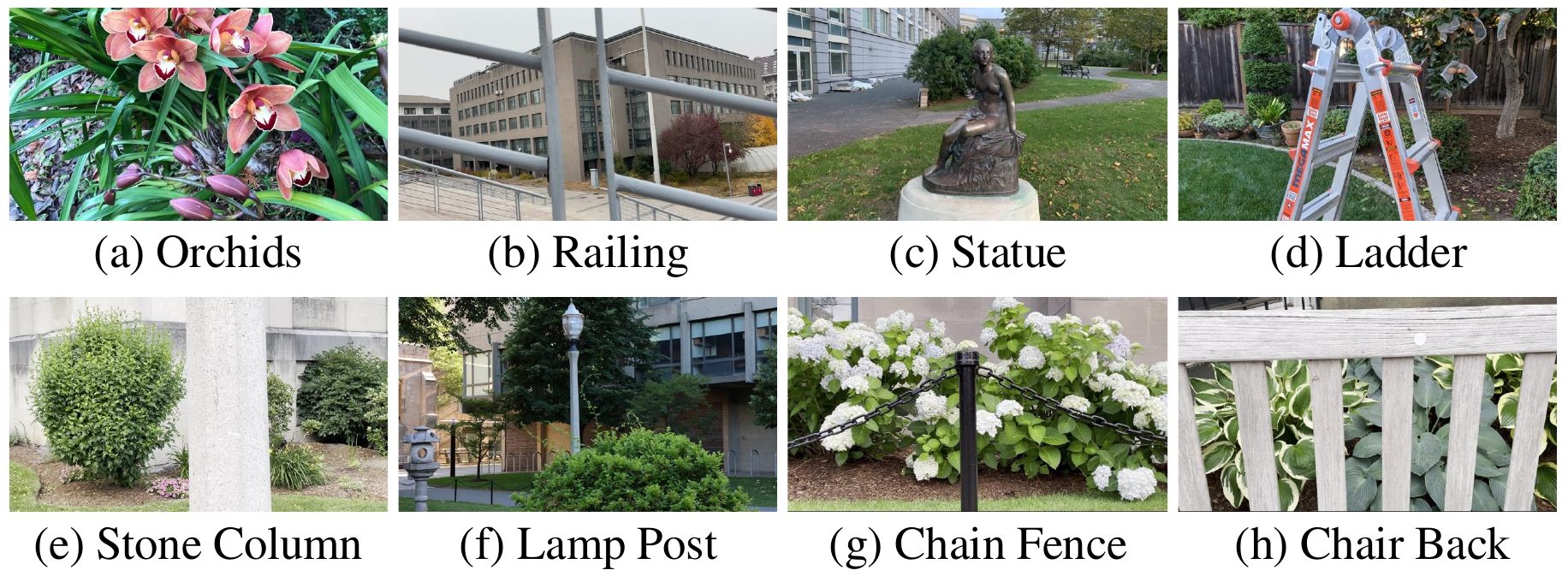}
    \caption{\textbf{The DeclutterSet.} \texttt{(a)Orchids} and \texttt{(f)Lamp Post} illustrate occluders at different distances: in (a), both the buds and flowers lie on the same near-depth plane close to the camera, while in (f), the occluding object is situated farther away in the mid-background; \texttt{(b)Railing} and \texttt{(c)Statue} resemble traditional occlusion and object removal settings commonly found in existing benchmarks; \texttt{(e)Stone Column} and \texttt{(g)Chain Fence} exhibit occlusions that scatter across different image regions as the viewpoint shifts; \texttt{(d)Ladder} and \texttt{(h)Chair Back} feature larger, irregularly shaped occluders and more pronounced viewpoint variations, posing further challenges to cross-view consistency and geometry recovery. Further details are provided in the supplementary material.}
    \vspace{-4mm}
    \label{fig:data_fig}
\end{figure}

\section{Experiments}
\label{sec:experiments}

\begin{table*}[t]
\centering
\caption{\textbf{Quantitative Comparisons With the Generative-Free Baseline.} Due to the lack of evaluation code or the inability to reconstruct all scenes, generative baselines are excluded. OCC-NeRF is selected as the only non-generative method that supports reconstruction on DeclutterSet. Under its original parameter settings, OCC-NeRF underperforms due to its rigid near-range removal strategy and lack of post-removal refinement. In contrast, DeclutterNeRF yields superior performance across all metrics and scenes, demonstrating improved robustness to complex scenarios. On average, our method improves PSNR by 68.4\%, SSIM by 238.0\%, and reduces LPIPS by 54.8\%.}
\vspace{-2mm}
\label{tab:quantitative_eva}
\begin{adjustbox}{max width=0.99\textwidth}
\begin{tabular}{@{}lccccccccc@{}}
\toprule
& \multicolumn{2}{c}{PSNR↑} & \multicolumn{2}{c}{SSIM↑} & \multicolumn{2}{c}{LPIPS↓} \\
\cmidrule(lr){2-3} \cmidrule(lr){4-5} \cmidrule(lr){6-7}

Scene & OCC-NeRF & Ours & OCC-NeRF & Ours & OCC-NeRF & Ours \\
\midrule
(a) Orchids & 10.76 & 20.86 & 0.213 & 0.894 & 0.371 & 0.130 \\
(b) Railing & 14.00 & 23.12 & 0.324 & 0.860 & 0.457 & 0.241 \\
(c) Statue & 16.13 & 24.87 & 0.197 & 0.902 & 0.502 & 0.135 \\
(d) Ladder & 13.17 & 21.37 & 0.074 & 0.656 & 0.534 & 0.352 \\
(e) Stone Column & 14.73 & 21.73 & 0.240 & 0.864 & 0.435 & 0.229 \\
(f) Lamp Post & 15.62 & 22.67 & 0.581 & 0.903 & 0.403 & 0.240 \\
(g) Chain Fence & 11.90 & 23.71 & 0.294 & 0.927 & 0.389 & 0.125 \\
(h) Chair Back & 10.59 & 21.67 & 0.118 & 0.887 & 0.432 & 0.143 \\
\midrule
Average & 13.36 & 22.50 & 0.255 & 0.862 & 0.440 & 0.199 \\
\bottomrule
\end{tabular}
\end{adjustbox}
\end{table*}

\begin{figure*}[t]
    \centering
    \includegraphics[trim={1cm 1.3cm 1cm 1.1cm},width=0.98\linewidth]
    {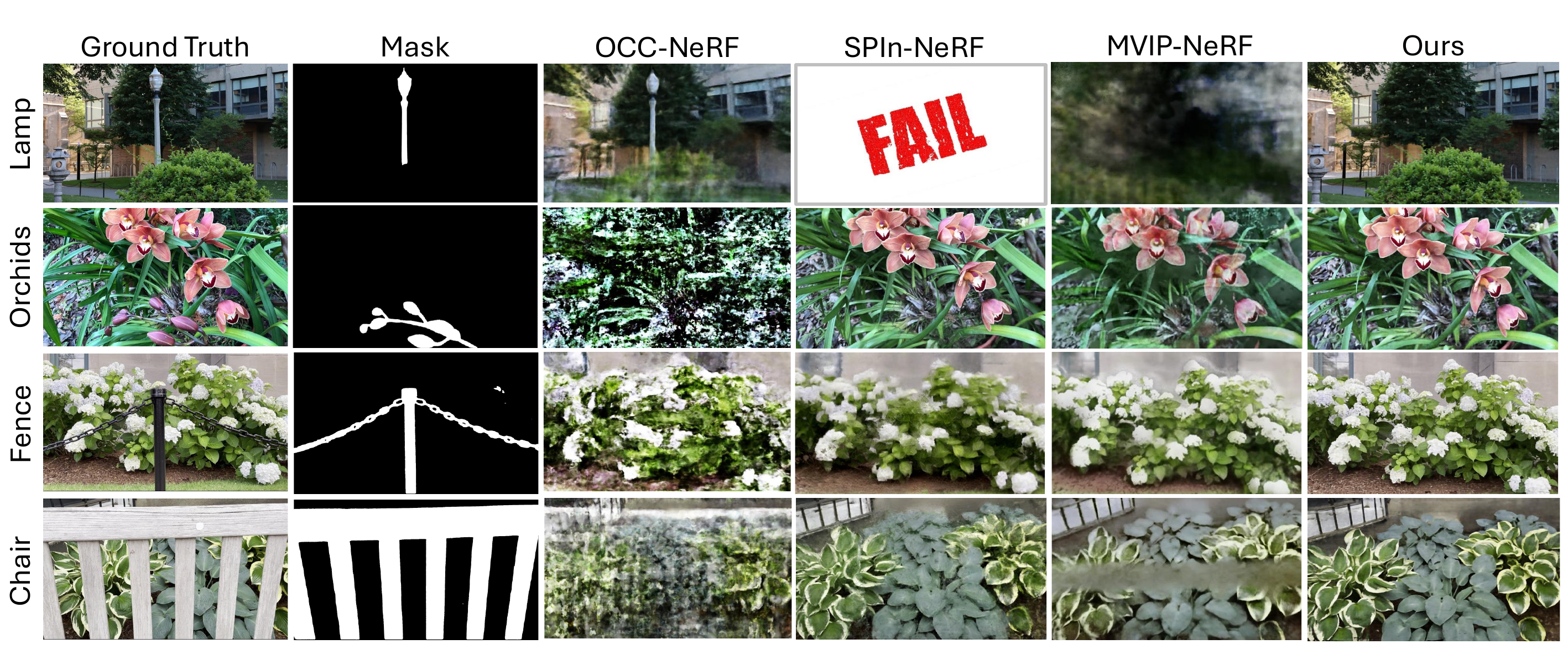}
    \caption{\textbf{Qualitative Comparisons With Baselines.} We compare DeclutterNeRF against OCC-NeRF, SPIn-NeRF and MVIP-NeRF across both standard and newly collected datasets. Our method reliably removes occluders at varying depths and achieves photorealistic reconstruction. In contrast, OCC-NeRF struggles in close-range scenes due to its distant-only rendering assumptions. SPIn-NeRF and MVIP-NeRF, designed on previous benchmarks, frequently suffer from inconsistent floaters and hallucinated artifacts when occluders shift their relative positions across views---a mode exposed by our DeclutterSet but overlooked in prior benchmarks.}
    \vspace{-4mm}
    \label{fig:res_fig}
\end{figure*}

\subsection{Experimental Setup}
\label{sec:experimental_setup}

\textbf{Dataset.} Due to the novelty of the occlusion removal and reconstruction problem and the limited availability of existing datasets, we follow the pattern of NeRF, which uses 8 scenes from the LLFF dataset~\cite{mildenhall2019llff}, and create a dataset comprising 8 occluded scenes. As shown in Fig.~\ref{fig:data_fig}, DeclutterSet comprises eight occluded scenes, including four sourced from existing benchmarks and four newly captured, ensuring a balanced distribution of occlusion types and scene layouts. Specifically, (a) Orchids is taken from the classic LLFF dataset, (b) Railing is from the OCC-NeRF dataset, and (c) Statue and (d) Ladder are from dataset IBRNet~\cite{wang2021ibrnet} which currently has become the mainstream data in object removal. For the data we constructed, (e) to (h), each scene consists of approximately 30 images captured using a Canon R6 Mark II or an iPhone 12 Pro. Following the mainstream approach, we created the test set by holding out $1/8$ of the images. The details for mask annotation and propagation are described in the supplementary material.

\noindent\textbf{Baselines \& Metrics.} We compare our method against both generative and non-generative state-of-the-art approaches for object and occlusion removal in NVS. Specifically, OCC-NeRF~\cite{zhu2023occlusion} serves as the generative-free baseline, while SPIn-NeRF~\cite{spinnerf} and MVIP-NeRF~\cite{MVIPNeRF} represent the generative baselines. We provide both qualitative and quantitative evaluations for the rendering results. For qualitative analysis, we include visualizations across all scenes in the main paper and the supplementary material. For quantitative evaluation, we report standard NeRF reconstruction metrics, PSNR, SSIM, and LPIPS, computed over non-occluded pixels only.

\noindent\textbf{Parameters.} Similar to most learning-based 3D reconstruction methods, DeclutterNeRF is also influenced by hyperparameters. We set the termination of frequency regularization at 10\% of the total iterations, begin camera parameter optimization at 20\% of the total iterations, and set the Occlusion Annealing Regularization $\lambda$ to 100. Our $\mathcal{L}_{photo}$ consists of three mainstream loss functions: $\mathcal{L}_{mse}$, $\mathcal{L}_{occ}$, and $\mathcal{L}_{s3im}$. The weights of $L_\text{occ}$ and $L_\text{s3im}$ are set to 0.01. More details can be found in the supplementary material.

\subsection{Comparison Results}

\textbf{Qualitative Evaluation.} Figure~\ref{fig:res_fig} compares our method's rendering results with OCC-NeRF, SPIn-NeRF, and MVIP-NeRF. The mask represents the occluding objects we aim to remove. Our pipeline enables selective removal of occlusions with minimal manual intervention, including the distant lamp and the unopened orchid buds in the foreground, while achieving photorealistic reconstructions. In contrast, OCC-NeRF only handles distant scenes adequately, often removing desired nearby objects and failing to reconstruct close-range details. For datasets like LLFF, which primarily consist of close-range scenes, OCC-NeRF's performance is significantly limited. Even for distant scenes, OCC-NeRF's depth warping strategy, impedes the optimization process, causing the model to struggle with complex geometries and leading to poor reconstruction quality that often appears smeared. Our joint camera parameter optimization strategy effectively avoids local minima traps and leverages this optimization to achieve high quality reconstructions.

The performance of SPIn-NeRF and MVIP-NeRF on our DeclutterSet also shows notable differences compared to our model. SPIn-NeRF's heavy reliance on COLMAP and pre-rendered depth priors impacts its reconstruction capability, often leading to failure rendering when depth information cannot be accurately recovered, particularly in distant scenes. More failure cases, parameter settings, and detailed analysis can be found in the supplementary material. Even MVIP-NeRF, despite claiming independence from depth priors, struggles with outdoor distant scenes. Moreover, in relatively simple scenes, these generative methods are highly prone to overfitting. As training progresses, reconstruction quality plateaus while artifacts increase, degrading overall results. Our method effectively avoids overfitting and underfitting issues caused by varying exposure levels in occluded regions, and it suppresses artifact generation, achieving optimal reconstruction results.

\noindent\textbf{Quantitative Evaluation.} We quantitatively evaluate our method against OCC-NeRF, the only generative-free baseline capable of handling all scenes in DeclutterSet. As shown in Table~\ref{tab:quantitative_eva}, DeclutterNeRF achieves consistent and significant improvements across all standard NeRF metrics. 
The performance gap is especially pronounced in challenging scenes such as Orchids, Chain Fence, and Chair Back, where OCC-NeRF’s fixed near-range removal strategy struggles to adapt to varying occlusion depths and scene complexity. This is particularly evident in the Orchids scene, as corroborated by Fig.~\ref{fig:res_fig}, where none of the flowers are retained in OCC-NeRF’s output---resulting in one of the lowest PSNR scores. In contrast, our method contributes to more robust, artifact-free reconstruction under diverse occlusion settings. 

Beyond accuracy, DeclutterNeRF also demonstrates high practical efficiency, enabled by our multi-stage architectural improvements. It completes training in under 10 hours on a single NVIDIA RTX 4090 GPU, with memory consumption kept below 10 GB. In comparison, OCC-NeRF requires over 30 hours of training, while the diffusion and distillation-based learning MVIP-NeRF demands more than 100 GB of GPU memory. This level of efficiency makes DeclutterNeRF more accessible and better suited for broader adoption and large-scale experimentation.

\subsection{Ablation Studies}

\begin{figure}[t]
    \centering
    \begin{subfigure}[b]{0.23\textwidth}
        \centering
        \includegraphics[width=\textwidth]{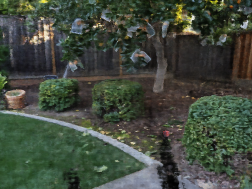}
        \caption{}
        \label{fig:subfig1}
    \end{subfigure}
    \hfill
    \begin{subfigure}[b]{0.23\textwidth}
        \centering
        \includegraphics[width=\textwidth]{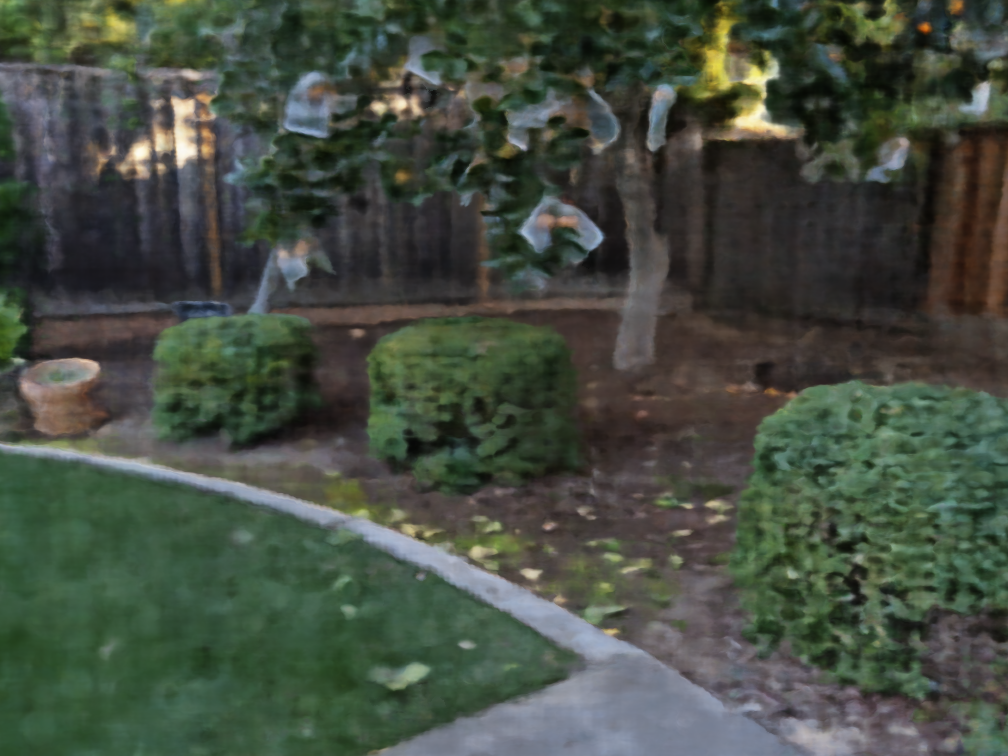}
        \caption{}
        \label{fig:subfig2}
    \end{subfigure}
    \caption{\textbf{Visual Ablation Studies for Occlusion Annealing Regularization (OAR).} Visual effects between introducing (iii) OAR on the Ladder scene: (a) before, (b) after.}
    \label{fig:ablation}
    \vspace{-4mm}
\end{figure}

We demonstrate the impact of ablation and the gradual introduction of each component in Table~\ref{tab_ablation}. Initially, we train (i) a masked NeRF, which simply uses the non-occluded mask areas for training. Subsequently, we introduce (ii) joint optimization for camera parameters, which improves all the rendering metrics. Notably, while the use of (iii) occlusion annealing regularization (OAR) results in a slight regression in rendering metrics, it addresses the main issues of artifacts and incomplete rendering. We show this in Fig.~\ref{fig:ablation}, which observably enhances the actual visual effect. Finally, we introduced the S3IM loss to further address the reconstruction costs caused by occlusions.

\begin{table}[t]
\centering
\caption{\textbf{Ablation Studies.} Metrics evaluation through ablation and gradual introduction of each module in our framework. Although the introduction of OAR leads to a slight drop in overall quantitative accuracy, it plays a critical role in cross-view generalization and is essential for successful reconstruction after occlusion removal. This effect is further illustrated in Fig~\ref{fig:ablation}.}
\begin{tabular}{lccc}
\toprule
Method & PSNR↑ & SSIM↑ & LPIPS↓ \\
\midrule
(i) (masked NeRF) & 19.59 & 0.56 & 0.38 \\
(ii) (+camera opt.) & 21.01 & \textbf{0.67} & \textbf{0.28} \\
(iii) (+OAR) & 20.89 & 0.61 & 0.29 \\
(iv) (+s3im loss) & \textbf{21.37} & 0.656  & 0.352 \\
\bottomrule
\end{tabular}
\label{tab_ablation}
\end{table}

\subsection{Limitations}

Our experiments assume that occluded regions are at least partially visible from other viewpoints, as we have no generative priors to reconstruct unseen parts of the scene. In cases where occlusions completely hide target content from all views, generative priors remain necessary. However, we believe that future generative approaches can build upon our framework---first maximizing reconstruction from observable data, then refining the remaining gaps through targeted generation. This layered strategy promises both efficiency and consistency for occlusion-aware scene recovery.
\section{Conclusion}
\label{sec:conclusion}

We introduced DeclutterSet, a dataset designed to reflect the real-world complexity of occlusions with diverse object layouts and viewpoint variations, addressing critical limitations of existing benchmarks. Based on this, we proposed DeclutterNeRF, a generative-free framework that leverages NeRF’s multi-view consistency, joint camera optimization, occlusion annealing regularization, and stochastic structural similarity loss. Our method achieves state-of-the-art performance on this specific task with minimal computational overhead. We hope this work offers a broader perspective on occlusion and object removal and serves as a foundation for future research, whether generative or optimization-based, in robust and efficient 3D scene reconstruction.
{
    \small
    \bibliographystyle{ieeenat_fullname}
    \bibliography{main}
}

\appendix

\clearpage
\setcounter{page}{1}
\maketitlesupplementary

\section{Method Details}
\label{sec:method_details}

\subsection{Architecture and Training Details}

DeclutterNeRF follows the core architecture and strategy of the original NeRF~\cite{mildenhall2020nerf}. Specifically, we build on NeRF\texttt{-{}-}~\cite{wang2021nerf} and apply DeclutterNeRF on top of this structure. Our model is implemented using PyTorch~\cite{Paszke2019PyTorchAI} and trained on a single NVIDIA GeForce RTX 4090 GPU. Since our dataset typically requires no more than 10GB of VRAM, and the image size can be adjusted flexibly to control memory usage, GPUs with significantly lower configurations can also be used to train our model. Unlike recent models that employ multiple MLPs and assign distinct names to each, we adhere to the original NeRF approach by using a single MLP for training and rendering. Our MLP consists of 8 fully connected ReLU hidden layers, each with 128 dimensions. Our further camera optimization algorithm mentioned in \textit{Sec.~3.2} and problems encountered based on the logic of NeRF\texttt{-{}-}.

For training settings, we use a scale factor of 4 and a batch size of 4096, with 200K iterations. This aligns with the training methods of current mainstream models. Even with a scale factor of 2 and a batch size of 8192, our GPU memory usage does not exceed 15 GB. We evenly distribute the batch samples across each input image, so the number of samples per image depends on the total number of images in this scene. We train our model using the Adam optimizer~\cite{Kingma2014AdamAM}.

\begin{figure*}[t]
    \centering
    \includegraphics[trim={0cm 0.6cm 0cm 0cm},width=\linewidth]{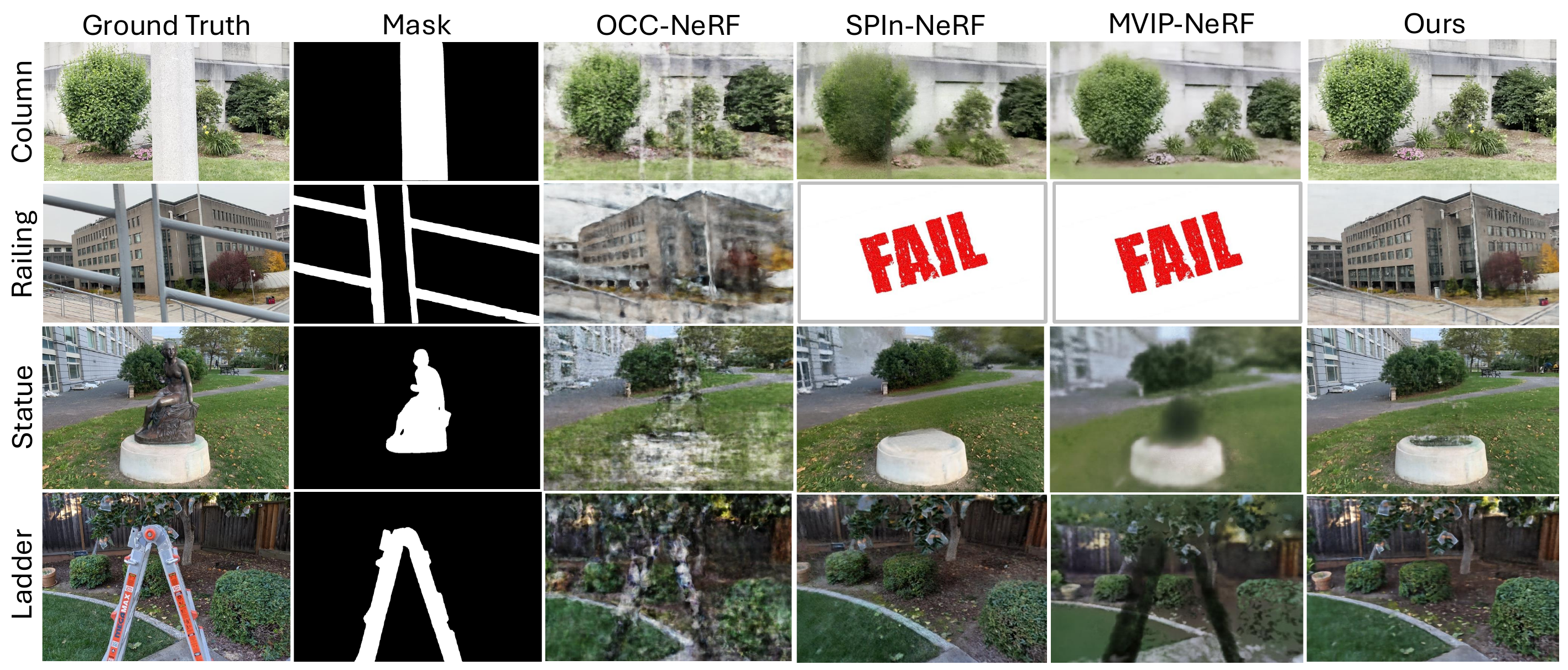}
    \caption{\textbf{Additional Qualitative Comparisons With Baselines.} Our method consistently produces desirable results, while generative models still suffer from artifacts and floaters during rendering. Notably, DeclutterNeRF maintains geometric fidelity and cross-view consistency in challenging occlusion scenarios with complex depth relationships. A detailed analysis of failure cases is provided in Sec.~\ref{sec:add_qua_res}.}
    \label{fig:res_fig_spl}
\end{figure*}

\subsection{Annotation Mapping Details}

We directly leverage OR-NeRF's efficient multiview segmentation approach to remove obstacles and construct our dataset~\cite{yin2023ornerf}. Its multiview segmentation process is both efficient and consistent. When given point prompts on a single view, the system projects these points into 3D space using COLMAP's sparse reconstruction, establishing correspondences between 2D points and the 3D point cloud. These 3D points are then projected back to all 2D images using camera parameters, creating consistent annotations across all views. Once annotations are propagated to all views, the SAM predicts masks for each view at approximately two frames per second, without requiring neural network training for each scene.

\subsection{Evaluation Settings}

Due to the irregular occlusion masks in occluded images, we rearrange valid pixels from ground truth and rendered images into rectangular formats suitable for SSIM and LPIPS patch-based evaluation. This rearrangement may introduce slight variations in metrics compared to methods that directly compare original images, as the structural changes can affect SSIM and LPIPS scores. However, these differences are typically minimal and do not impact the overall evaluation results.

Considering the unavoidable occlusions when capturing real-world scenes, we calculate the rendering accuracy only within the valid visible regions using masks. Therefore, we suggest readers interpret the quantitative evaluation metrics reasonably and place more emphasis on the qualitative results, which demonstrate the true rendering performance in scenes with occlusion removal.

\section{Dataset Details}

\subsection{Dataset Building Process}

For the DeclutterSet, we capture each scene using either a Canon R6 Mark II camera or an iPhone 12 Pro, maintaining consistent exposure and focus settings throughout the capture process. To ensure high-quality multi-view inputs, we record continuous video while moving the camera in a smooth arc trajectory around the scene. From each recording, we extract 30-35 sequential frames at regular intervals, creating a forward-facing dataset similar to the classic NeRF format. We pay attention to select scenes with varying occlusion characteristics - different depths, scales, and geometric complexity. Camera parameters are estimated using COLMAP's structure-from-motion pipeline. For occlusion annotation, we used OR-NeRF's efficient multiview segmentation approach, requiring only point prompts on a single view to generate consistent masks across all views.

\subsection{Considerations}
\label{sec:dataset_details} 

While OCC-NeRF~\cite{zhu2023occlusion} provides some occlusion datasets, community feedback (as evidenced by multiple issues raised in its repository) has identified several issues with their data. These include blurry images, missing parameters, and even mismatches in ground truth for testing. Even the authors' model and code failed to reproduce their reported results.

To address these shortcomings, we constructed \textit{DeclutterSet}, which includes a variety of occlusion types, varying occlusion sizes and camera motions, and different occluder distances. As stated in the main text, it combines reliable data from existing references and is augmented with newly captured scenes, offering a new and robust benchmark for the community.

\subsection{Samples Exhibition}

\begin{figure*}[t]
    \centering
    \begin{tabular}{@{}c@{\hspace{0.5mm}}c@{\hspace{0.5mm}}c@{\hspace{0.5mm}}c@{}}
        \includegraphics[width=0.22\textwidth]{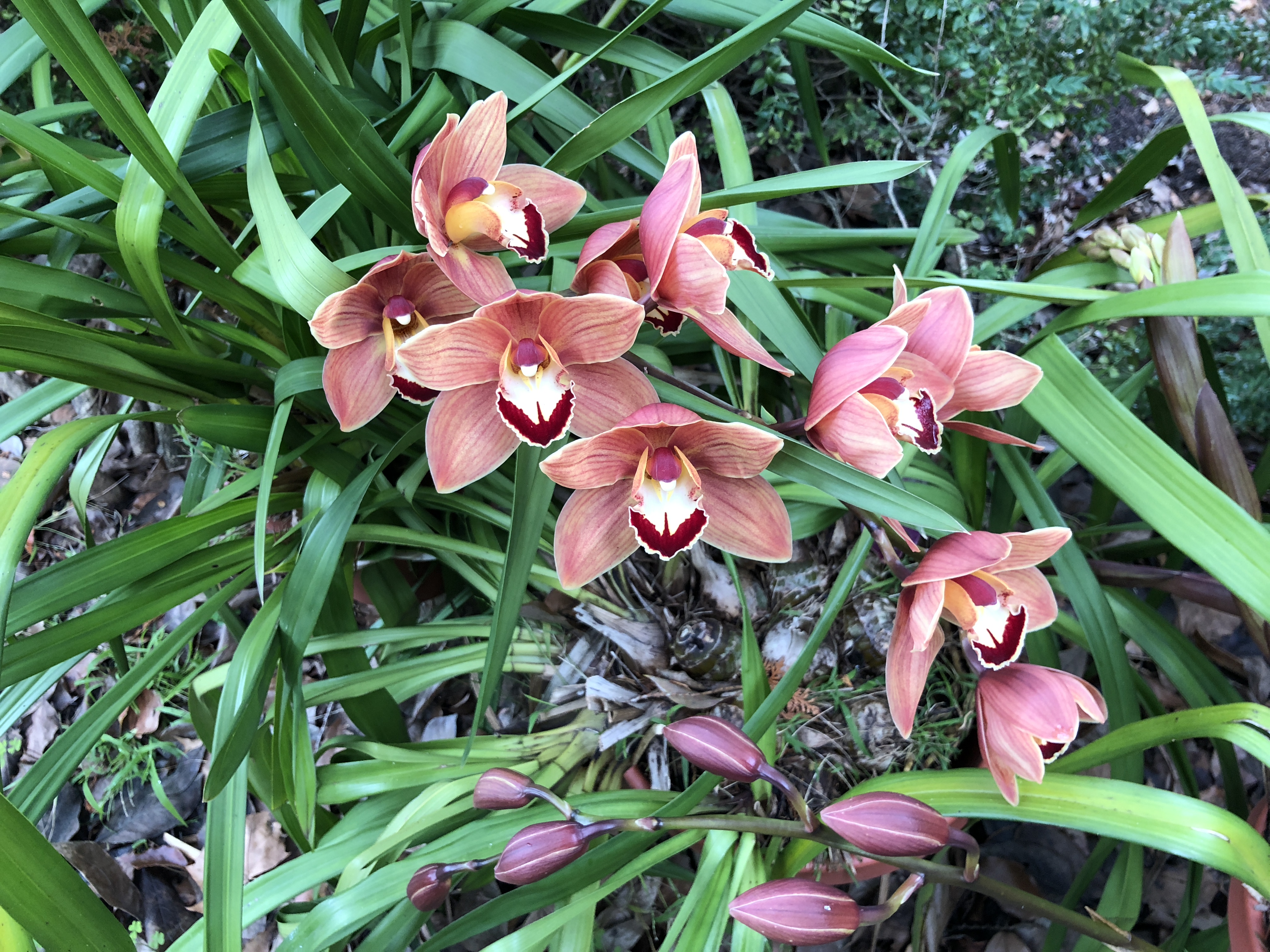} &
        \includegraphics[width=0.22\textwidth]{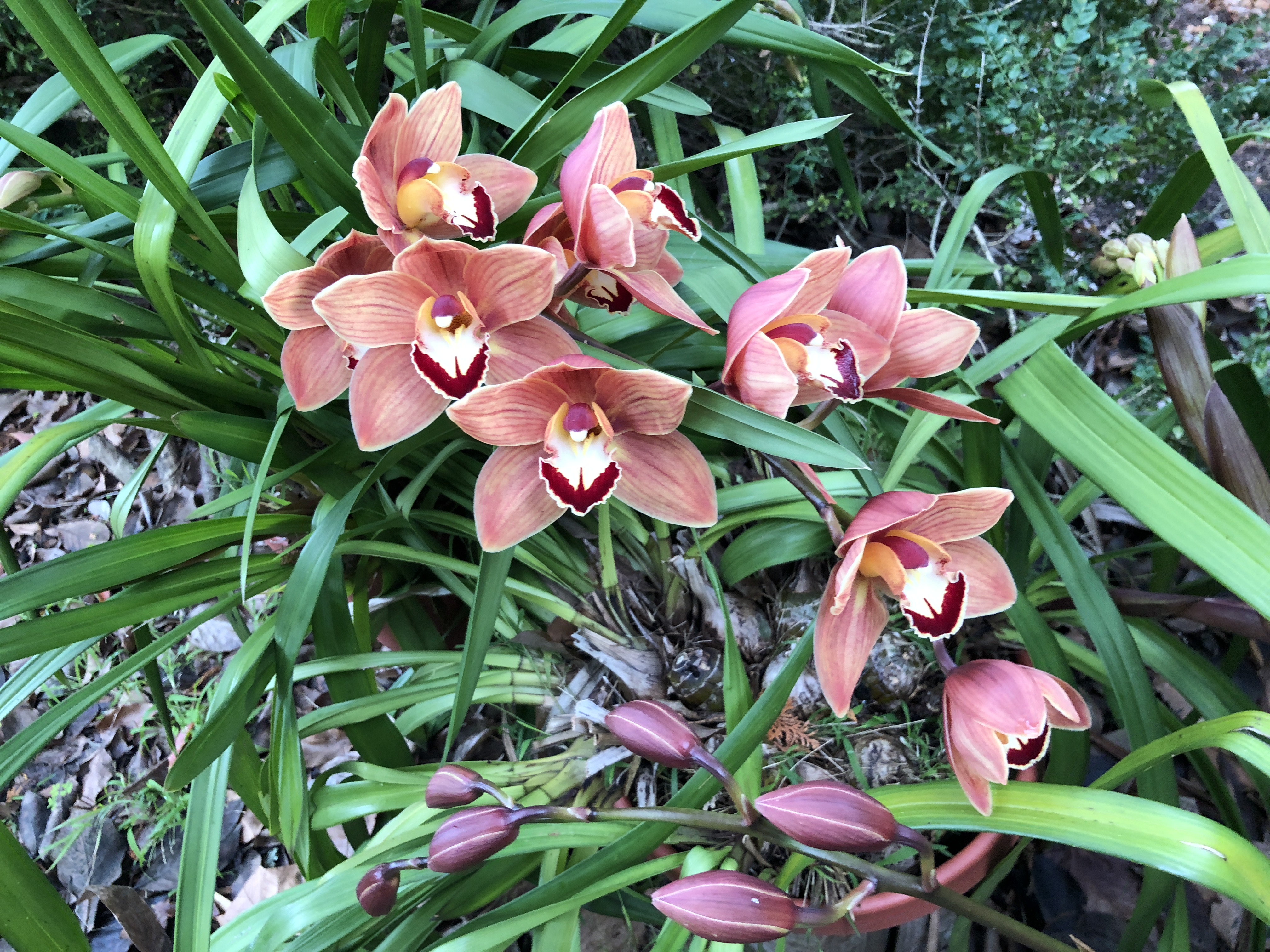} &
        \includegraphics[width=0.22\textwidth]{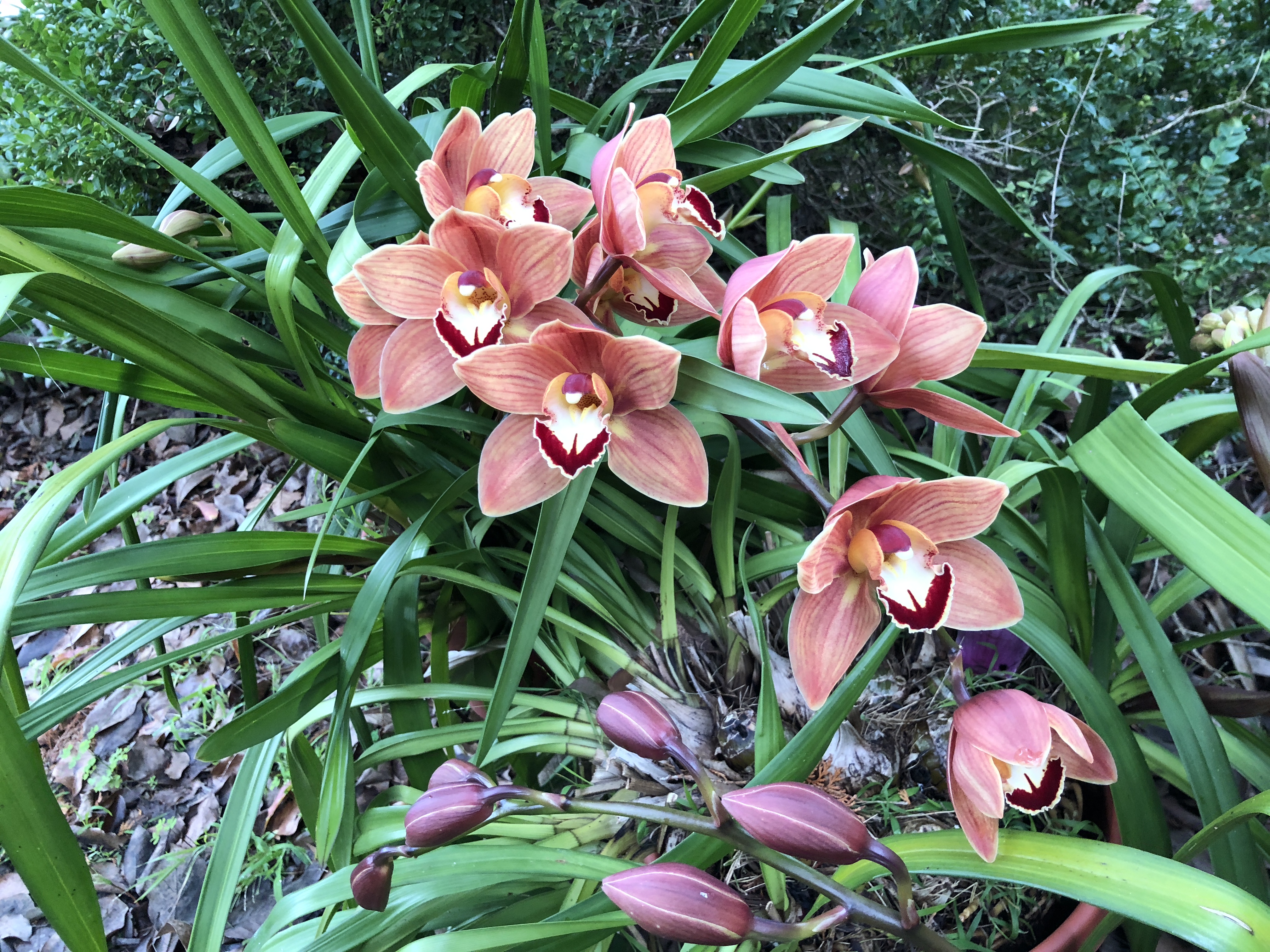} &
        \includegraphics[width=0.22\textwidth]{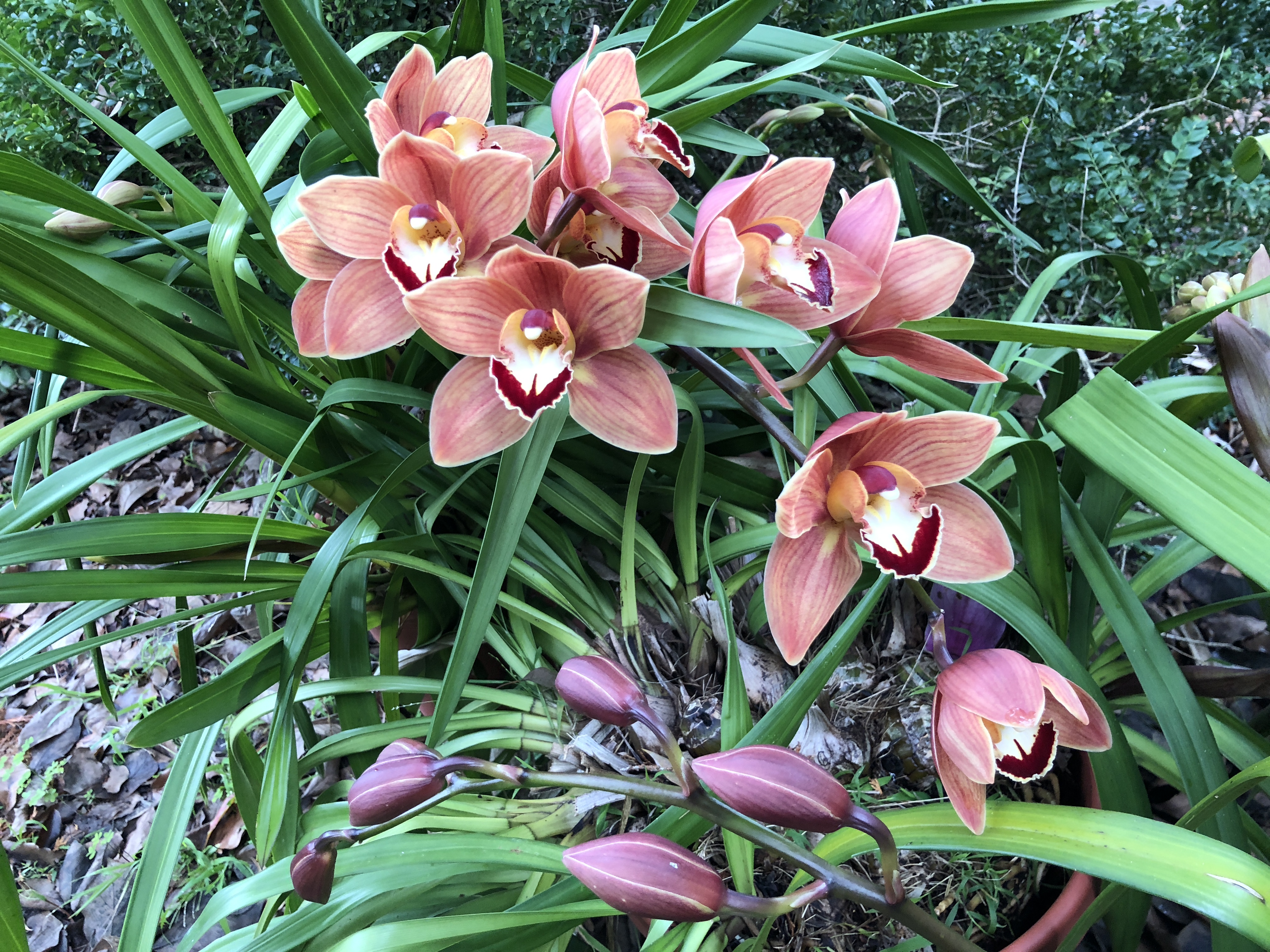} \\ [-0.7mm]
        \includegraphics[width=0.22\textwidth]{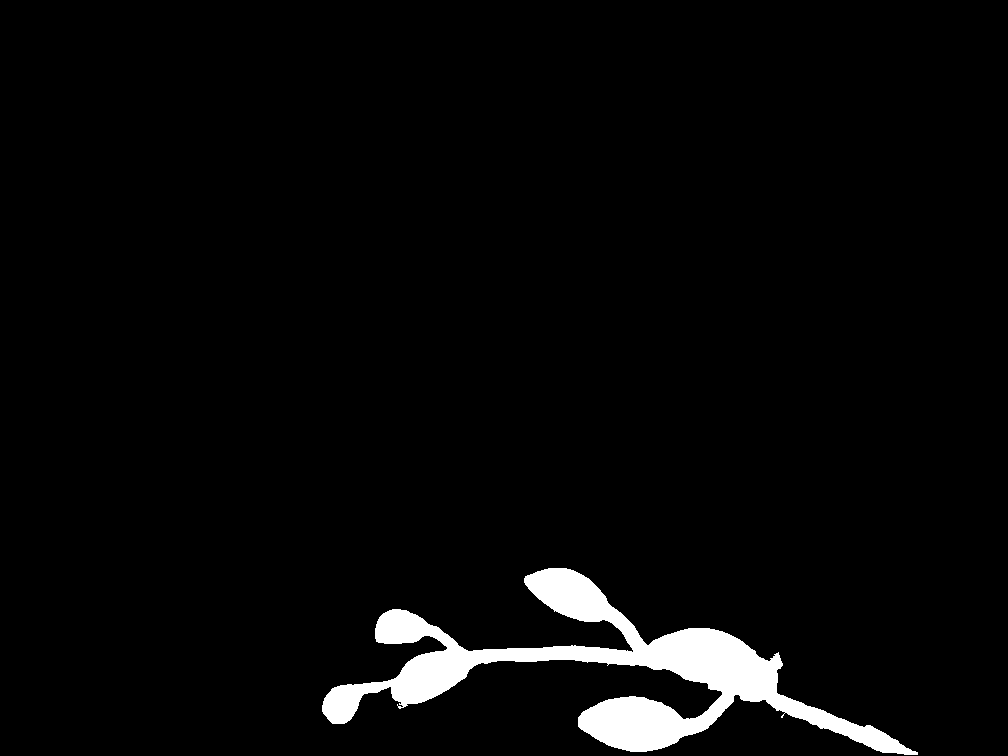} &
        \includegraphics[width=0.22\textwidth]{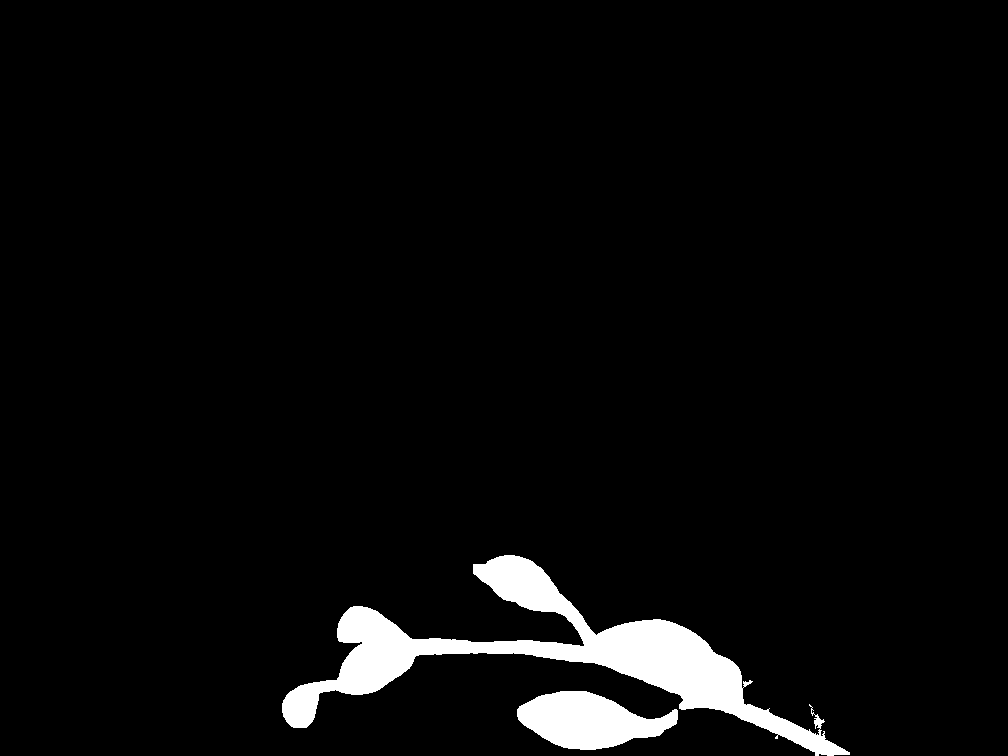} &
        \includegraphics[width=0.22\textwidth]{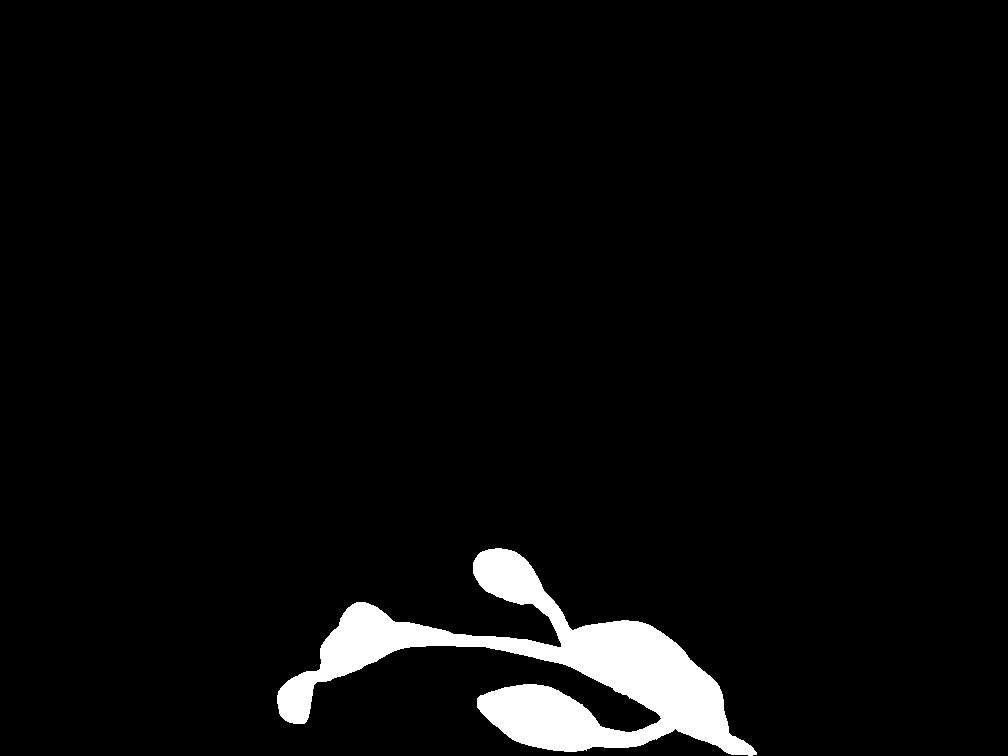} &
        \includegraphics[width=0.22\textwidth]{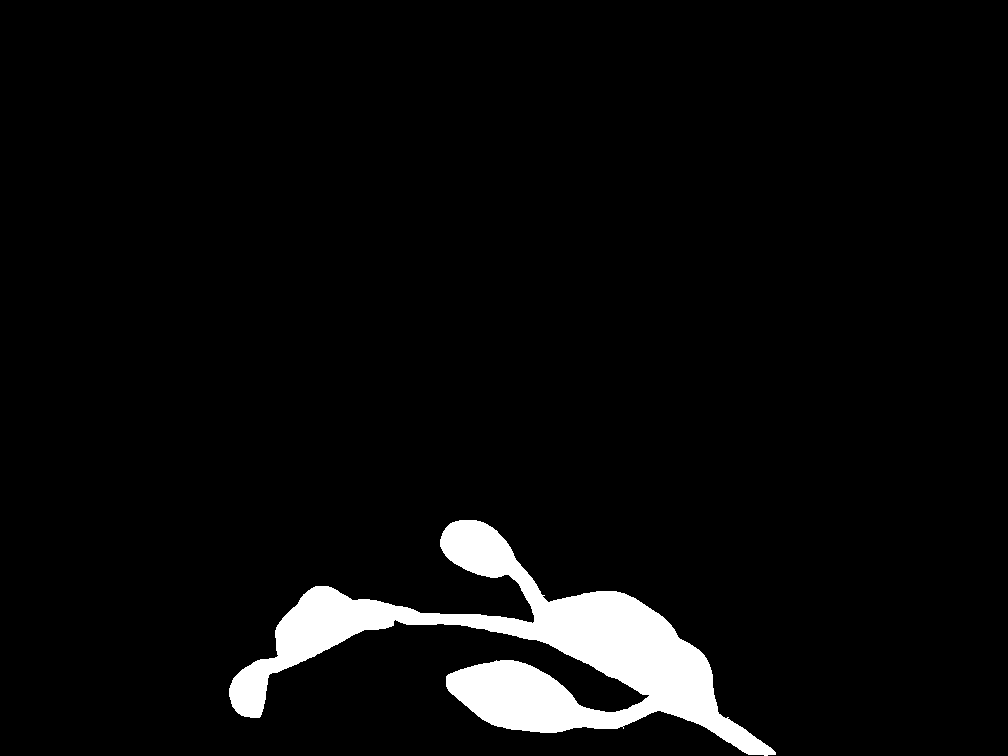} \\ [-0.7mm]
        \includegraphics[width=0.22\textwidth]{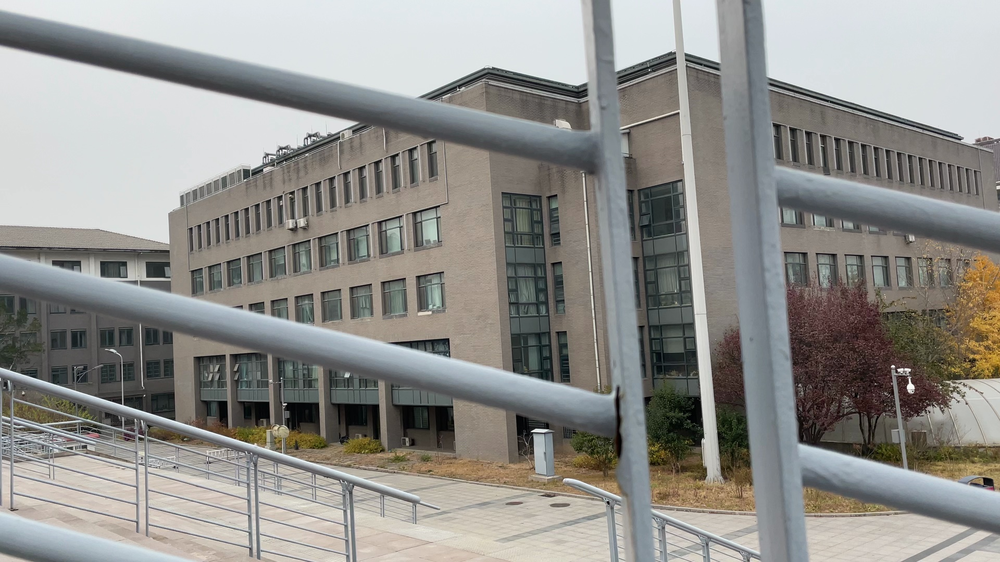} &
        \includegraphics[width=0.22\textwidth]{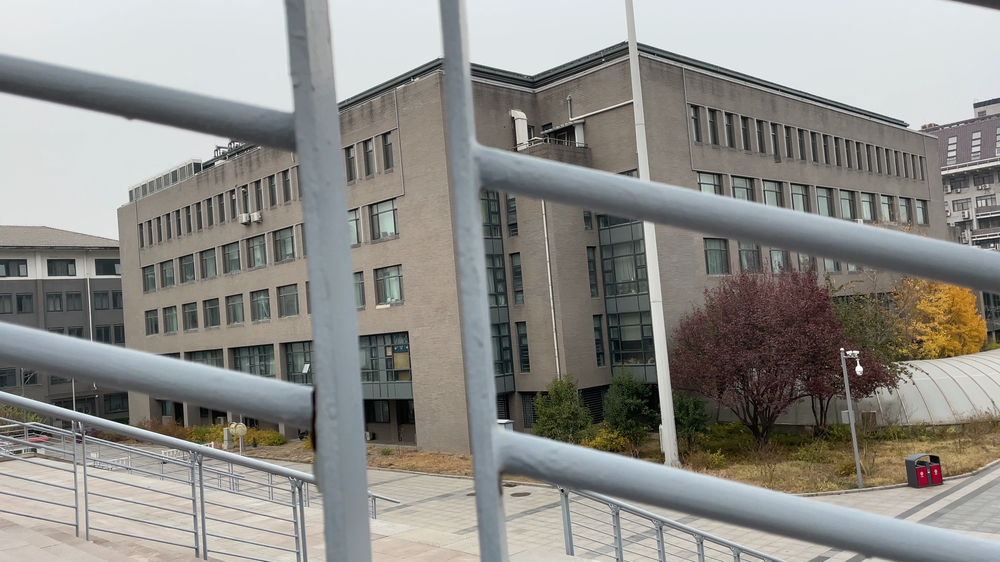} &
        \includegraphics[width=0.22\textwidth]{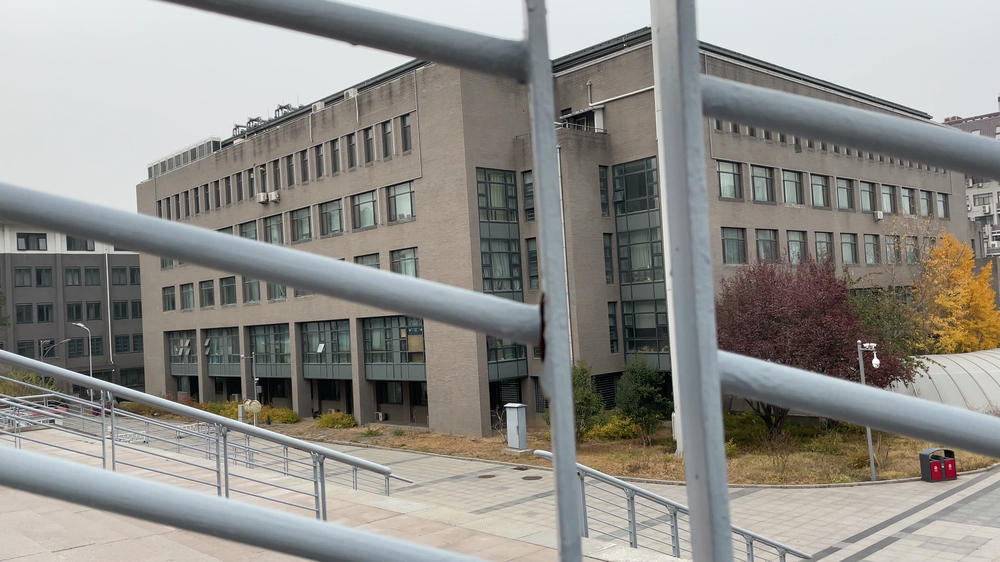} &
        \includegraphics[width=0.22\textwidth]{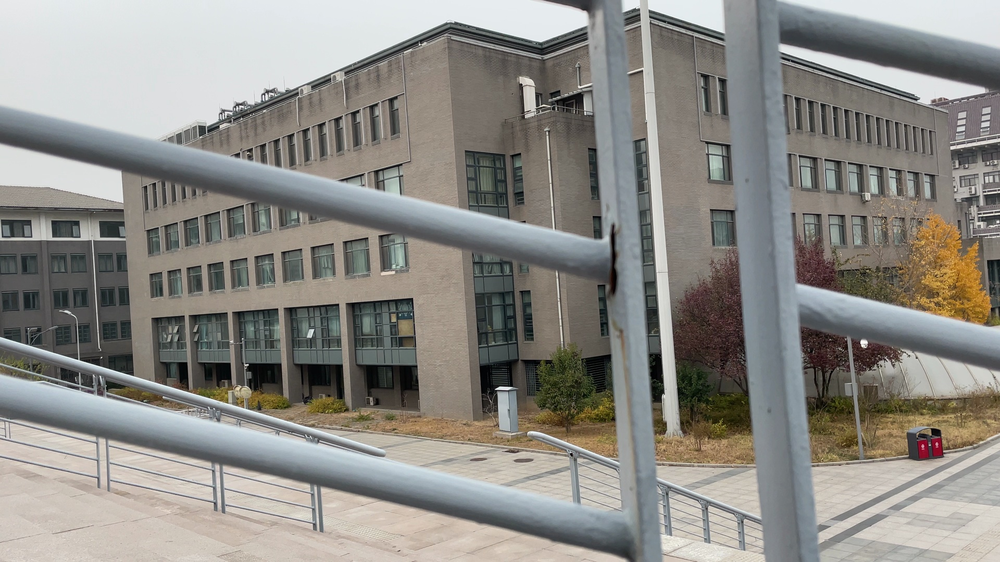} \\ [-0.7mm]
        \includegraphics[width=0.22\textwidth]{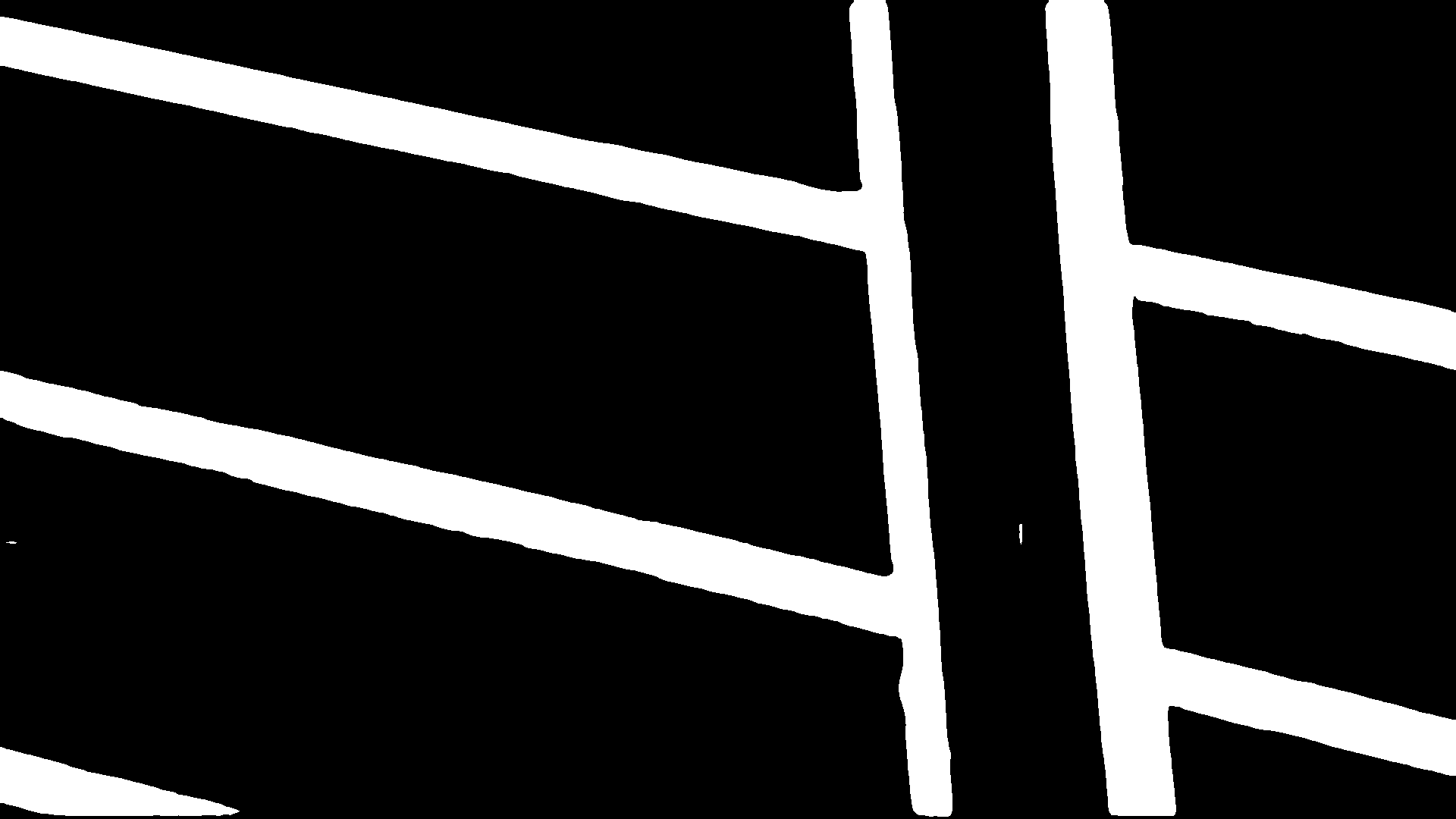} &
        \includegraphics[width=0.22\textwidth]{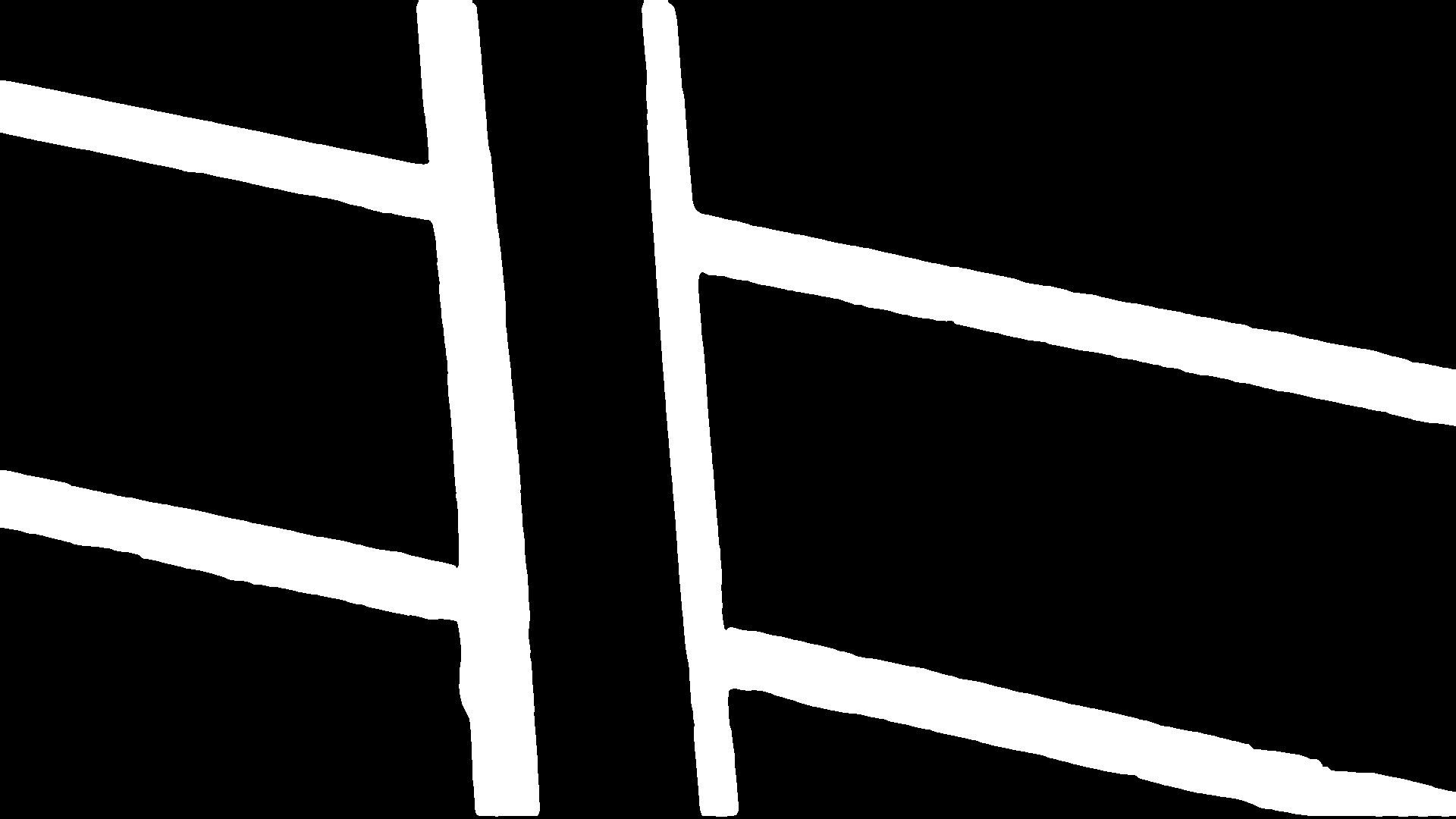} &
        \includegraphics[width=0.22\textwidth]{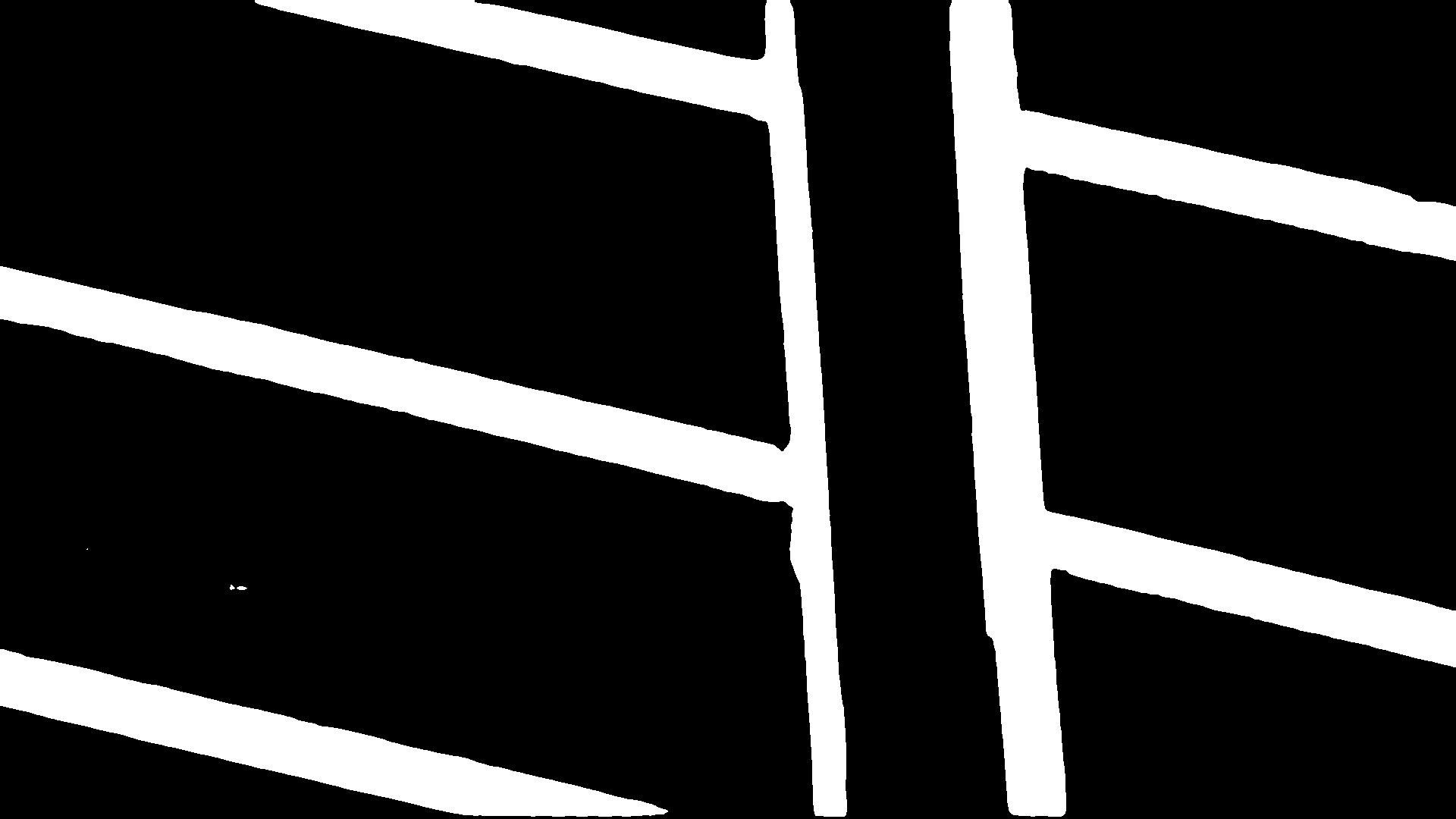} &
        \includegraphics[width=0.22\textwidth]{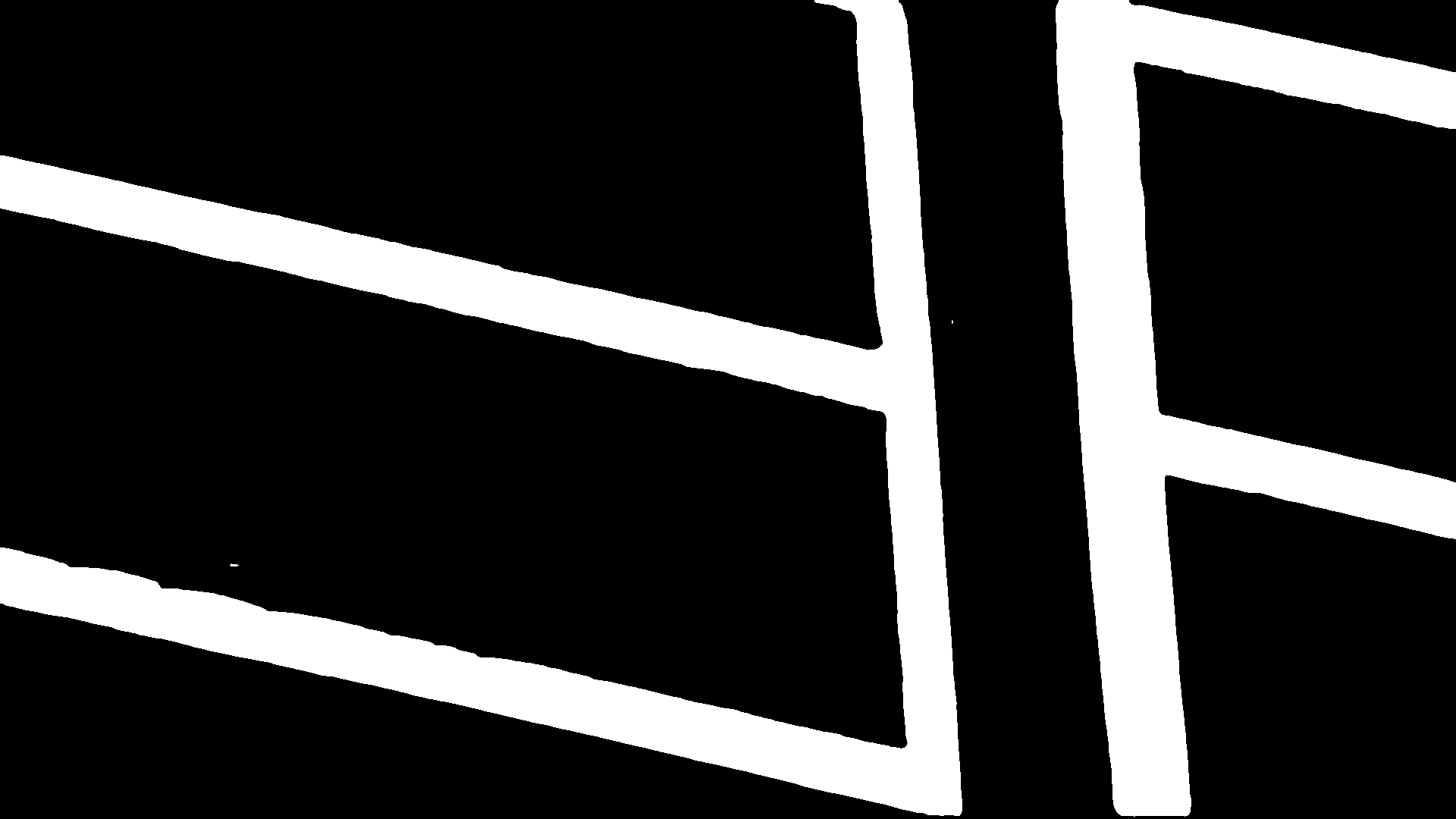} \\ [-0.7mm]
        \includegraphics[width=0.22\textwidth]{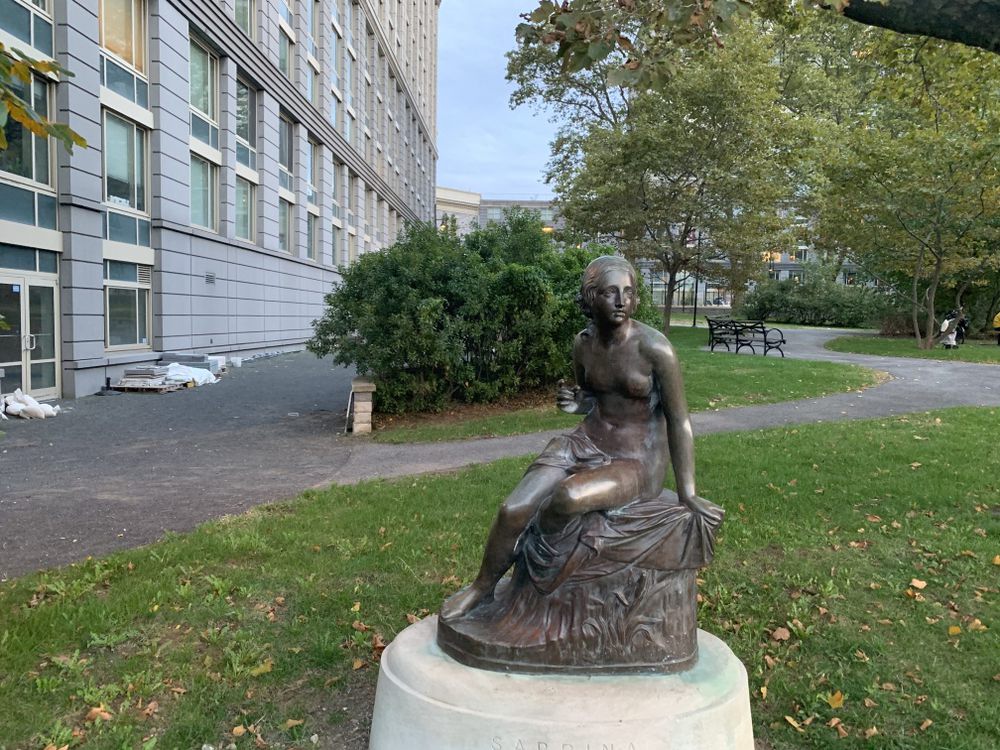} &
        \includegraphics[width=0.22\textwidth]{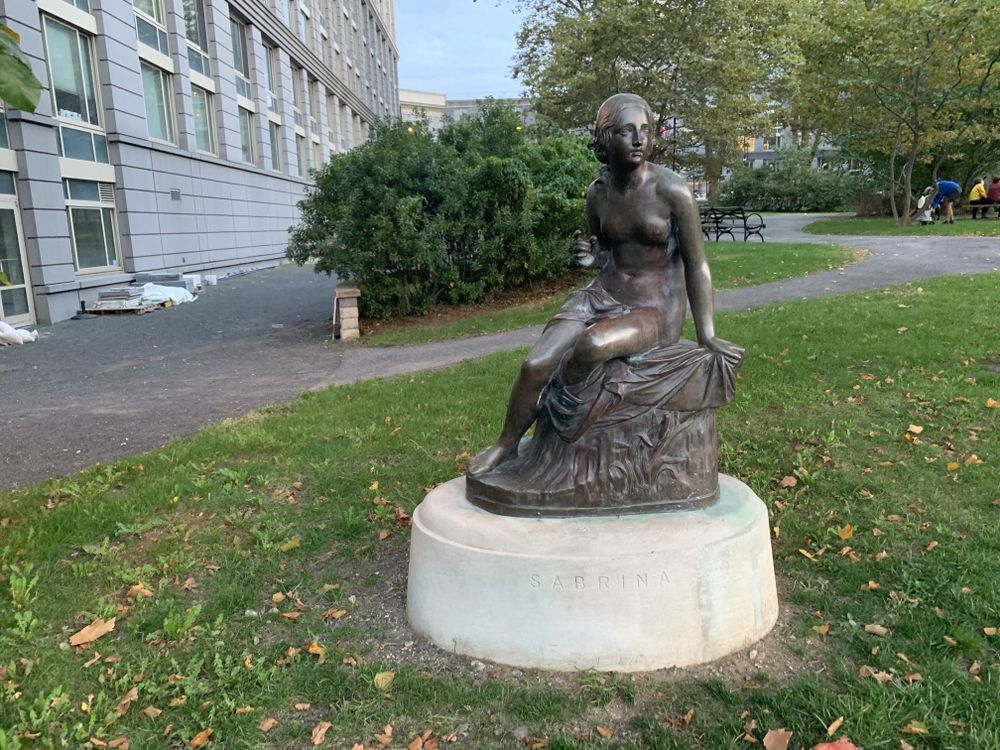} &
        \includegraphics[width=0.22\textwidth]{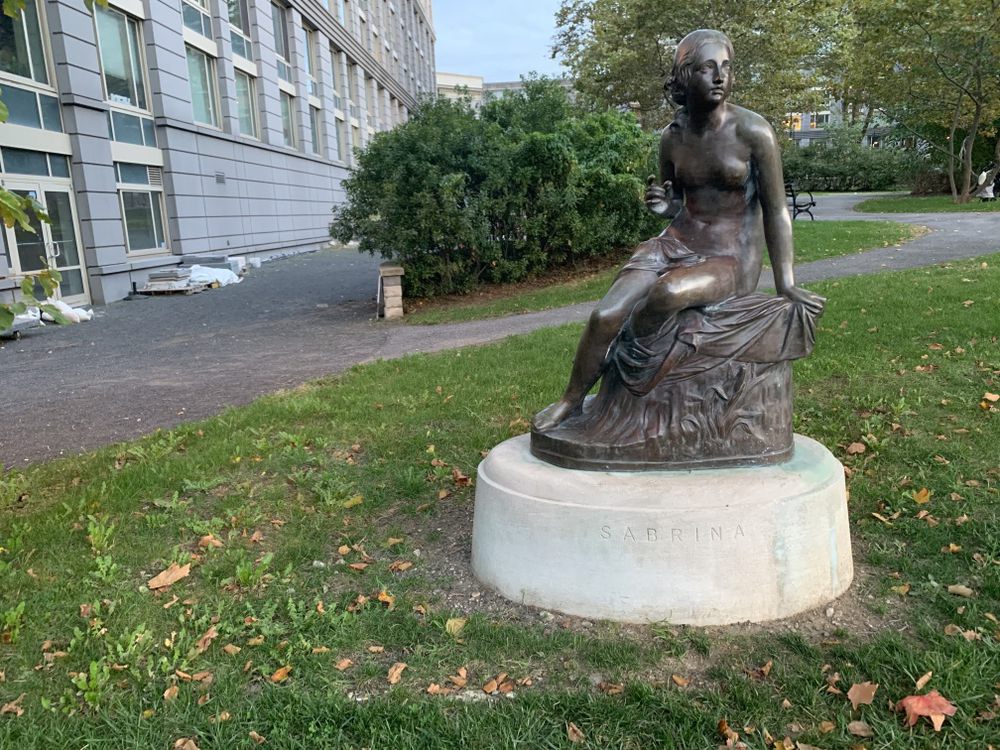} &
        \includegraphics[width=0.22\textwidth]{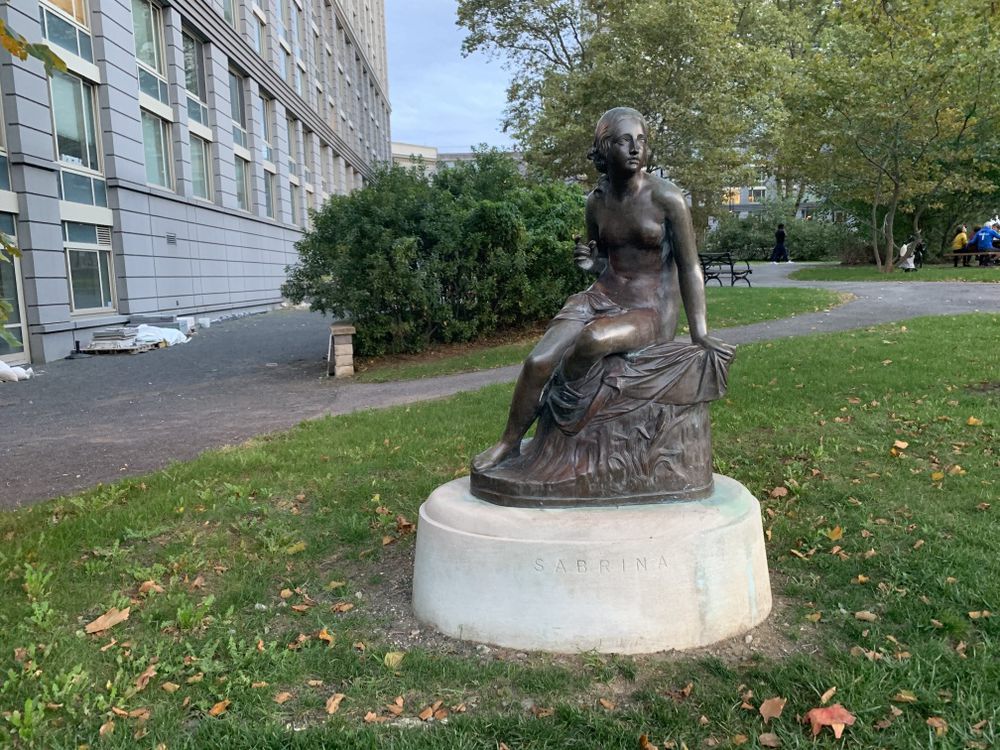} \\ [-0.7mm]
        \includegraphics[width=0.22\textwidth]{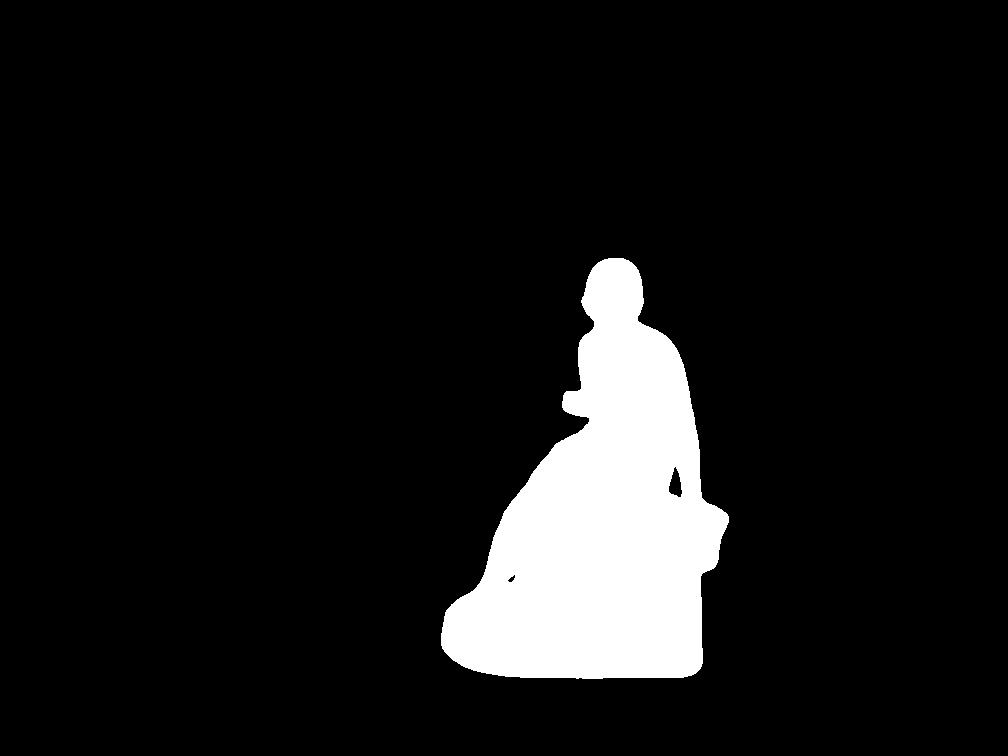} &
        \includegraphics[width=0.22\textwidth]{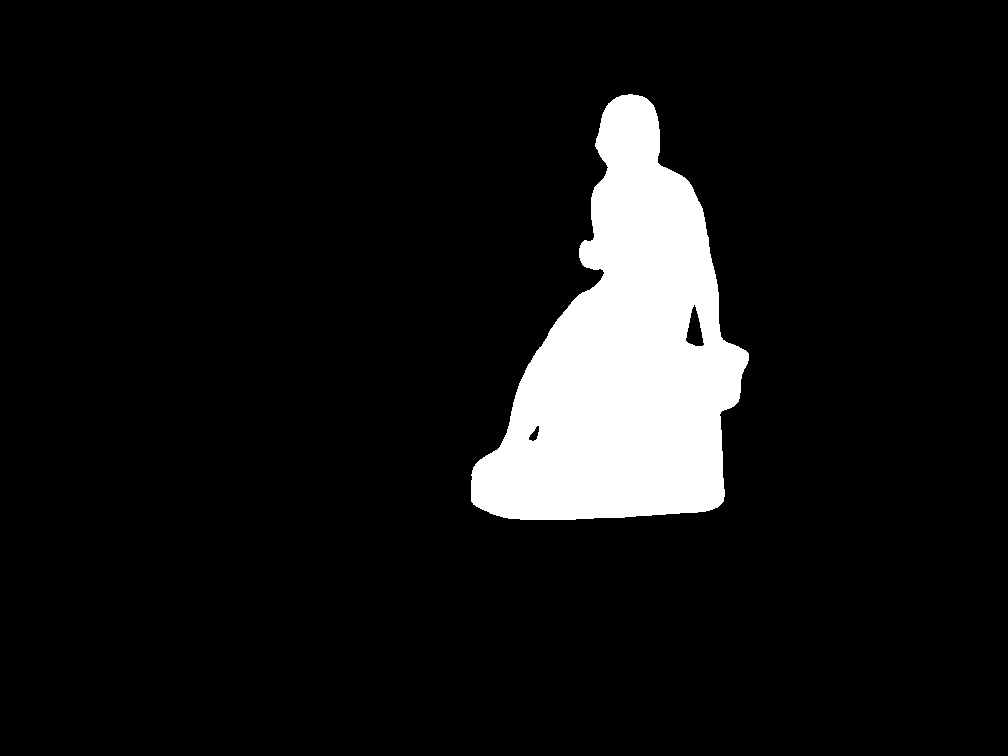} &
        \includegraphics[width=0.22\textwidth]{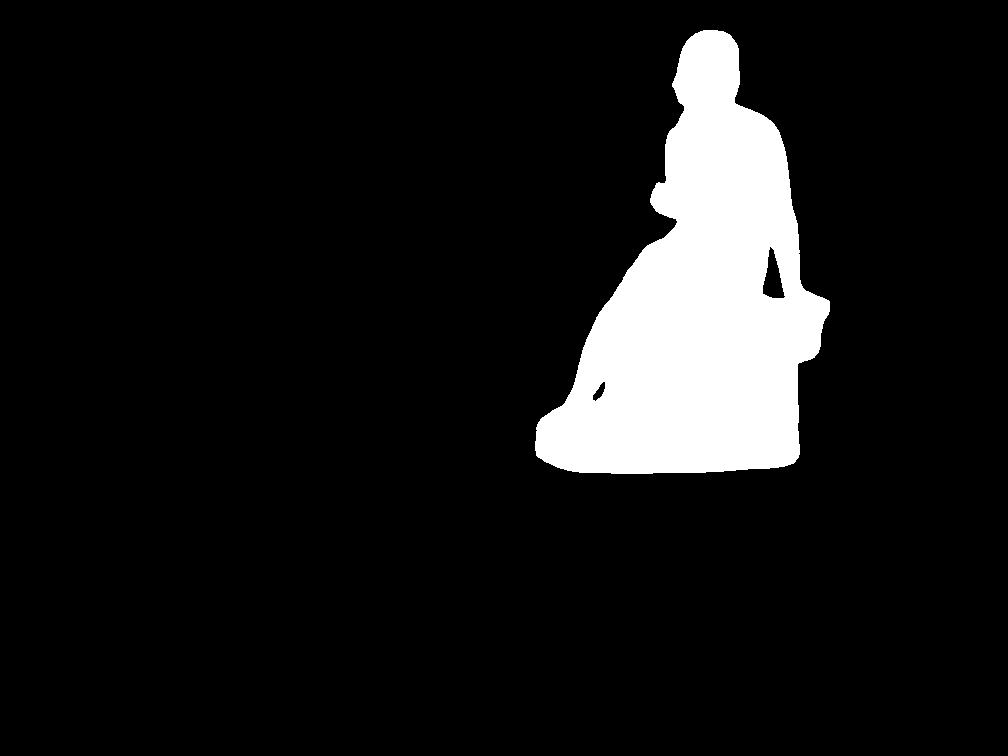} &
        \includegraphics[width=0.22\textwidth]{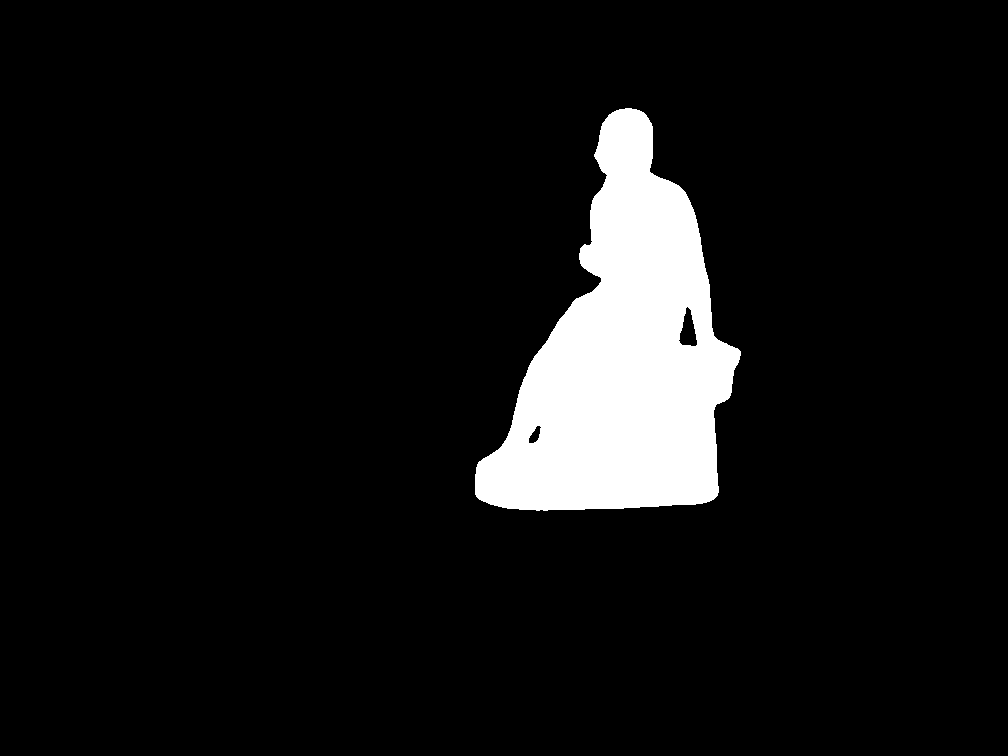} \\ [-0.7mm]
        \includegraphics[width=0.22\textwidth]{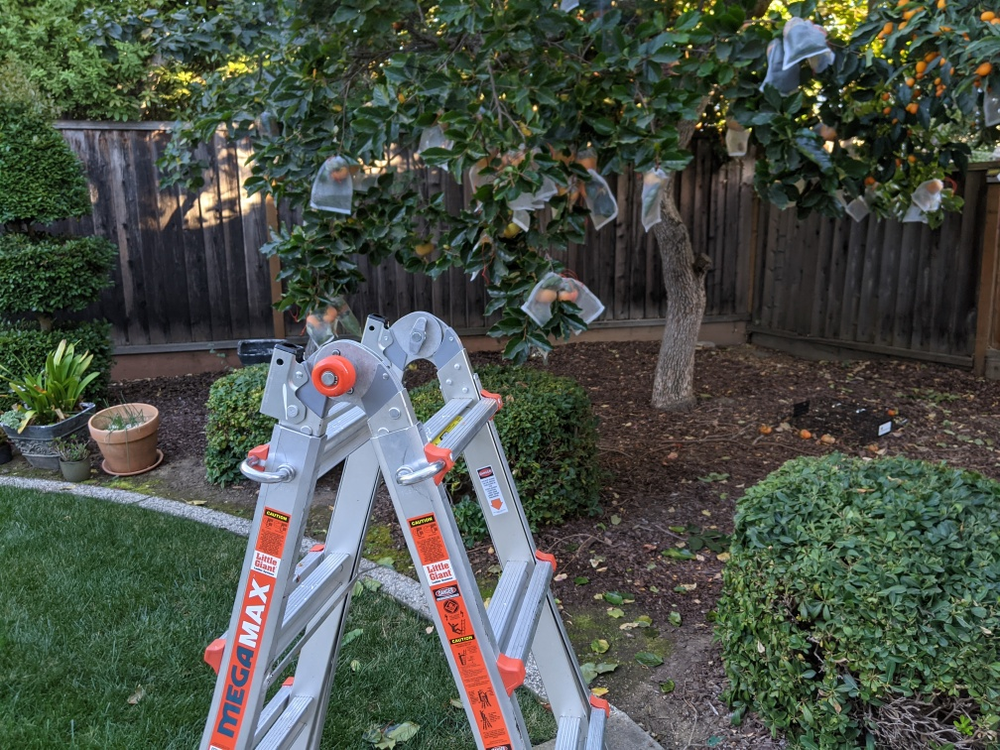} &
        \includegraphics[width=0.22\textwidth]{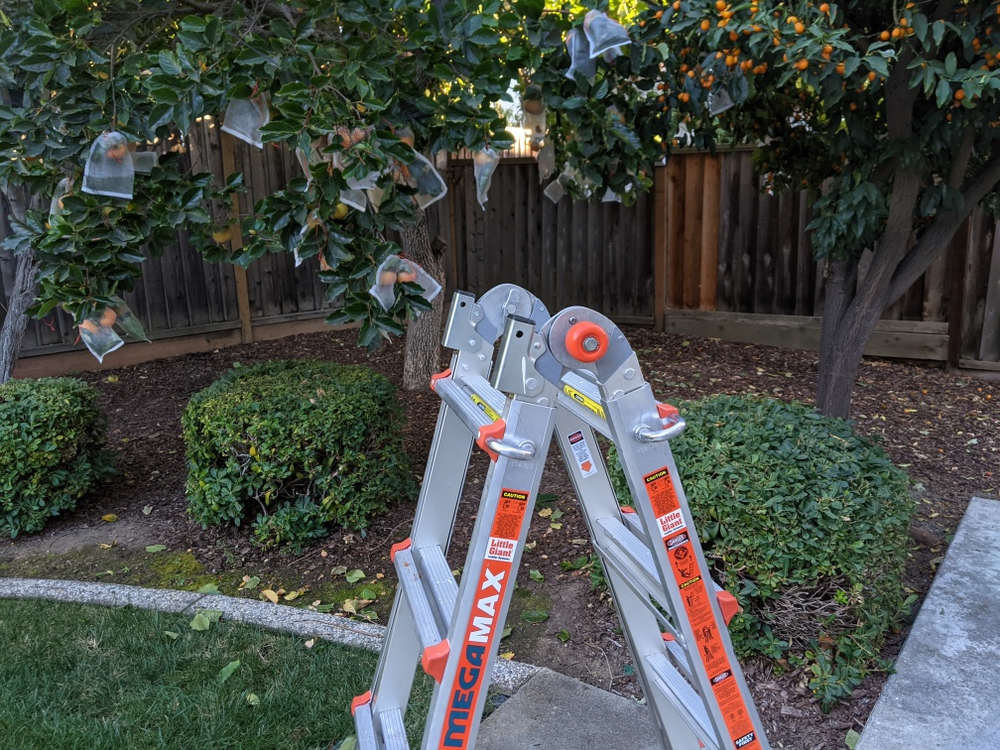} &
        \includegraphics[width=0.22\textwidth]{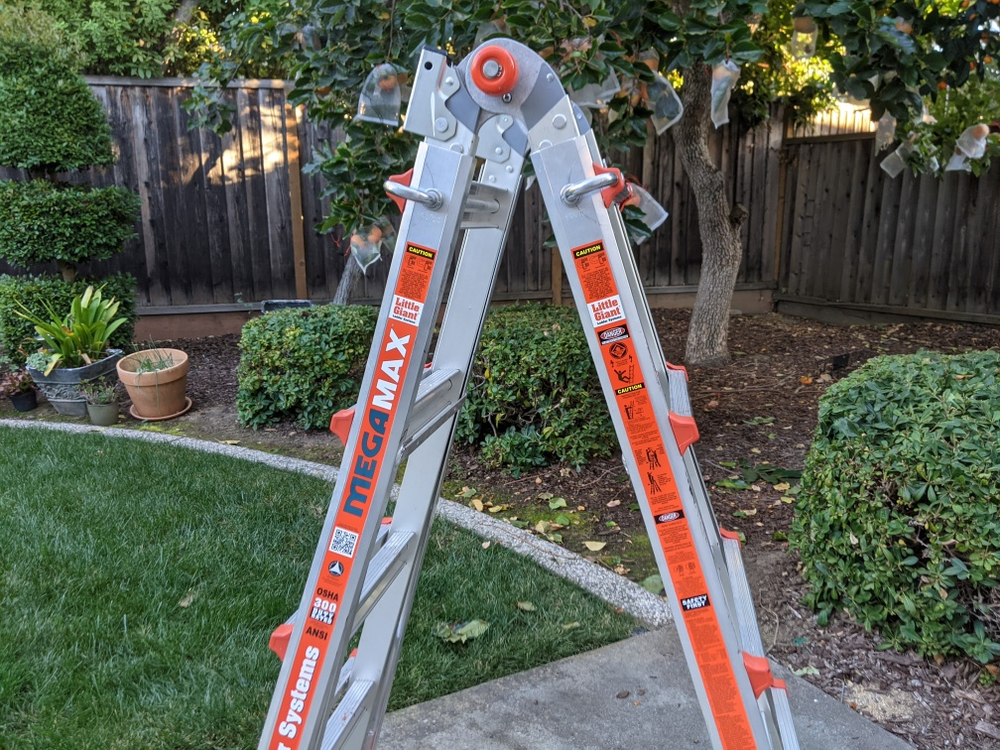} &
        \includegraphics[width=0.22\textwidth]{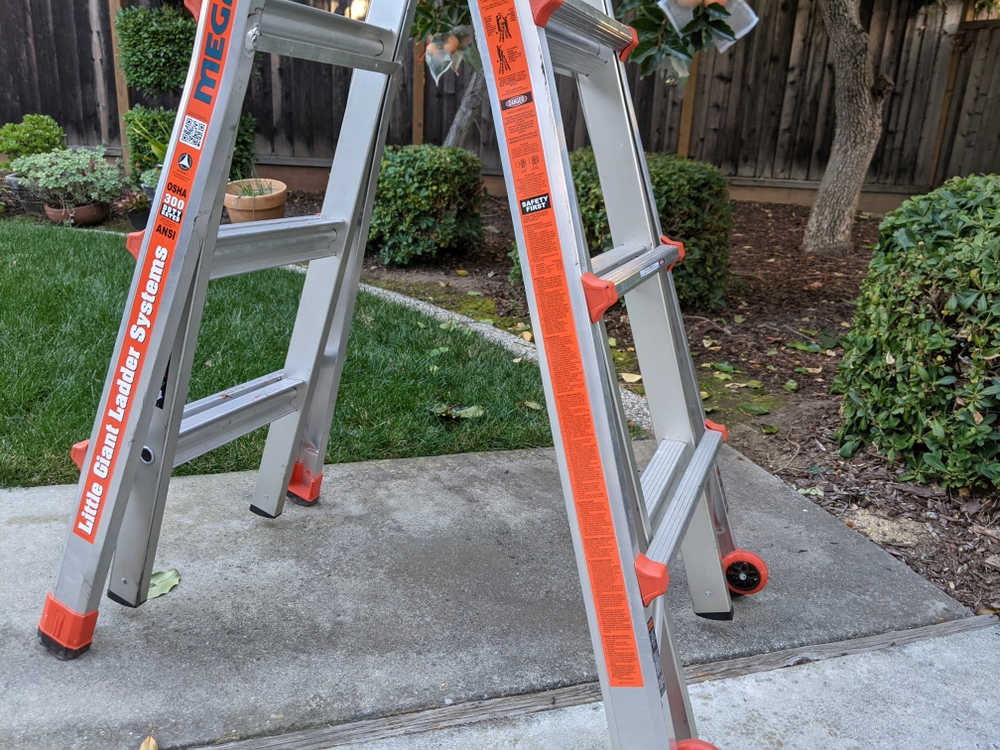} \\ [-0.7mm]
        \includegraphics[width=0.22\textwidth]{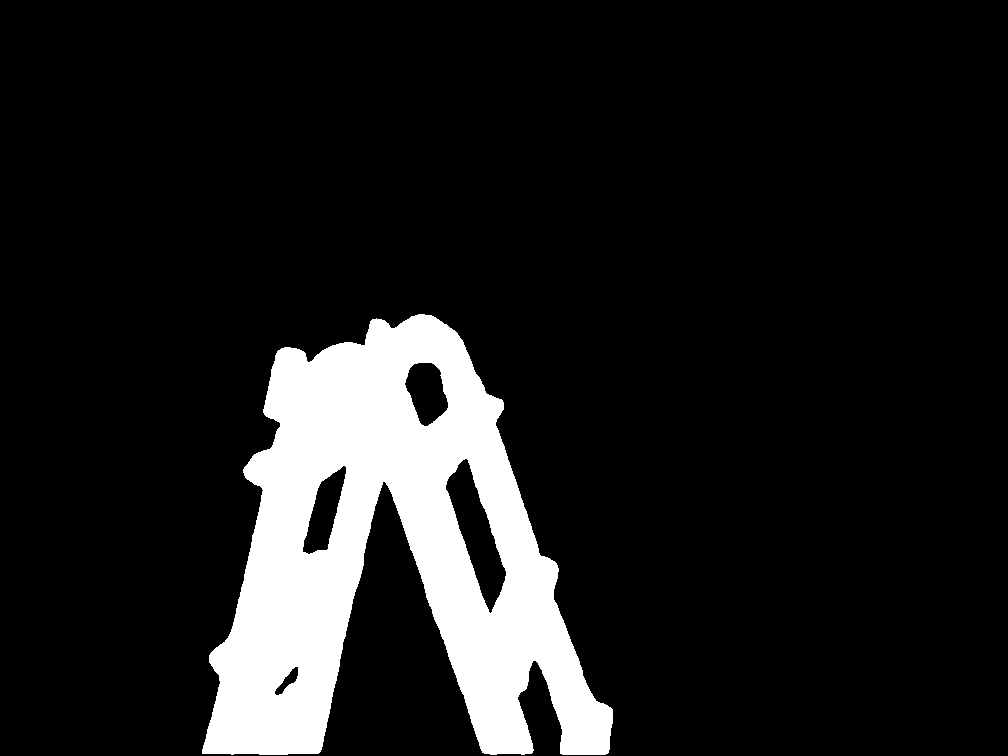} &
        \includegraphics[width=0.22\textwidth]{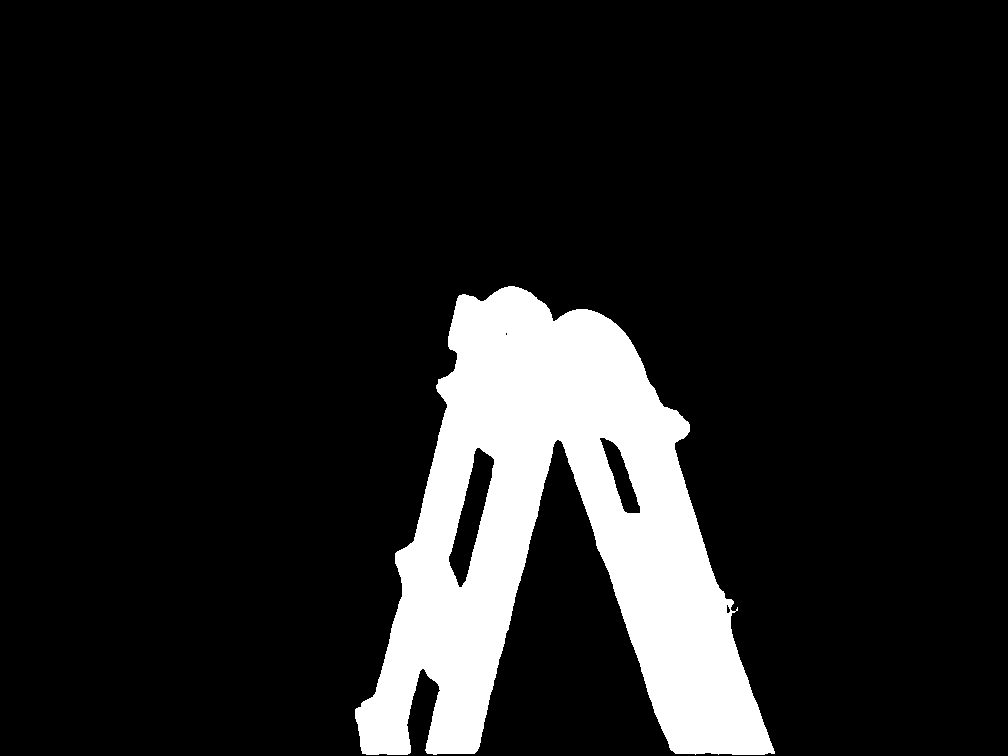} &
        \includegraphics[width=0.22\textwidth]{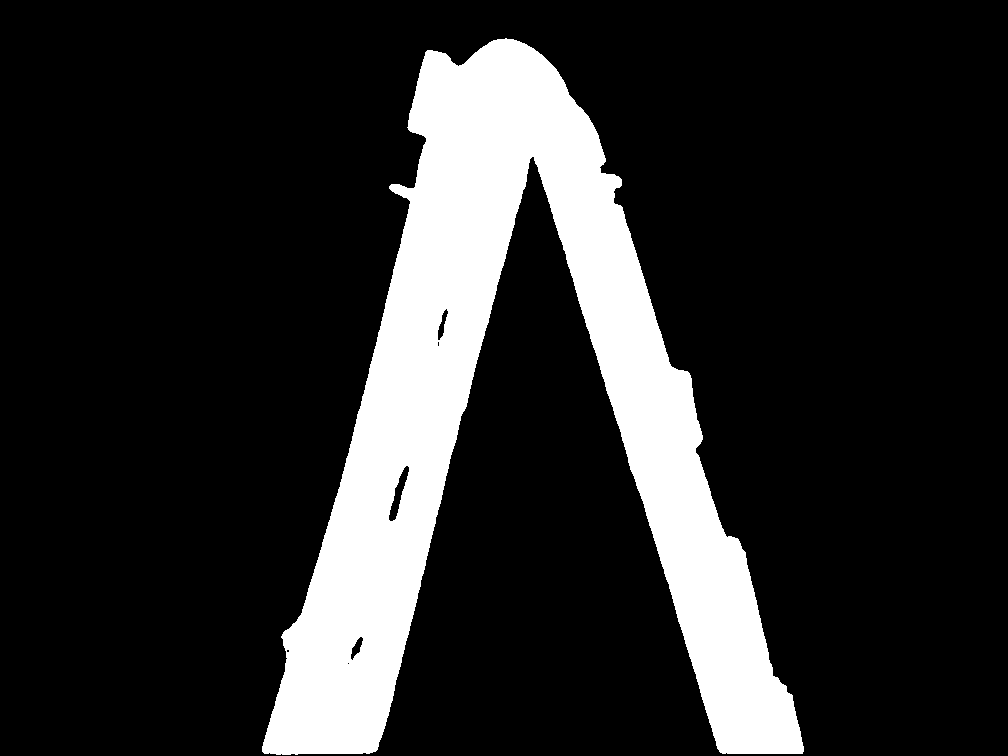} &
        \includegraphics[width=0.22\textwidth]{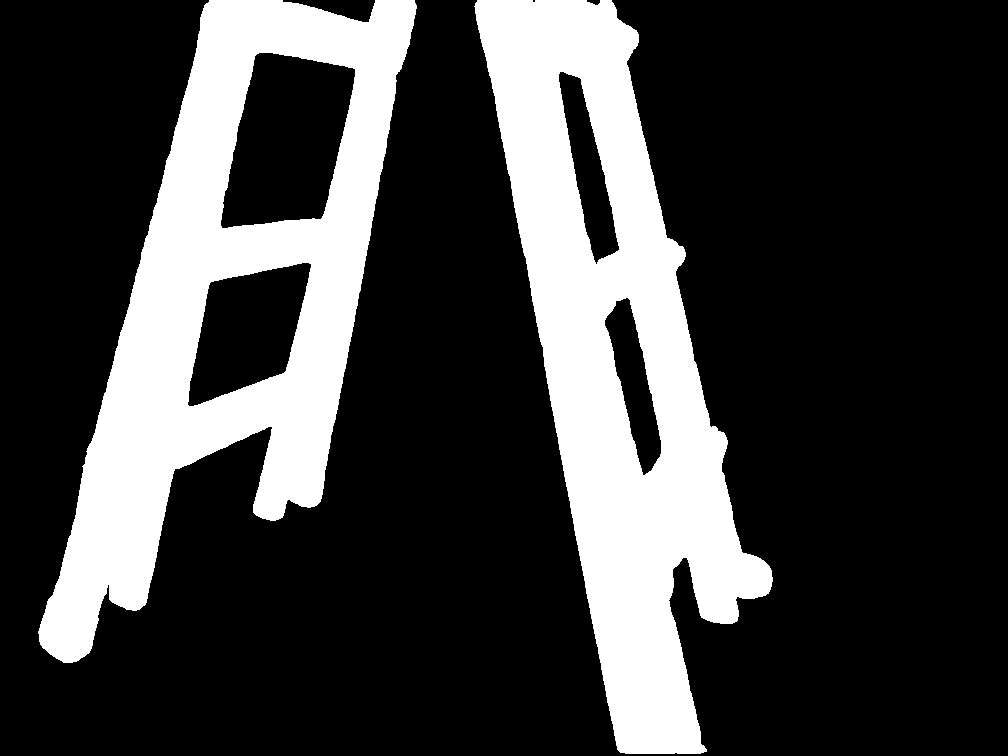}
    \end{tabular}
    \vspace{-2mm}
    \caption{\textbf{DeclutterSet Illustration (Part I).} From the top to the bottom: (a) Orchids, (b) Railing, (c) Statue, (d) Ladder.}
    \label{fig:data_1}
\end{figure*}

\begin{figure*}[t]
    \centering
    \begin{tabular}{@{}c@{\hspace{0.5mm}}c@{\hspace{0.5mm}}c@{\hspace{0.5mm}}c@{}}
        \includegraphics[width=0.22\textwidth]{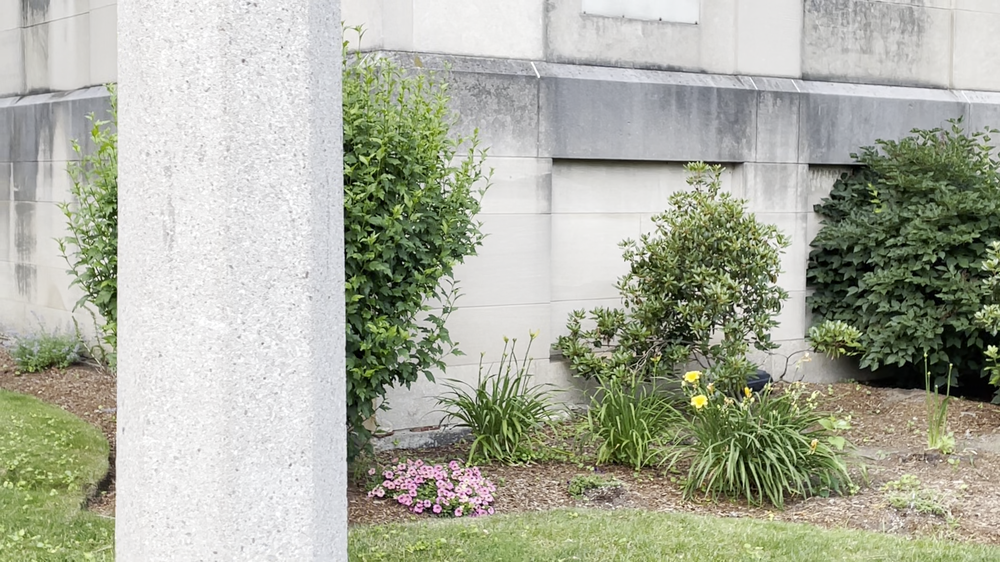} &
        \includegraphics[width=0.22\textwidth]{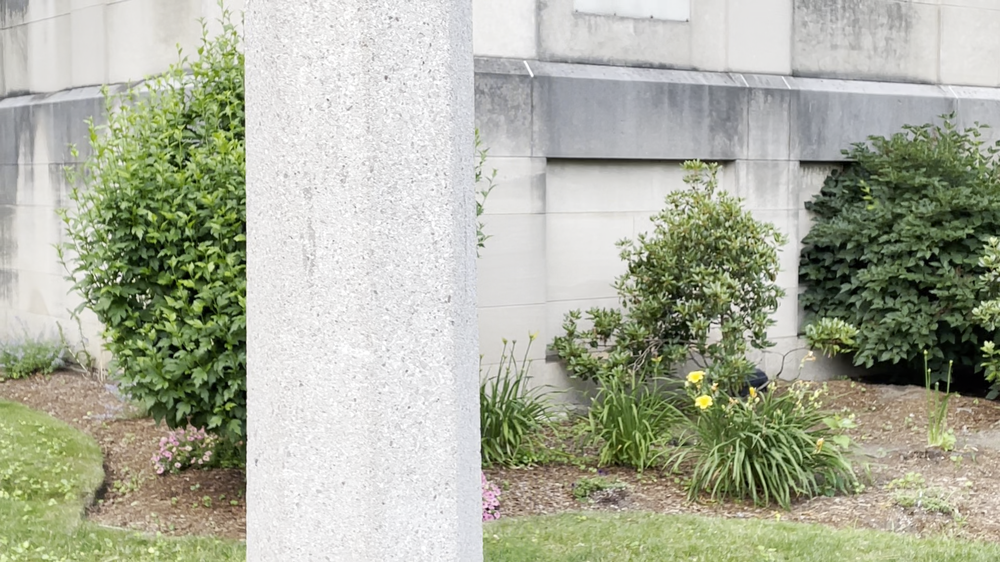} &
        \includegraphics[width=0.22\textwidth]{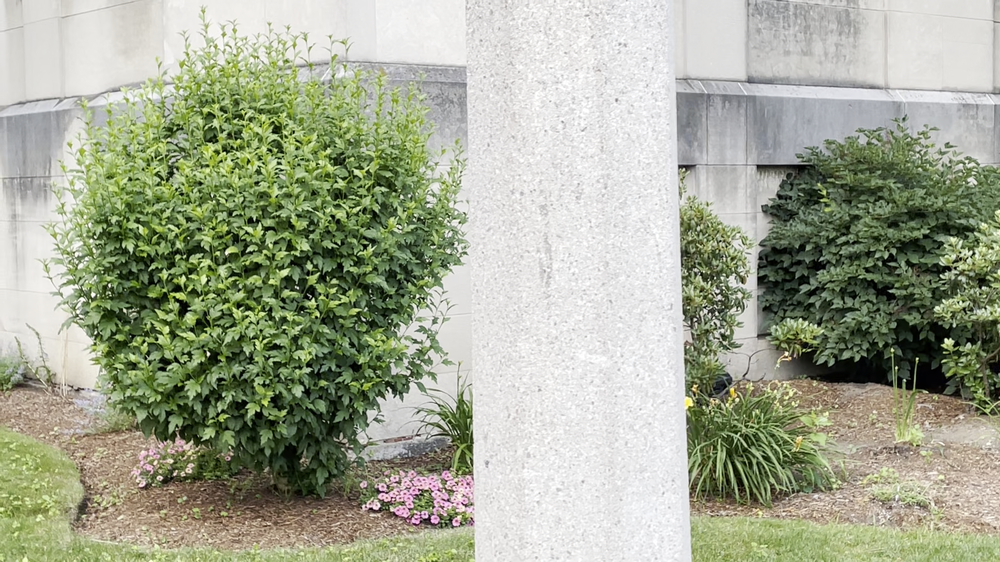} &
        \includegraphics[width=0.22\textwidth]{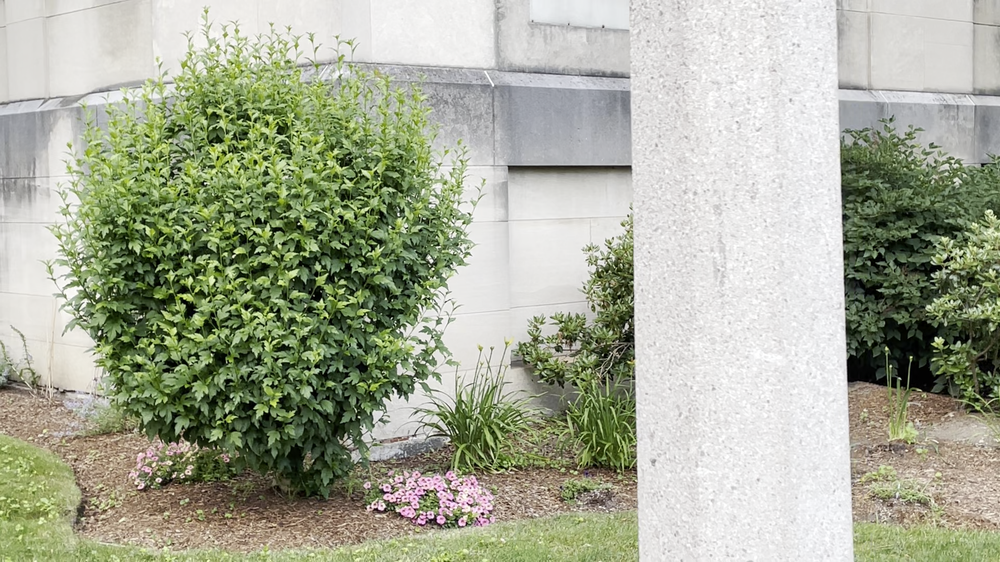} \\ [-0.7mm]
        \includegraphics[width=0.22\textwidth]{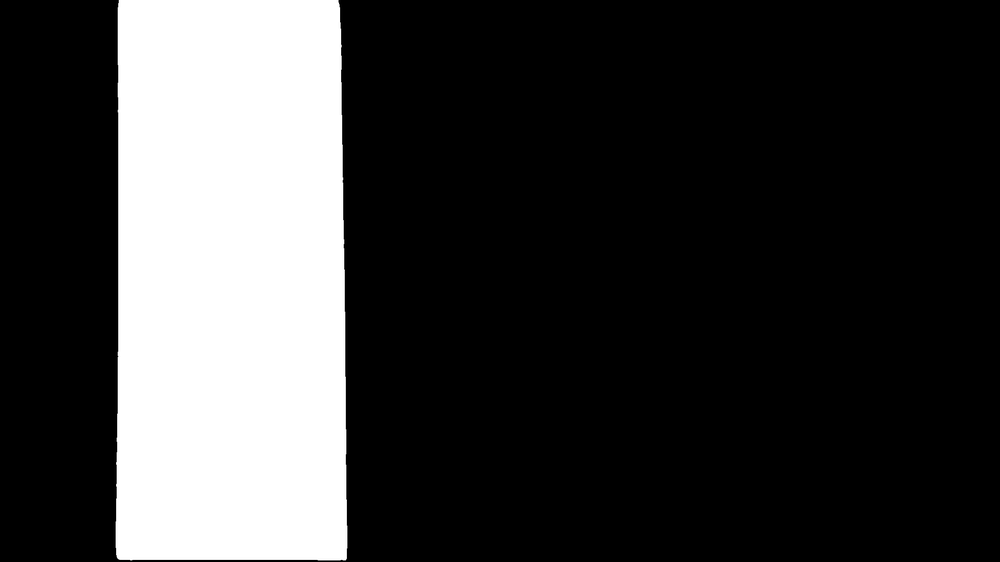} &
        \includegraphics[width=0.22\textwidth]{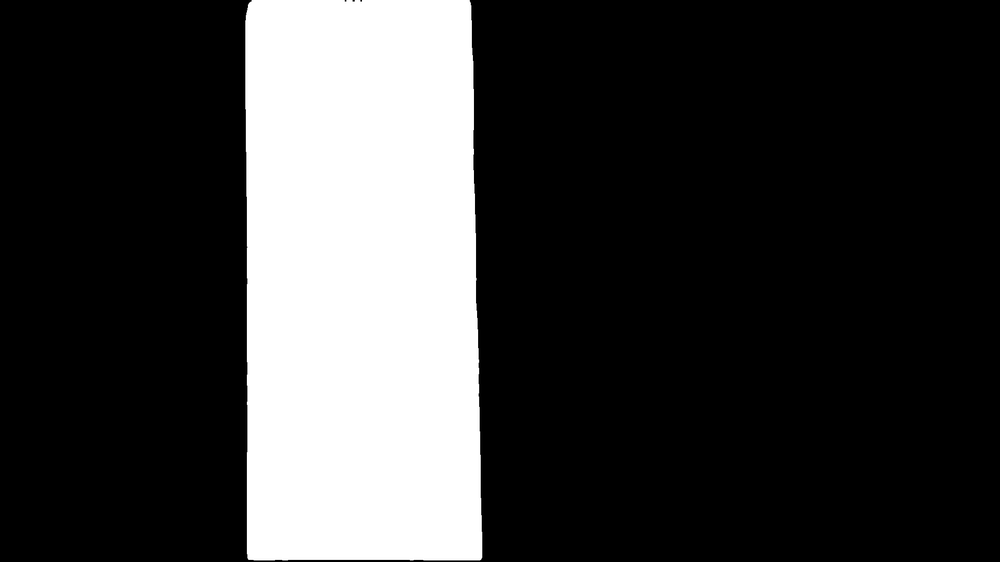} &
        \includegraphics[width=0.22\textwidth]{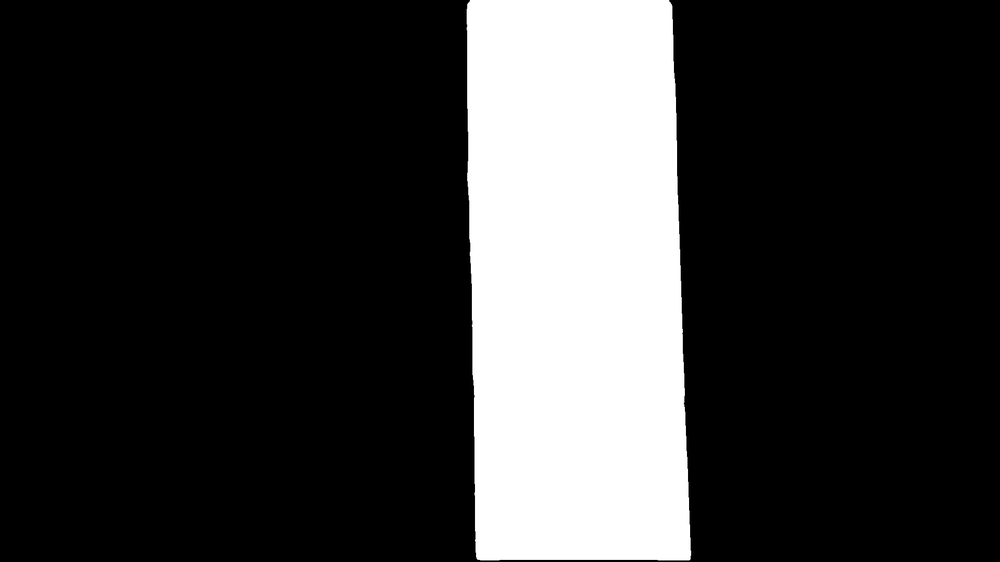} &
        \includegraphics[width=0.22\textwidth]{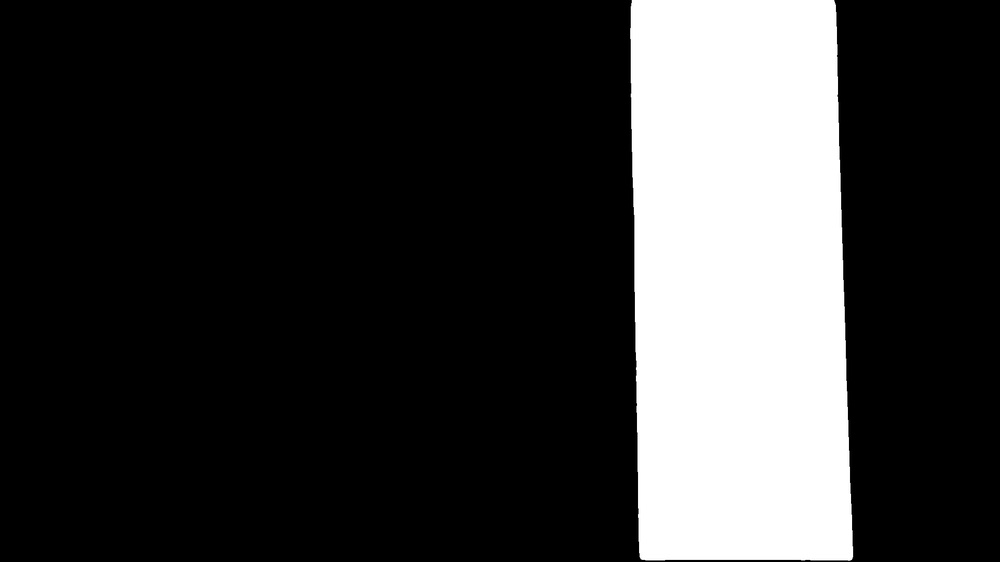} \\ [-0.7mm]
        \includegraphics[width=0.22\textwidth]{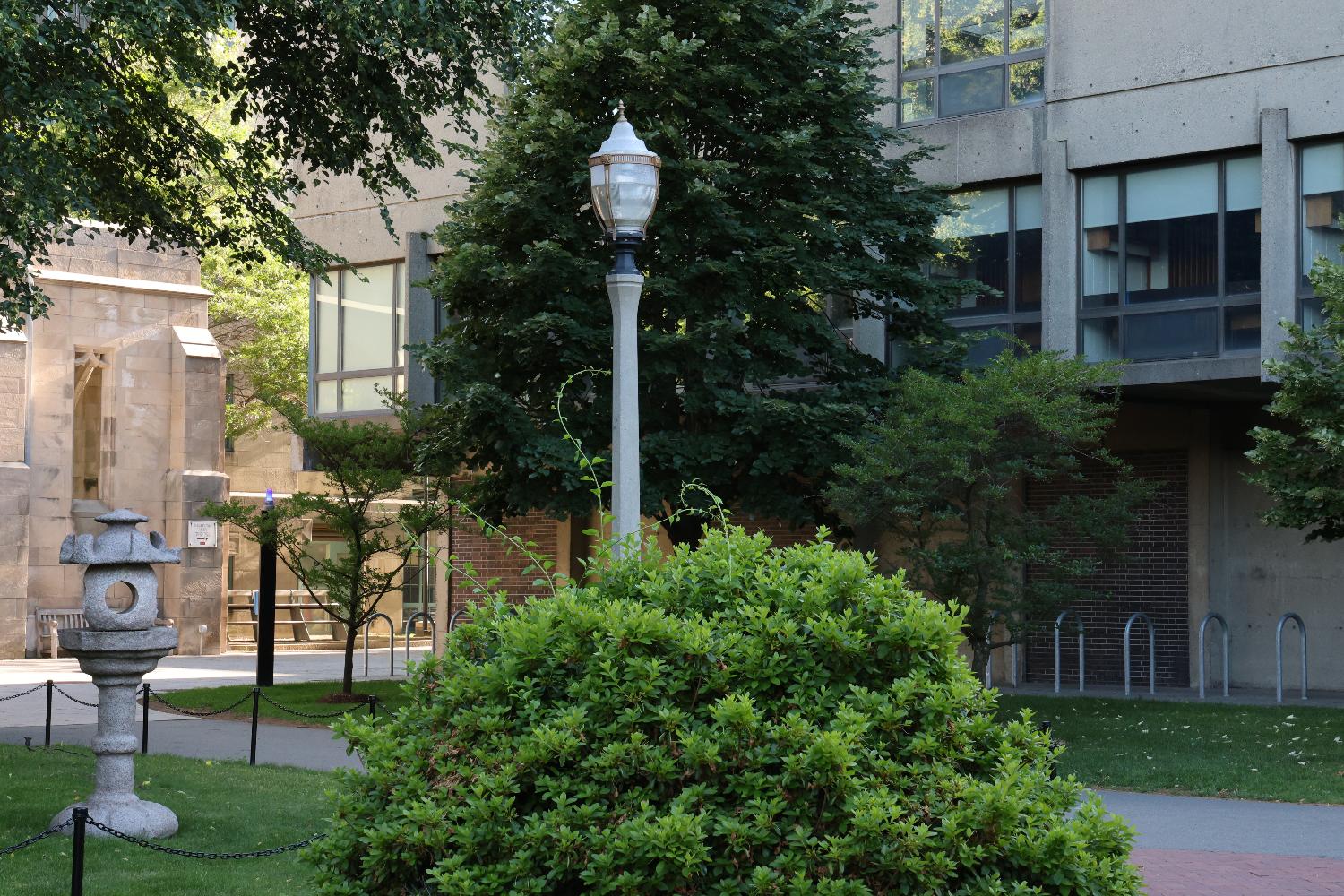} &
        \includegraphics[width=0.22\textwidth]{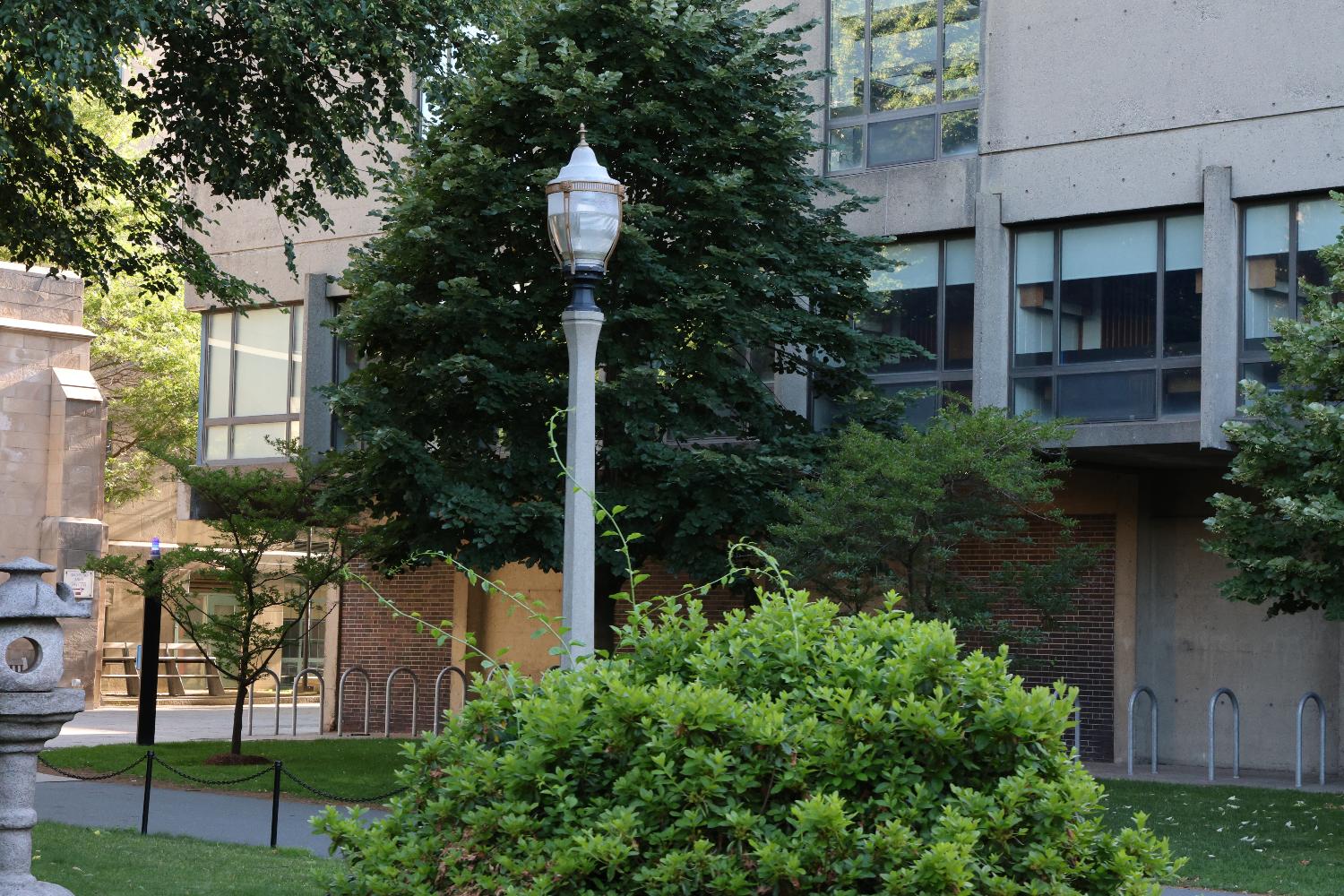} &
        \includegraphics[width=0.22\textwidth]{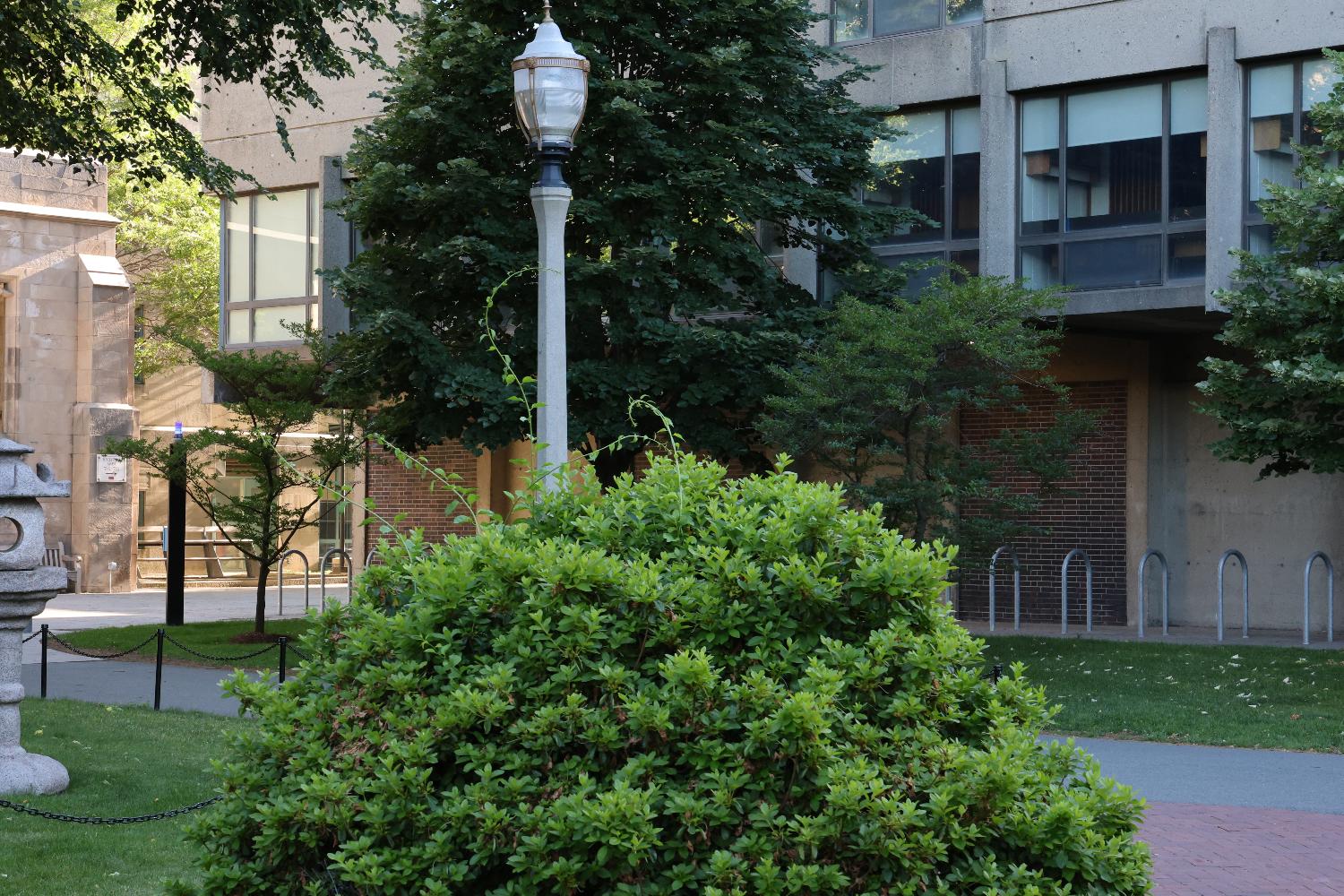} &
        \includegraphics[width=0.22\textwidth]{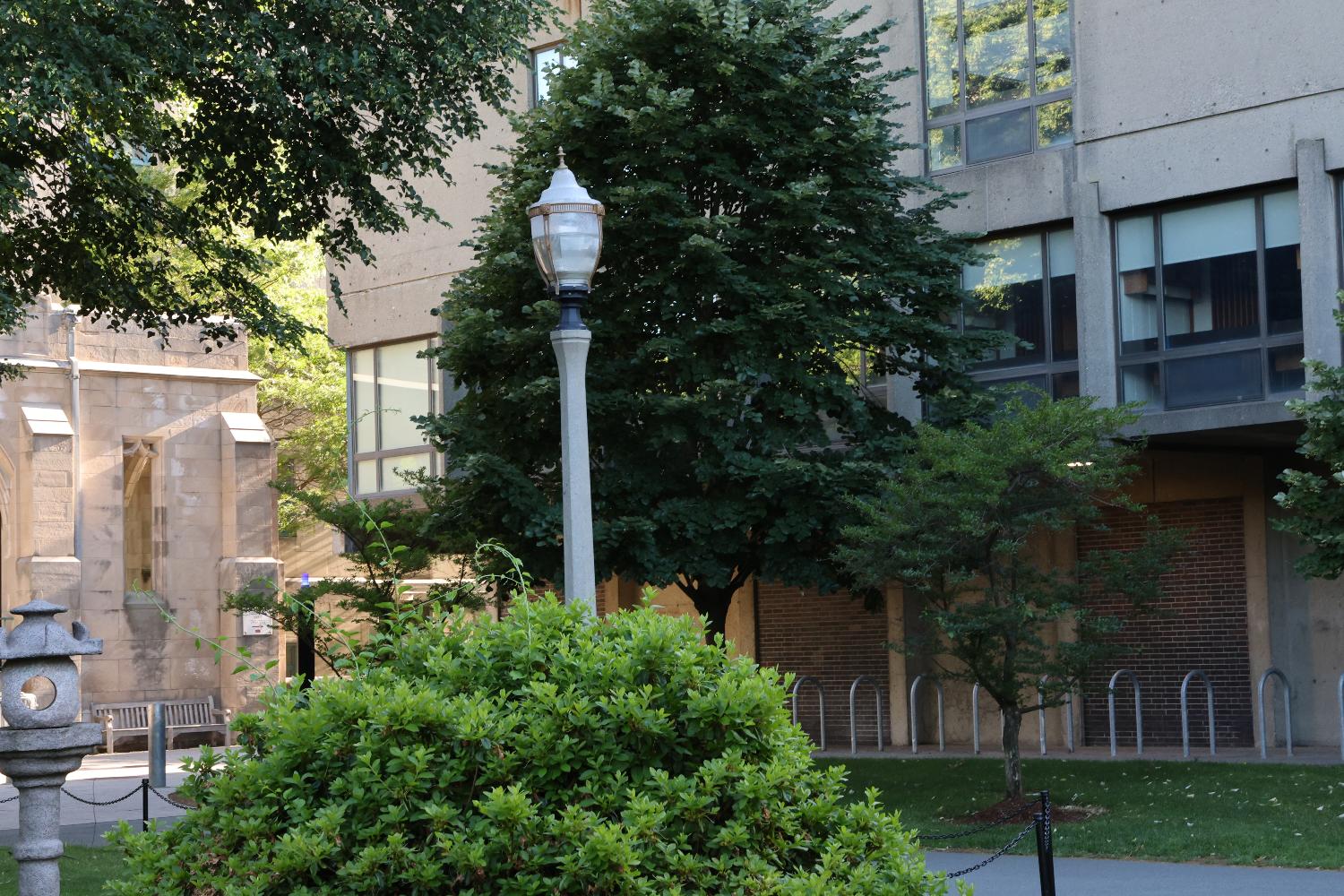} \\ [-0.7mm]
        \includegraphics[width=0.22\textwidth]{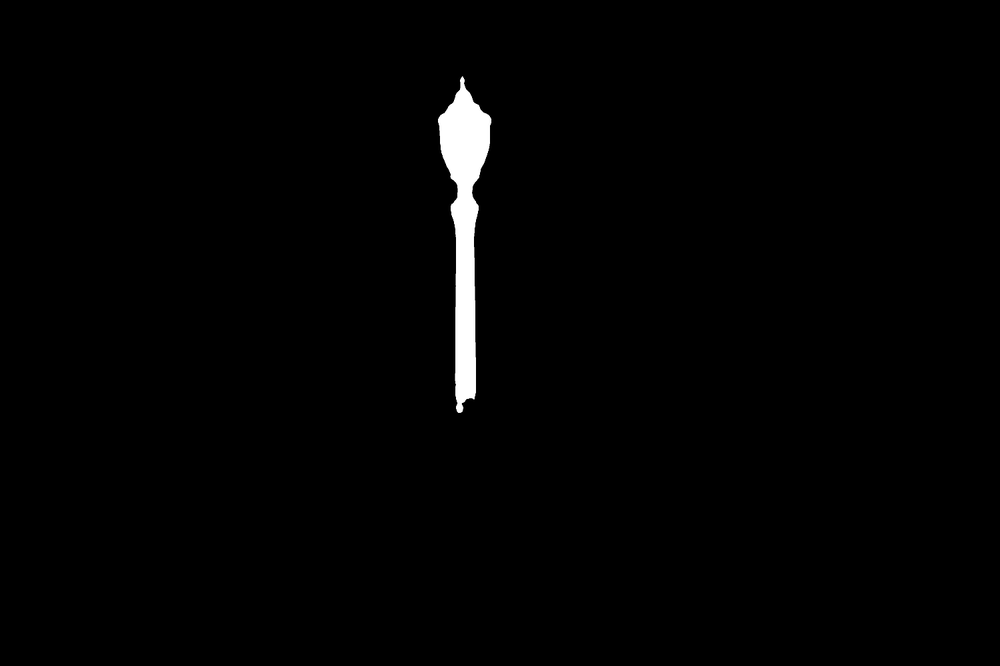} &
        \includegraphics[width=0.22\textwidth]{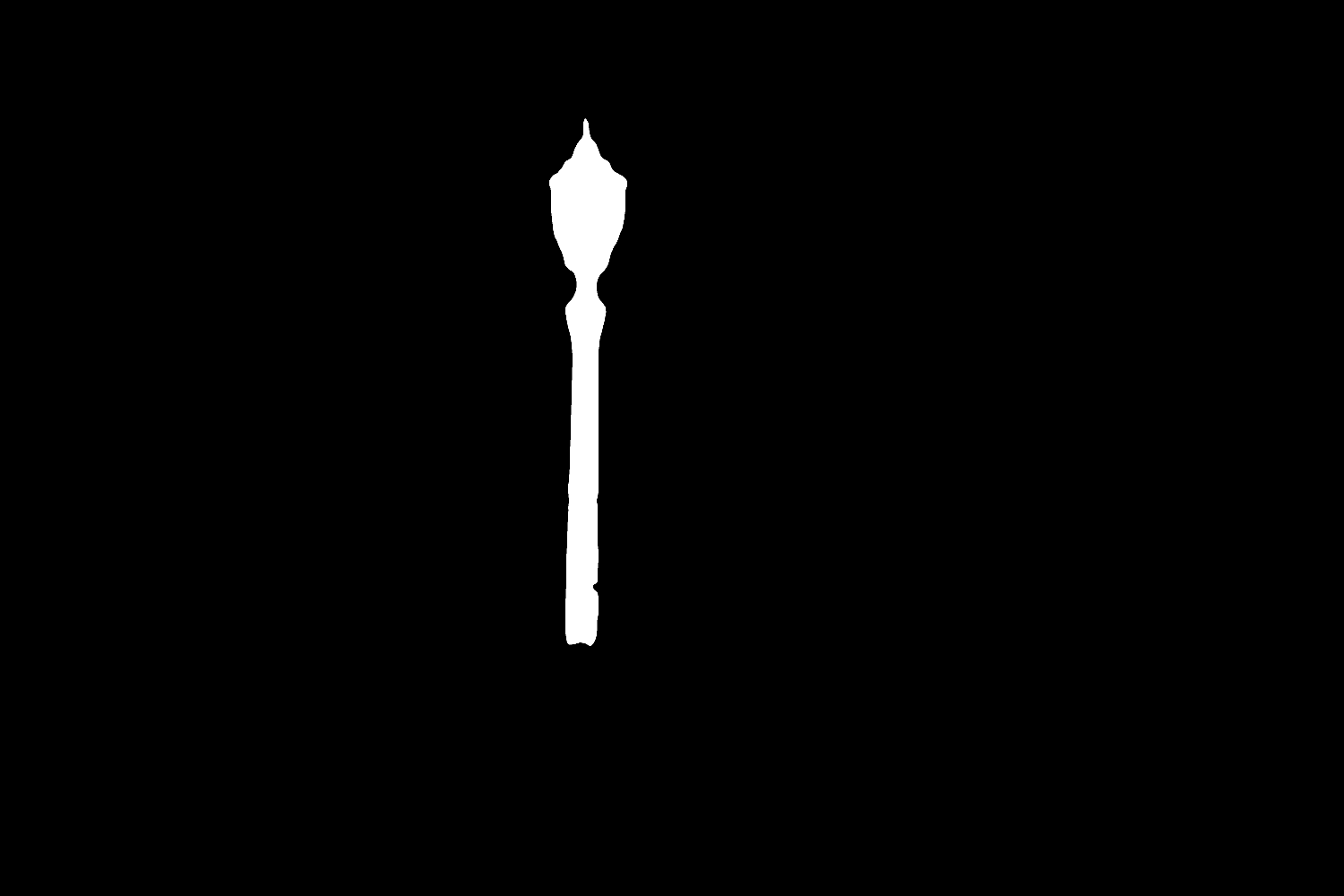} &
        \includegraphics[width=0.22\textwidth]{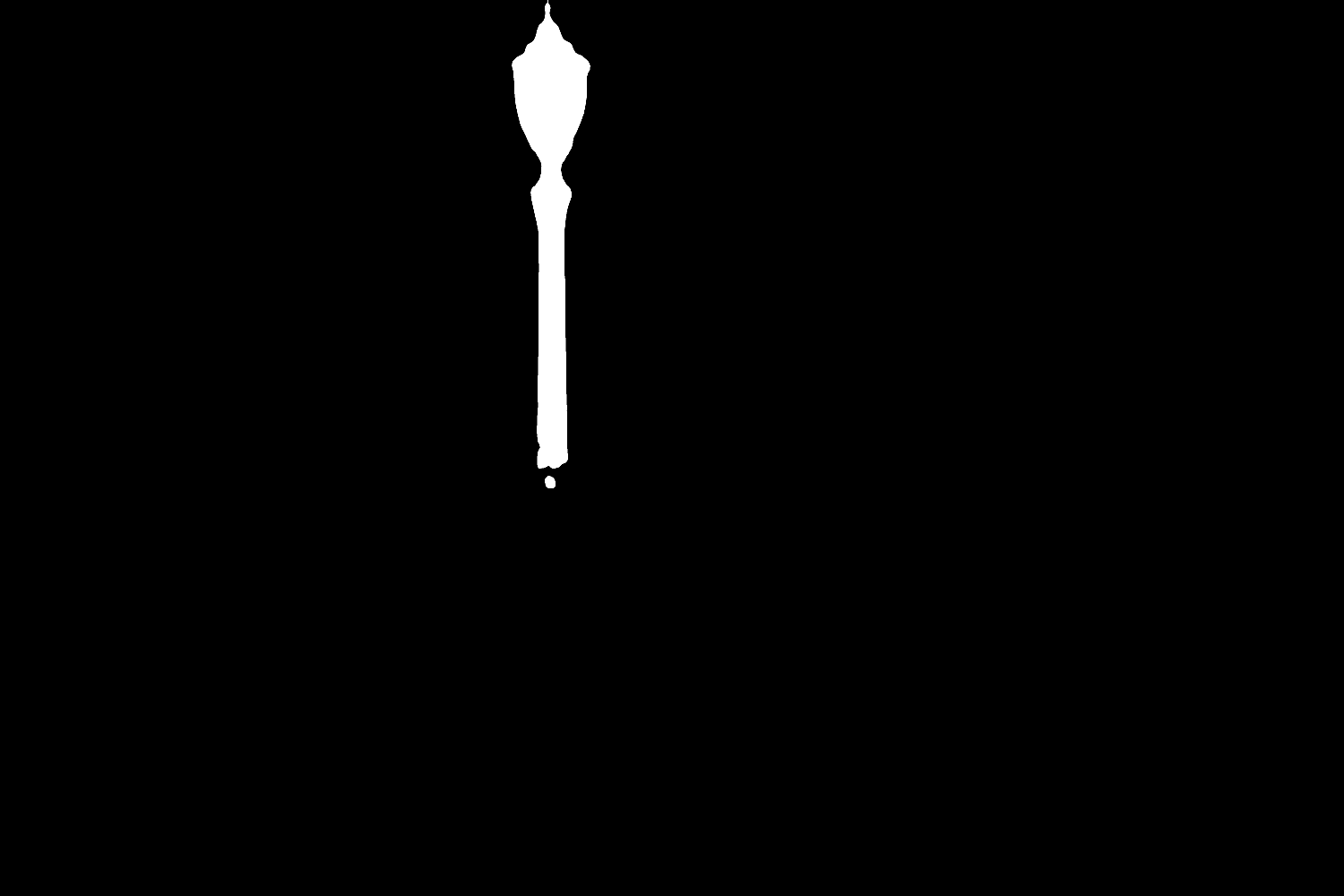} &
        \includegraphics[width=0.22\textwidth]{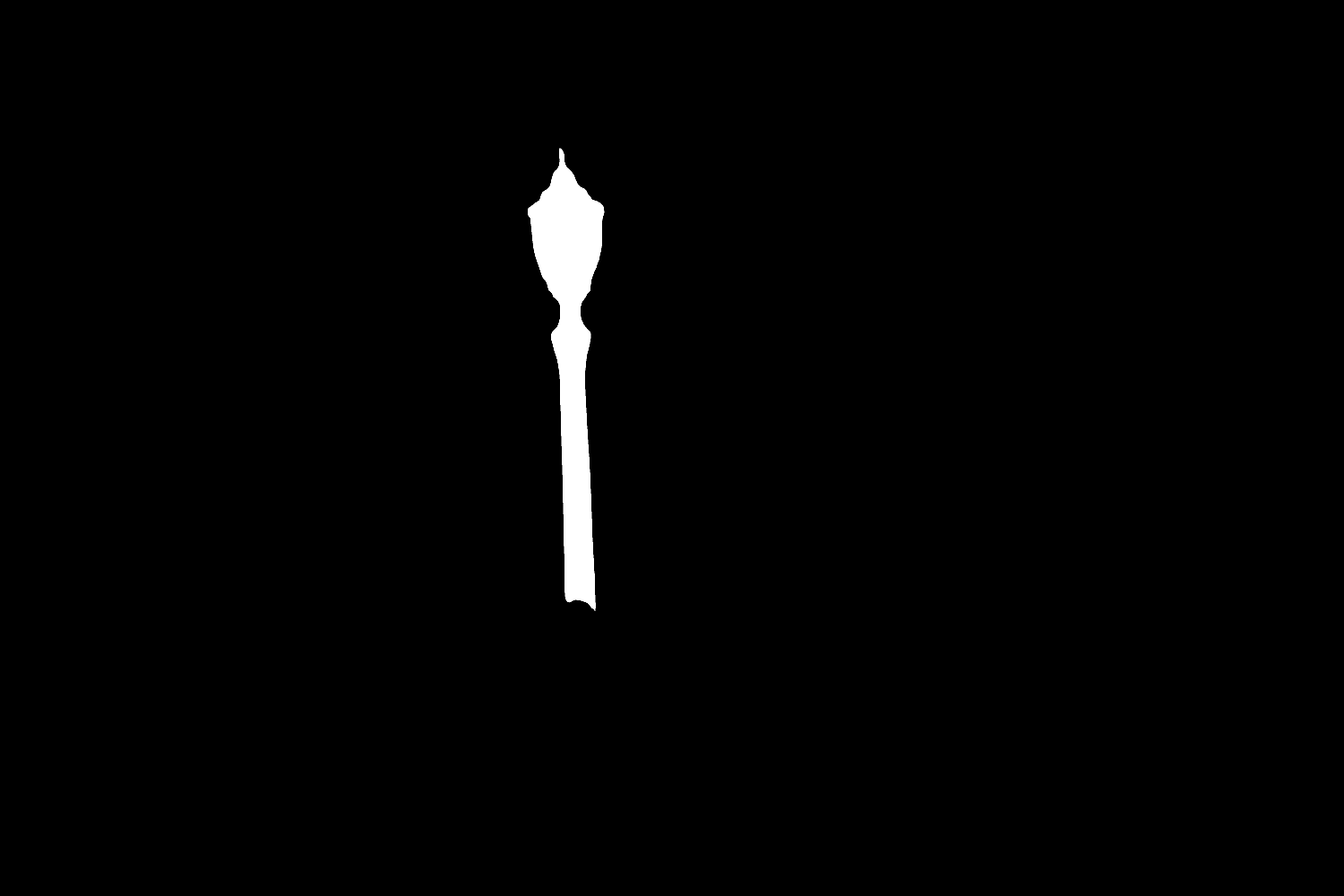} \\ [-0.7mm]
        \includegraphics[width=0.22\textwidth]{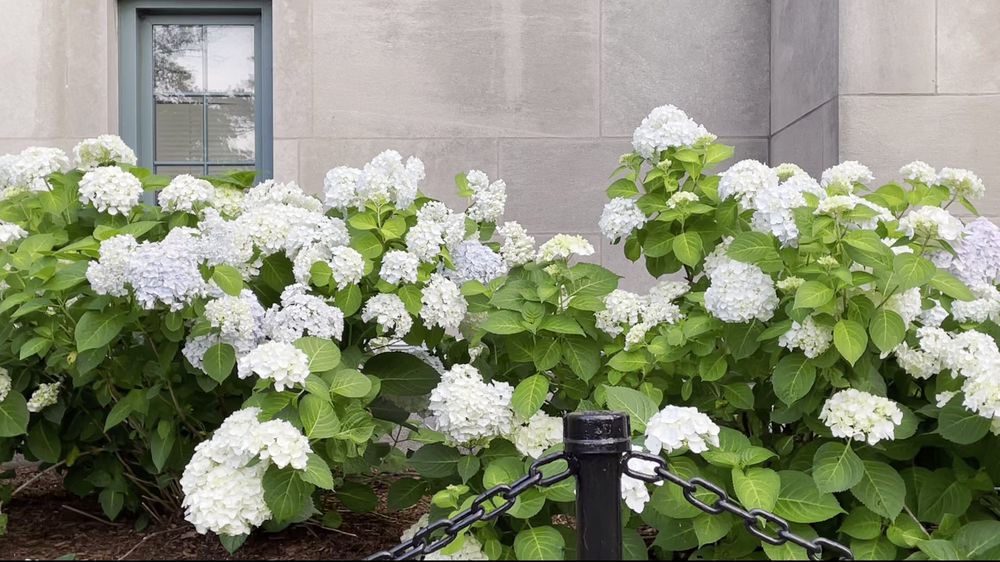} &
        \includegraphics[width=0.22\textwidth]{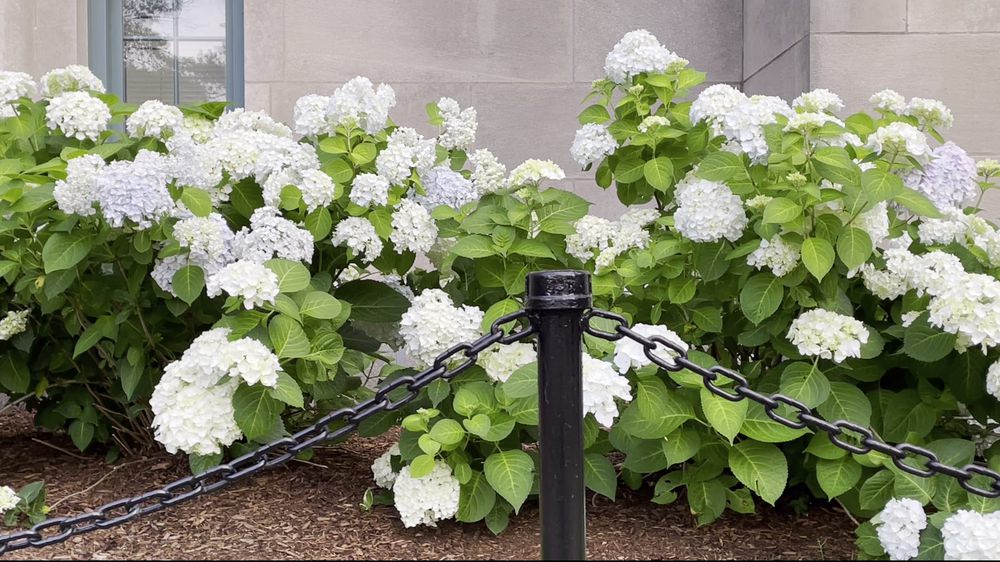} &
        \includegraphics[width=0.22\textwidth]{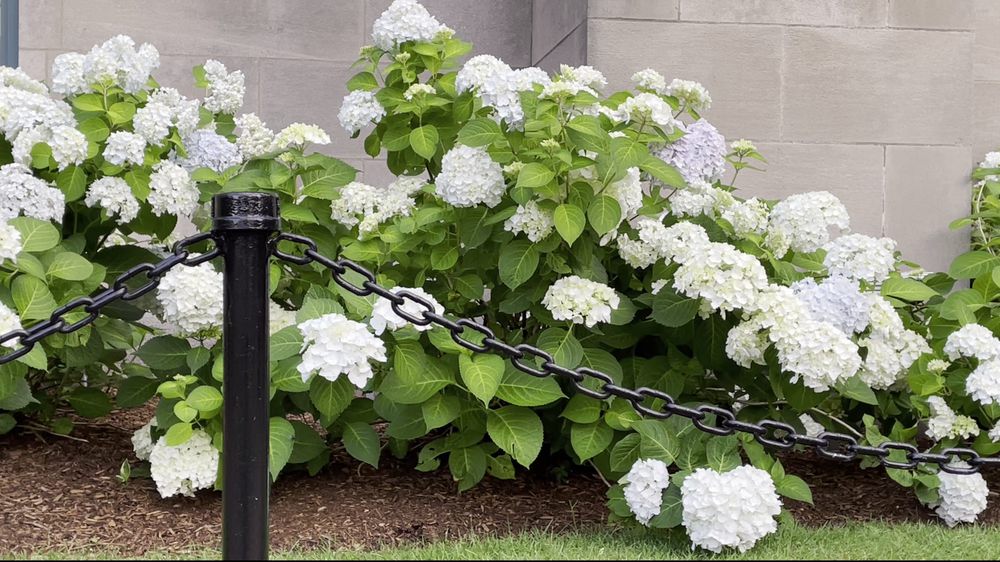} &
        \includegraphics[width=0.22\textwidth]{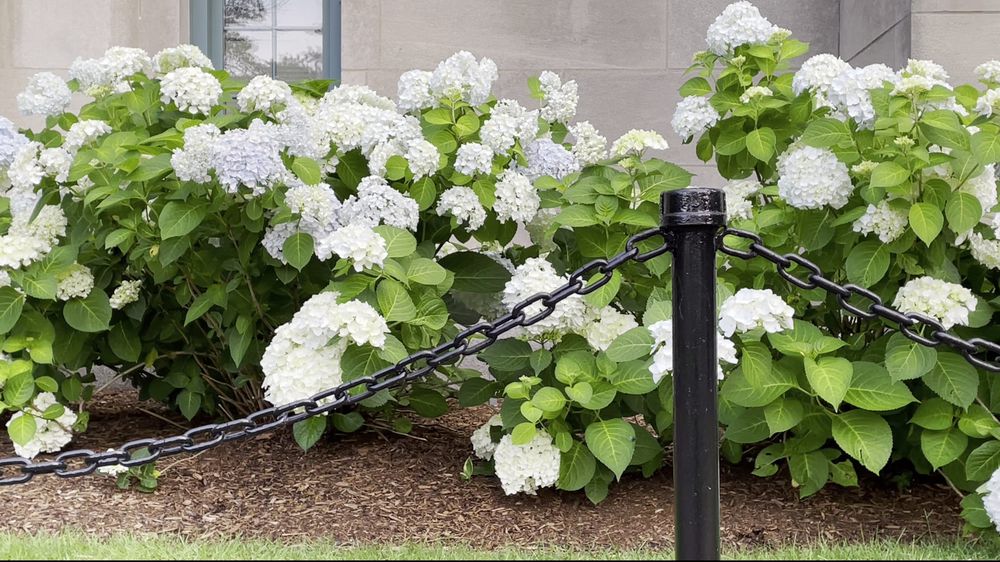} \\ [-0.7mm]
        \includegraphics[width=0.22\textwidth]{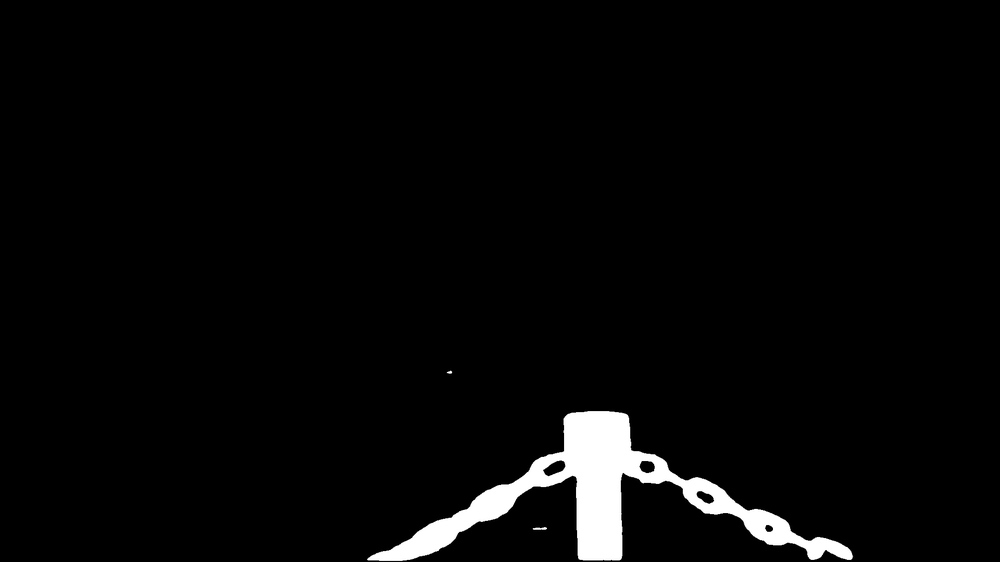} &
        \includegraphics[width=0.22\textwidth]{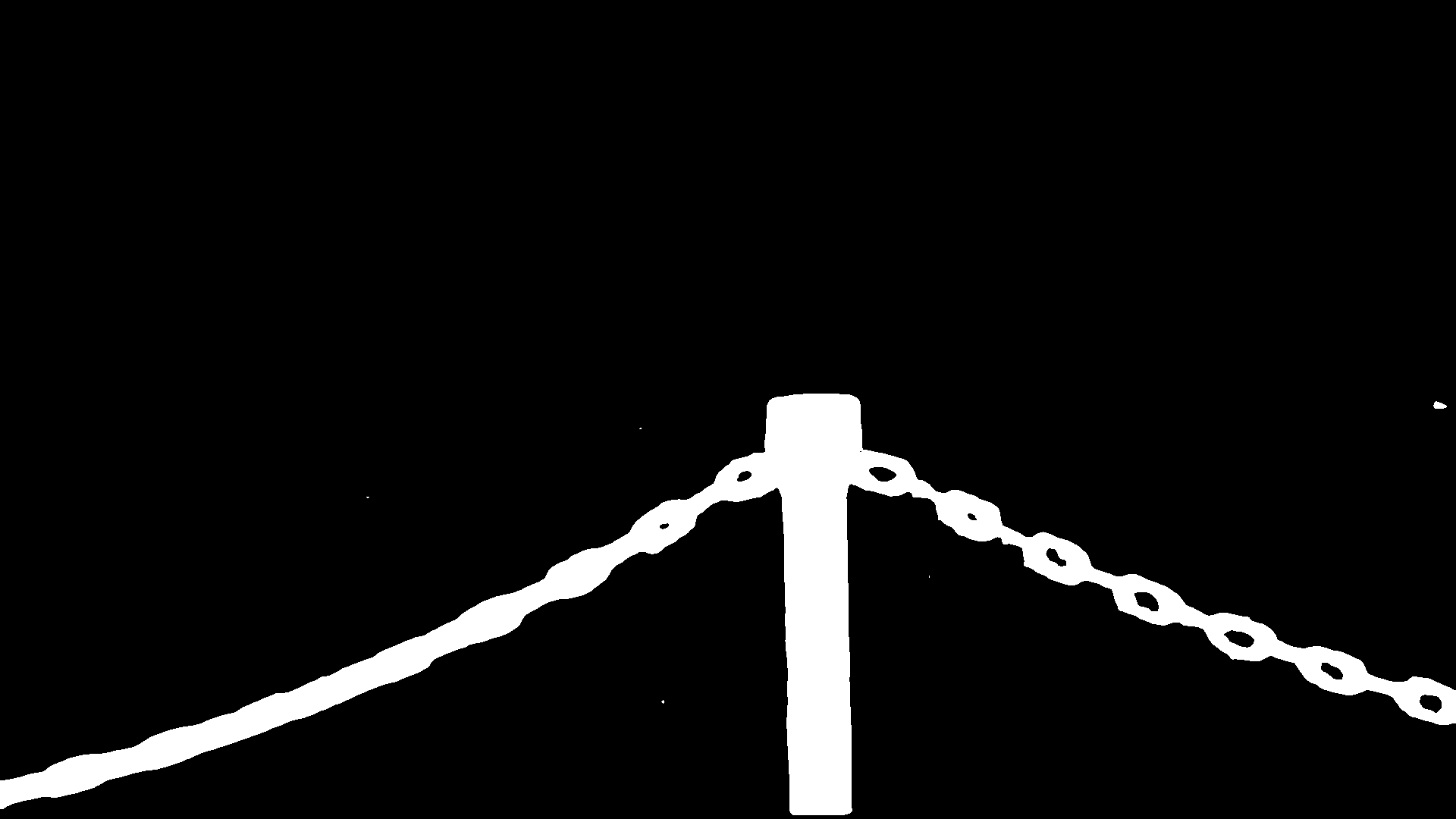} &
        \includegraphics[width=0.22\textwidth]{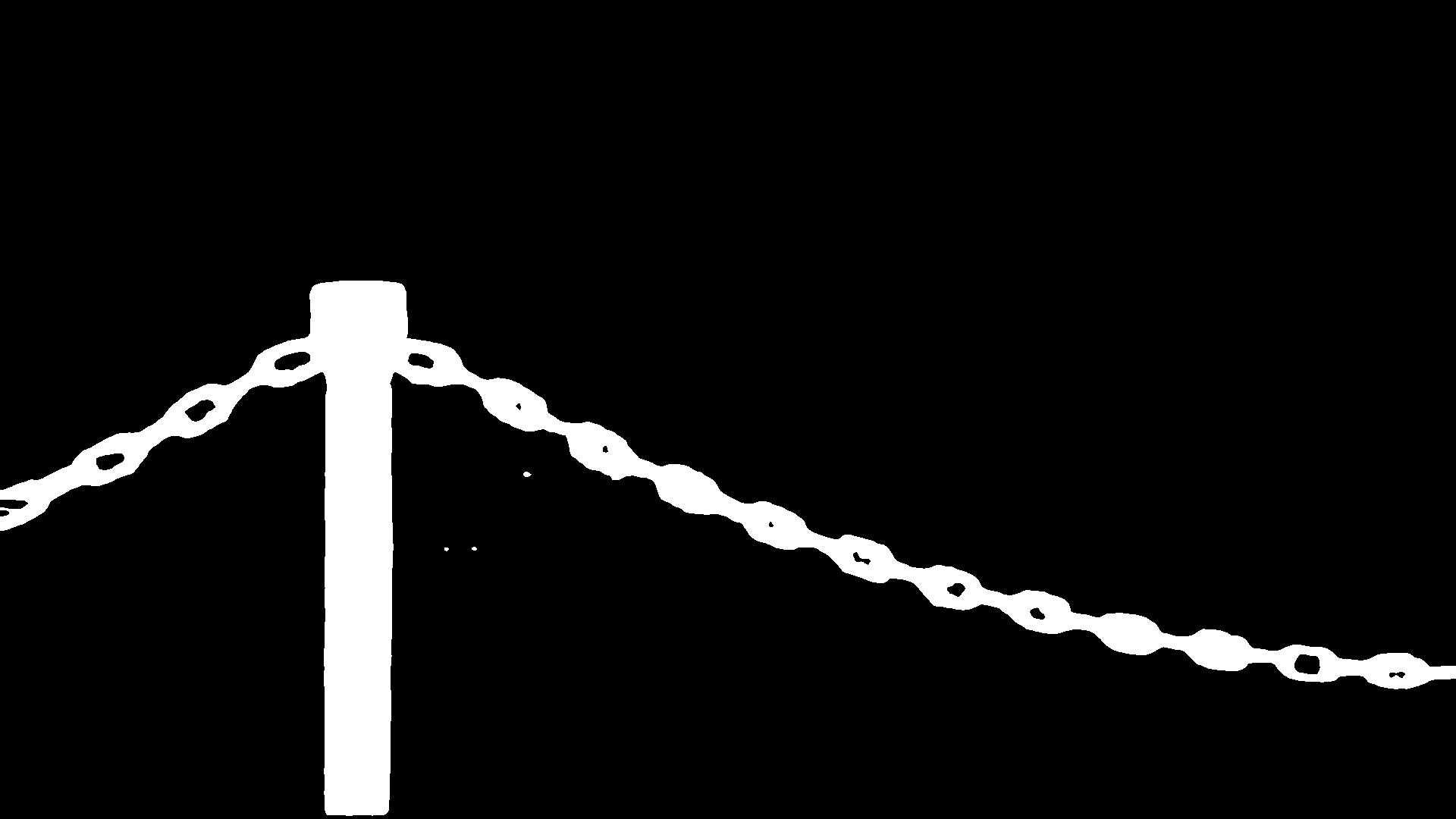} &
        \includegraphics[width=0.22\textwidth]{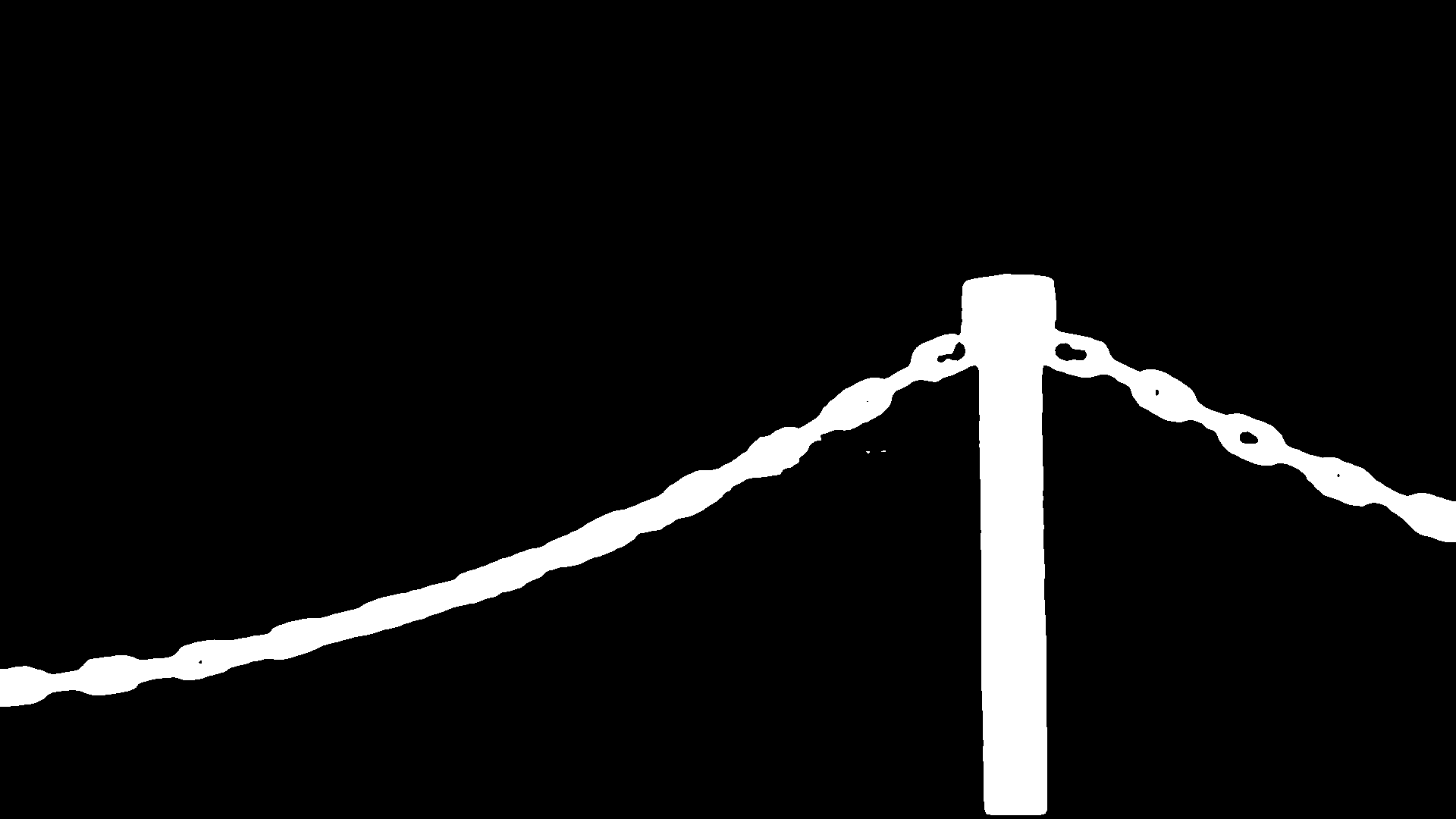} \\ [-0.7mm]
        \includegraphics[width=0.22\textwidth]{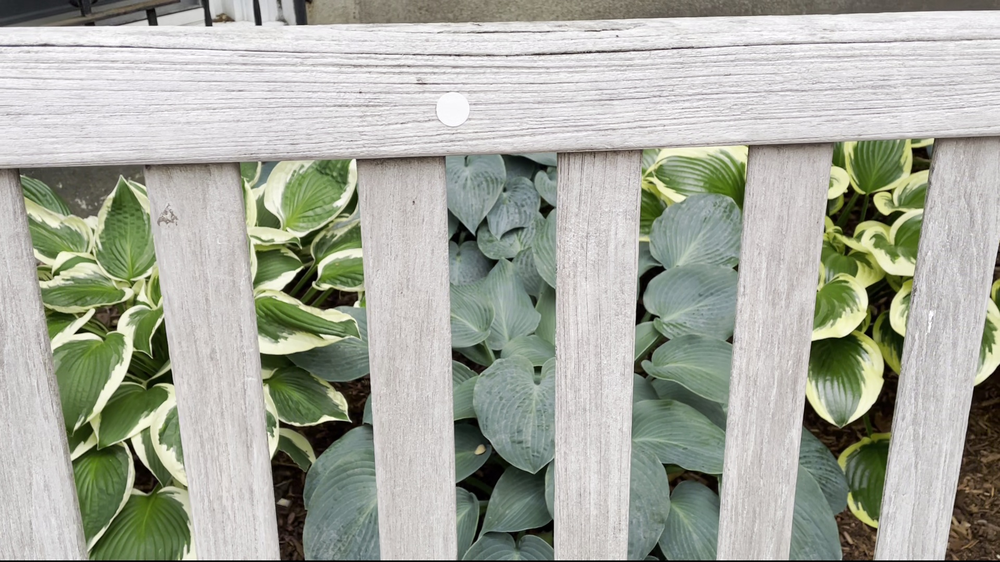} &
        \includegraphics[width=0.22\textwidth]{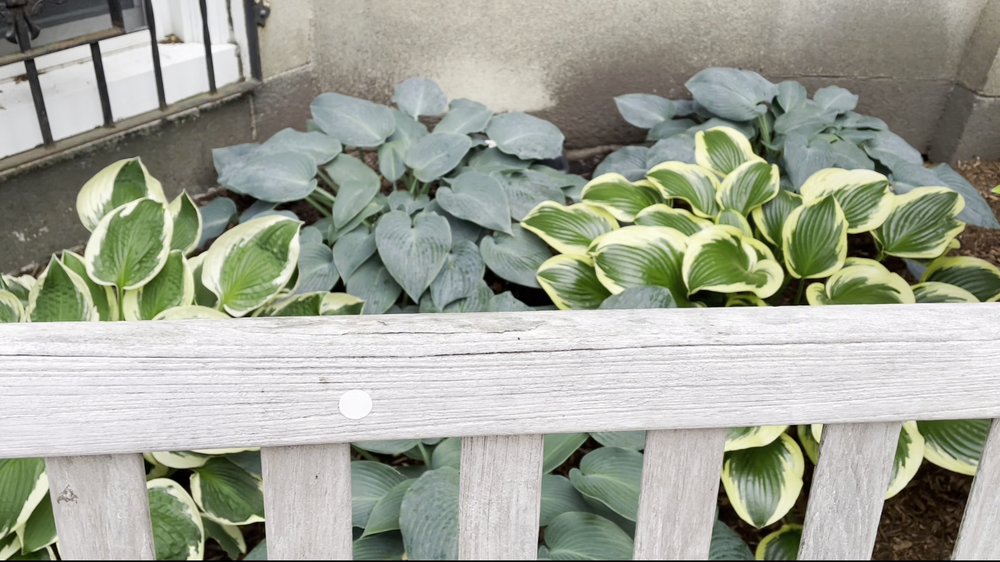} &
        \includegraphics[width=0.22\textwidth]{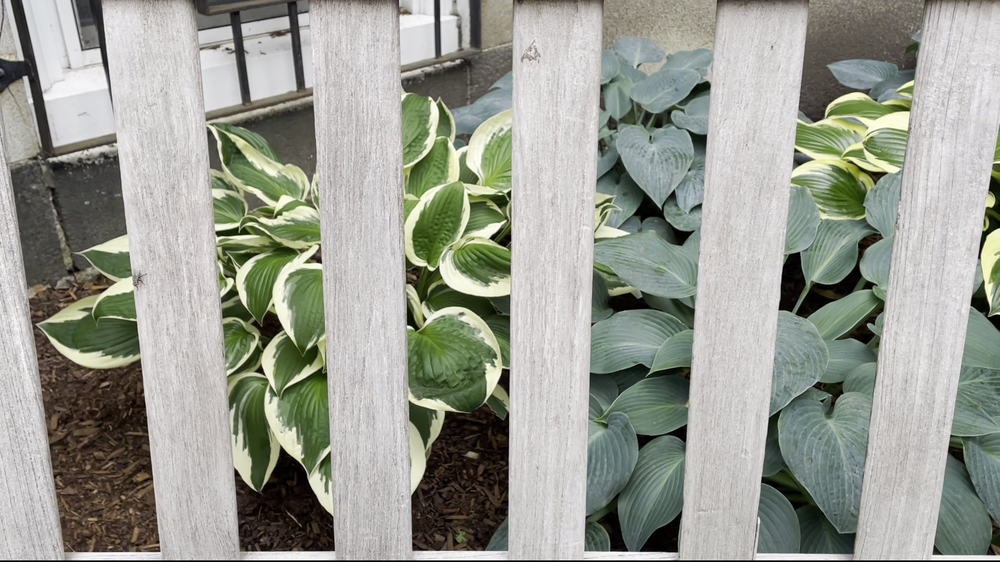} &
        \includegraphics[width=0.22\textwidth]{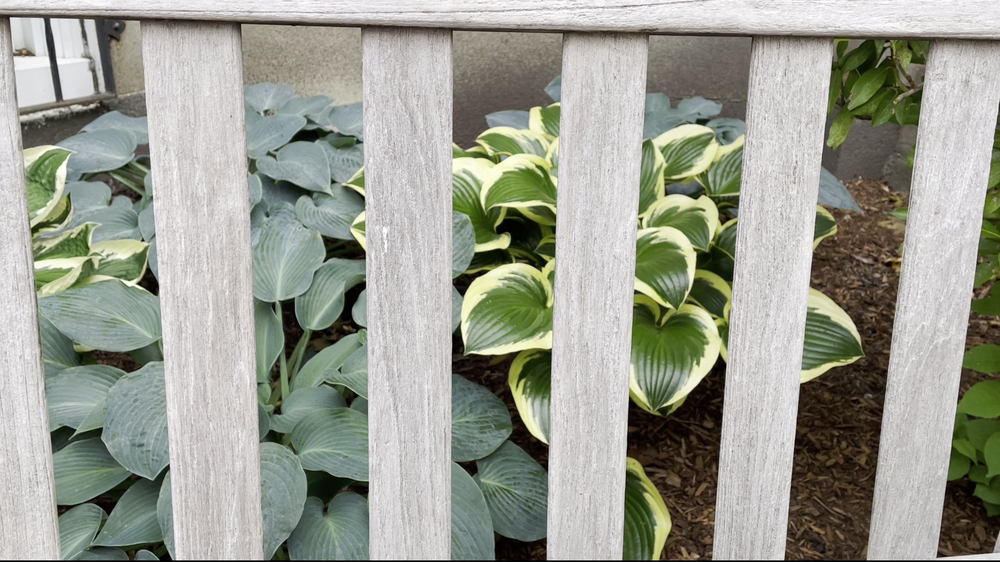} \\ [-0.7mm]
        \includegraphics[width=0.22\textwidth]{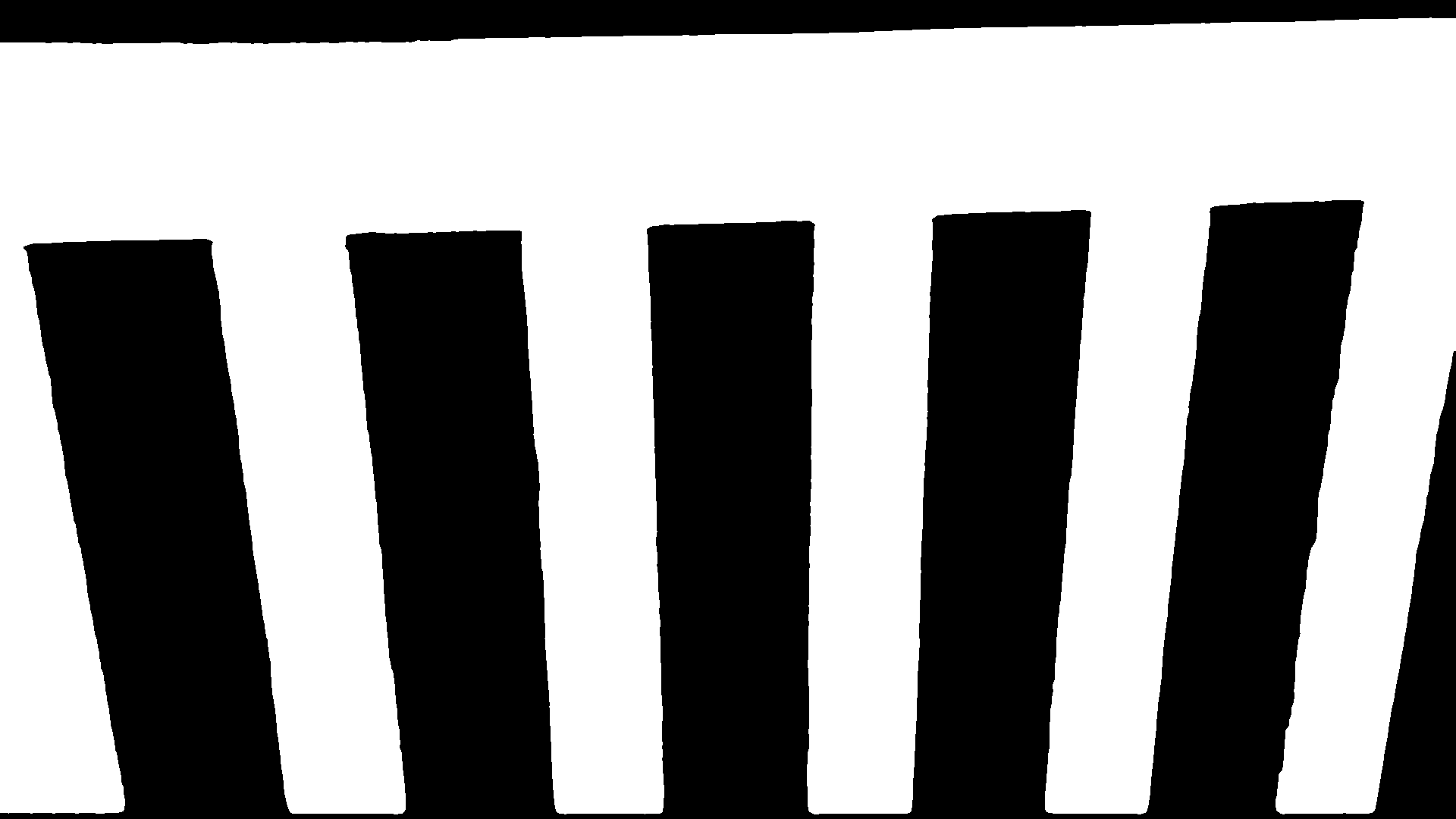} &
        \includegraphics[width=0.22\textwidth]{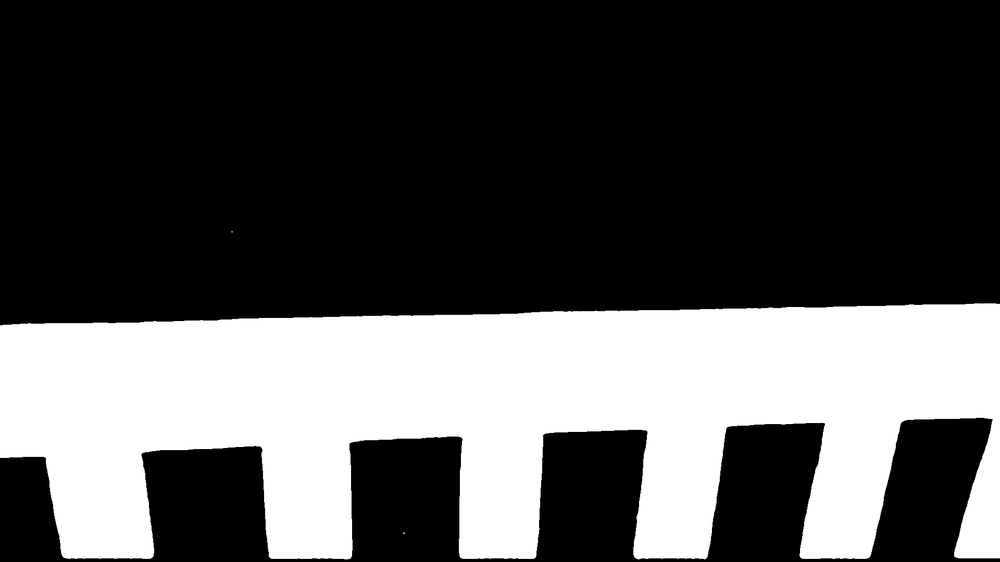} &
        \includegraphics[width=0.22\textwidth]{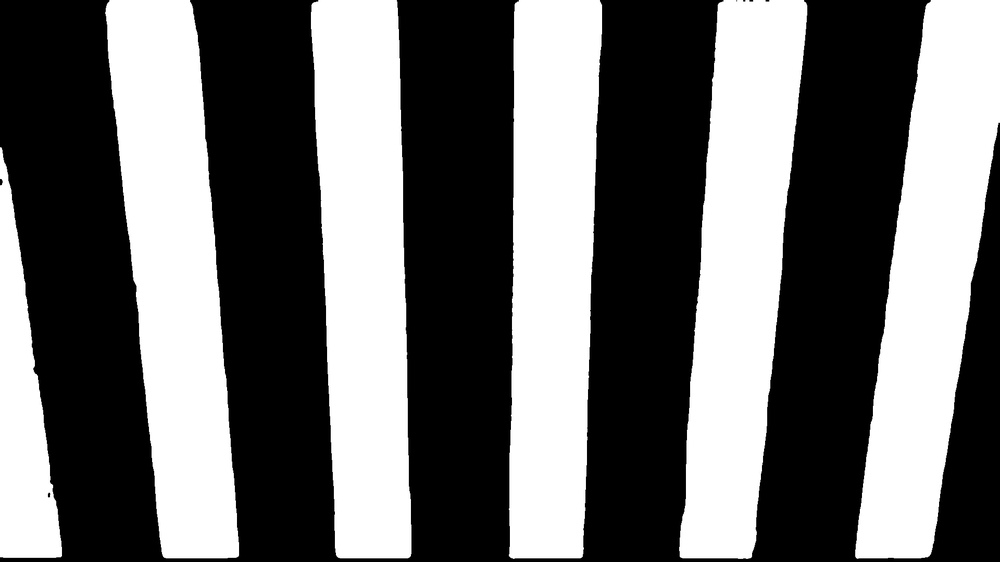} &
        \includegraphics[width=0.22\textwidth]{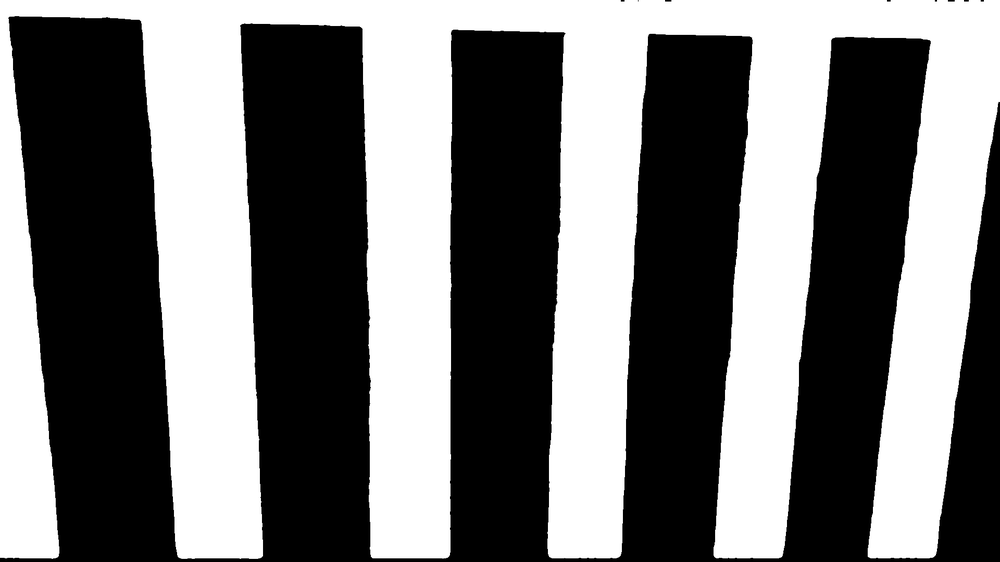}
    \end{tabular}
    \vspace{-2mm}
    \caption{\textbf{DeclutterSet Illustration (Part II).} (e) Stone Column, (f) Lamp Post, (g) Chain Fence, (h) Chair Back.}
    \label{fig:data_2}
\end{figure*}

Figure~\ref{fig:data_1} and Fig.~\ref{fig:data_2} show more samples from our DeclutterSet. We select image frames that are evenly distributed to characterize our dataset:  (i) wider distance distribution, (ii) larger occluded regions, (iii) greater relative motion between viewpoints and occluders, and (iv) more uncertain occluder shapes and mask layouts.

\section{Additional Qualitative Results}
\label{sec:add_qua_res}

Figure~\ref{fig:res_fig_spl} shows additional visual results on our collected dataset. Beyond normal results, our method demonstrates remarkable robustness by producing high-quality renderings even when faces with incorrect camera parameters from OCC-NeRF data. This issue originates from the OCC-NeRF dataset itself. Specifically, while incorporating existing scenes to complement DeclutterSet, we observed that the \textit{Railing} scene in OCC-NeRF suffers from camera calibration inconsistencies. Although we attempted to re-estimate the camera poses using COLMAP, the anomalies persisted. Nonetheless, we retained this scene in our dataset to reflect the realistic challenges posed by imperfect calibration---an inherent difficulty in occlusion removal tasks. As shown, baseline methods without camera parameter optimization fail to generate converged results and coherent reconstructions. OCC-NeRF produces only blurred representations, while our method successfully recovers a clear scene despite the adverse calibration conditions.

\noindent\textbf{Failure Cases.} The label ``FAIL'' in qualitative results is used to denote two distinct failure cases. (i) For SPIn-NeRF, it indicates that reconstruction was not accessible even before rendering, due to the lack of reliable depth information provided by COLMAP. (ii) For MVIP-NeRF, it refers to a failure that occurred during rendering, where the training process did not converge, resulting in extremely blurred and semantically meaningless images. 

To balance reconstruction quality and memory usage when using SPIn-NeRF with COLMAP, we uniformly apply a downsampling factor of $4$.

\section{Statement}
\label{sec:statement}

\subsection{Ethics Statement}
Due to concerns about the misuse of generative models and image processing techniques, both 2D and 3D generation have to face these issues. Our DeclutterNeRF, which does not employ any generative priors, mitigates these concerns to a certain extent. This approach helps to avoid potential ethical issues associated with generative models while still achieving effective results in our specific domain.

\subsection{Open Source Statement}
Through extensive experimentation with numerous baseline methods, we have identified some opportunities for improvement in the field. Many technical repositories lack proper maintenance and guidance. We recognize that to achieve occlusion removal in NeRF, 3DGS and similar fields, it is first necessary to remove the barriers that exist in the dissemination and communication of these technologies. To this end, all code and data will be open-sourced under the MIT license for community use, fostering transparency and collaborative advancement in the field.

\clearpage


\end{document}